\documentclass{article}

\usepackage{PRIMEarxiv}

\usepackage[utf8]{inputenc} 
\usepackage[T1]{fontenc}    
\usepackage{hyperref}       
\usepackage{url}            
\usepackage{booktabs}       
\usepackage{amsfonts}       
\usepackage{nicefrac}       
\usepackage{microtype}      
\usepackage{lipsum}
\usepackage{fancyhdr}       
\usepackage{graphicx}       
\graphicspath{{media/}}     
\usepackage{amsmath}
\usepackage{tabularray}
\usepackage{adjustbox}
\usepackage{pgffor}  
\usepackage{tabularx} 
\usepackage{threeparttable}
\usepackage{booktabs,caption}
\usepackage{algpseudocode}
\usepackage{algorithm}
\title{Harnessing Collective Intelligence Under a Lack of Cultural Consensus
}

\author{
  Necdet Gürkan \\
  The School of Business \\
  Stevens Institute of Technology \\
  Hoboken, New Jersey\\
  \texttt{ngurkan@stevens.edu} \\
   \And
  Jordan W. Suchow \\
  The School of Business \\
  Stevens Institute of Technology \\
  Hoboken, New Jersey\\
  \texttt{jws@stevens.edu} \\
}

\begin{document}
\maketitle

\begin{abstract}
Harnessing collective intelligence to drive effective decision-making and collaboration benefits from the ability to detect and characterize heterogeneity in consensus beliefs. This is particularly true in domains such as technology acceptance or leadership perception, where a consensus defines an intersubjective truth, leading to the possibility of multiple ``ground truths'' when subsets of respondents sustain mutually incompatible consensuses. Cultural Consensus Theory (\textsc{cct}) provides a statistical framework for detecting and characterizing these divergent consensus beliefs. However, it is unworkable in modern applications because it lacks the ability to generalize across even highly similar beliefs, is ineffective with sparse data, and can leverage neither external knowledge bases nor learned machine representations. Here, we overcome these limitations through Infinite Deep Latent Construct Cultural Consensus Theory (i\textsc{dlc-cct}), a nonparametric Bayesian model that extends \textsc{cct} with a latent construct that maps between pretrained deep neural network embeddings of entities and the consensus beliefs regarding those entities among one or more subsets of respondents. We validate the method across domains including perceptions of risk sources, food healthiness, leadership, first impressions, and humor. We find that i\textsc{dlc-cct} better predicts the degree of consensus, generalizes well to out-of-sample entities, and is effective even with sparse data. To improve scalability, we introduce an efficient hard-clustering variant of the i\textsc{dlc-cct} using an algorithm derived from a small-variance asymptotic analysis of the model. The i\textsc{dlc-cct}, therefore, provides a workable computational foundation for harnessing collective intelligence under a lack of cultural consensus and may potentially form the basis of consensus-aware information technologies.
\end{abstract}

\keywords{collective intelligence, consensus beliefs, bayesian modelling, cultural consensus theory}

\section{Introduction}
Collective intelligence refers to decision-making processes that draw on the collective opinions of multiple individuals. This often results in higher-quality decisions compared to those made by a single individual, a phenomenon commonly termed the ``wisdom of the crowd'' \cite{wallsten1983state, surowiecki2005wisdom, howe2006rise}. A multiplicity of methods can serve as information-pooling mechanisms to derive collective intelligence, ranging from structured communication techniques to prediction markets and algorithmic aggregation methods. Technologies that implement these methods and facilitate the sharing of opinions and knowledge have further enhanced collective intelligence, leading to innovative approaches and methodologies for problem-solving \cite{malone2010collective}. This integration is readily apparent in practical applications; for instance, businesses now routinely leverage crowdsourcing platforms to perform tasks such as creative design, ideation, and prediction \cite{bonabeau2009decisions, bayus2013crowdsourcing, da2020harnessing}.

From their origins in taking the median response of the group \cite{galton1907vox}, traditional aggregation and belief-merging methods have been premised on the notion that each respondent provides a partial view of an objective truth (e.g., the weight of the ox) or of a single intersubjective consensus truth (e.g., who is the group's leader), while acknowledging that individuals may have unequal domain knowledge, vigilance, or bias that causes their responses to vary \cite{dawid1979maximum, palley2023boosting, budescu2015identifying}. The effectiveness of aggregation methods for deriving collective intelligence then hinges on their ability to parcel out these sources of signal, noise, and bias in the judgments respondents make, with more effective aggregation mechanisms coming closer to that objective or intersubjective truth.

However, these methods are often applied in domains where there is a lack of cultural consensus, with different subsets of people forming conflicting culturally constructed beliefs. For example, in one subset, a leader may be perceived as someone who is authoritative and decisive, while in another subset, leadership may be associated with collaborative and inclusive decision-making \cite{giessner2008license}. One subset may view face tattoos as taboo, whereas another may find them to be stylish and meaningful expressions of individual or group identity \cite{broussard2018tattoo}. A subset may avoid certain technologies because of privacy concerns, while another views them as vital for improving the quality of life \cite{lai2021has}. It quickly becomes evident that many concepts are socially learned and deeply influenced by respondents' personal and cultural beliefs, leading to the possibility of multiple incommensurate views \cite{berger2023social}.

Cultural Consensus Theory (\textsc{cct}) provides a statistical framework for information pooling in domains where there may be a lack of cultural consensus, enabling those who use it to infer the beliefs and attitudes that influence social practices and the degree to which respondents know or express those beliefs \cite{romney1986culture}. Consensus models based on \textsc{cct} then provide an opportunity to simultaneously study both individual and group-level differences by examining the extent to which a respondent conforms to the consensus within one or more subsets, and facilitating the representation of how people differ in terms of their level of knowledge and response biases. Researchers have applied the \textsc{cct} framework to find a practical and concise definition of beliefs that are accepted by a group sharing common knowledge. Applications are found in a wide variety of domains, including cognitive evaluation \cite{heshmati2019does}, organizational culture \cite{organizationalculture}, and online communities \cite{gurkan2022learning}. 

However, \textsc{cct} has several limitations that preclude its use as a computational foundation for harnessing collective intelligence in modern applications. First, it treats each question--answer pair independently, which prevents generalization across questions. This hinders the model's performance in the sparse data regime and leads to a cold-start problem when the model cannot generalize to new questions, even those highly similar to questions previously asked. Therefore, the number of questions required to characterize a culture's consensus beliefs scales linearly with the number of culturally held beliefs. Finally, \textsc{cct} can leverage neither existing structured knowledge bases nor pretrained learned representations that could provide relevant information about known entities and their relations. These constraints hinder the application of \textsc{cct} in more complex or dynamic cultural contexts where the interrelatedness of beliefs and the availability of preexisting knowledge could play a pivotal role.


These issues find solution in modern machine-learning techniques that enable one to make predictions about out-of-sample items by learning a model over latent representations. Machine learning algorithms can analyze vast quantities of data from various modalities, including text, images, and audio, and identify patterns and relationships that generalize beyond the training data \cite{yang2022getting, liu2020predicting, chen2023deep}. Leveraging these powerful techniques, it is now possible to create vector-feature representations of words, sentences, visual scenes, and images of objects. These high-dimensional representations at times approximate human mental representations \cite{lecun2015deep}. Although they are not comprehensive theories of human cognition, vector representations of various real-world objects and concepts have been used as inputs to linear models that can predict individual and aggregate evaluations on a wide range of topics, including perceived risk \cite{bhatia2019predicting}, first impressions based on faces \cite{peterson2022deep, gurkan2022cultural}, perceptions of leadership \cite{bhatia2022predicting}, and evaluation of creative writing \cite{johnson2022divergent}.

Even so, machine-learning methods are not immune to the very same limitations that beset traditional aggregation methods, and they similarly fail in domains that lack a clear cultural consensus. For instance, a computer vision algorithm might be used to predict a label from an image without recognizing that different subsets of respondents have distinct normative responses regarding which label ought to be applied. Or, a natural-language processing algorithm might be used to predict whether user-generated content is consistent with the norms of an online community, without recognizing that different subsets of respondents have distinct normative responses regarding what content is deemed acceptable. This is particularly relevant in domains where a consensus among respondents defines an intersubjective truth, leading to the possibility of multiple ``ground truths'' when subsets of respondents sustain mutually incompatible consensuses.

To combine the strengths of \textsc{cct} and machine-learning methods while addressing their respective limitations, we propose the i\textsc{dlc-cct}, an extension to the \textsc{cct} that (1) allows culturally held beliefs to take the form of a \textit{deep latent construct}: a fine-tuned deep neural network that maps the features of a concept or entity to the consensus response among a subset of respondents, and (2) draws these deep latent constructs from a Dirichlet Process using the stick-breaking construction. The approach therefore aligns pretrained machine representations to both group- and individual-level judgments, effectively capturing variations in belief processes and behaviors across them under a lack of cultural consensus. We evaluate the i\textsc{dlc-cct} on people’s judgments of various phenomena, including risk sources, leadership effectiveness, first impressions of faces, and humor. This refined technique has broad scientific and practical applicability in domains where (1) social scientists and organizations currently study group-level variation but do not leverage machine-learning methods to gain insight, and (2) computer scientists and behavioral scientists apply machine-learning methods to harness collective intelligence without considering conflicting culturally constructed beliefs.

The plan for the paper is as follows. In Section 2, we introduce advancements in deep learning and applications of word, sentence, and image embeddings. Section 3 reviews the \textsc{cct} and provides a mathematical formalism for continuous responses that serves as the foundation for our extended model. Following that, in Section 4 we introduce our extended deep latent-construct \textsc{cct} model, the i\textsc{dlc-cct}. Sections 5 and 6 describe the datasets and methods used to validate the model. Finally, in Sections 7 and 8 we present and discuss the results obtained from fitting the model to data across the several domains.

\section{Machine representations}

Recent advancements in deep learning have given rise to expressive high-dimensional vector representations, known as embeddings \cite{lecun2015deep}. These embeddings capture the relationships, patterns, and similarities of entities such as words or objects. Deep learning involves training large multi-layered neural networks, consisting of input, hidden, and output layers \cite{lecun2015deep}. The input layer receives data (e.g., image or text), while hidden layers transform the data into intermediate representations, and the output layer generates a response (e.g., label or action). These intermediate representations, or embeddings, effectively capture expressive features of high-dimensional data points, allowing for successful mapping to responses. Embeddings are versatile inputs for a variety of predictive and explanatory models \cite{gabel2022product, guzman2023measuring, yang2022getting, chen2023deep, lix2022aligning}. Researchers employ different deep-learning techniques to generate these embeddings based on the specific application \cite{ganaie2022ensemble}. Subsequent sections will discuss word, sentence, and image embeddings in detail.

\subsection{Sentence and Word Representations}

The field of distributional semantics concentrates on automatically deriving embeddings for real-world concepts from large-scale natural language data \cite{boleda2020distributional}. These models are based on the understanding that a significant portion of human knowledge is embedded in word co-occurrence patterns \cite{bhatia2019distributed}. By leveraging these patterns, researchers can create word embeddings as vectors in high-dimensional semantic spaces or as collections of probabilistic semantic topics \cite{griffiths2007topics, jones2007representing, mikolov2013distributed, pennington2014glove}. The resulting vectors show that closely associated words, which are often discussed and referenced in similar contexts, have similar representations and are therefore closer to each other in the semantic space. 

In most use cases, ``pre-trained'' distributional semantic representations for concepts serve as inputs for more sophisticated models. These models are then ``fine-tuned'' using participant data, allowing them to approximate the structured knowledge inherent in participants' responses. Applications include predicting the perceived riskiness of various hazards \cite{bhatia2019predicting}, evaluating perceived leadership \cite{bhatia2022predicting}, modeling stereotypes and prejudices that bias social judgments \cite{caliskan2017semantics}, and extracting team cognitive diversity \cite{lix2022aligning}. In addition, recent deep learning advancements can generate vector representations that capture key aspects of sentence meaning \cite{devlin2018bert, vaswani2017attention, liu2019roberta}, which can then be fine-tuned for downstream tasks using secondary machine learning models (see \cite{wolf2020transformers} for an intuitive explanation). Researchers have applied these methods to study people's semantic cognition, such as commonsense \cite{bosselut2019comet}, relation knowledge \cite{bouraoui2020inducing}, similarity judgements \cite{marjieh2022predicting}, and individual's propensity ratings \cite{singh2022representing}.

\subsection{Image representations}

Deep neural networks have demonstrated the ability to match or surpass human performance on benchmarks for image classification \cite{he2015delving}, detection \cite{ribli2018detecting}, and recognition \cite{lu2015surpassing}. Training these networks on large-scale database of images enables them to learn versatile feature sets that generalize effectively to real-world settings. The representations discovered by deep neural networks can be used in models of human behavior for perceptual tasks, such as predicting the memorability of objects in images \cite{khosla2015understanding} and predicting human typicality judgments \cite{lake2015deep}. While their predictive power is evidence of relevance to human judgments, these models fail to fully capture the structure of human psychological representations \cite{zhuang2021unsupervised}. Comparison of these two representations is challenging because human psychological representations cannot be directly observed. Researchers thus have combined representations obtained from machines with psychological models to examine the correspondence between the two \cite{peterson2016adapting} and bring them into closer alignment \cite{ battleday2020capturing, peterson2022deep}.

\section{Cultural Consensus Theory}

The underlying principle of consensus-based models is that, for many tasks, a group's central tendency often yields a precise outcome. The collective response can serve as a substitute for the actual answer when assessing individual group members—those whose judgments are closer to the group's central tendency (across multiple questions) are presumed to possess greater knowledge. Consequently, consensus-based models can be employed to estimate an individual's level of knowledgeability when there is no available ground truth. Cultural Consensus Theory (\textsc{cct}), developed by \cite{romney1986culture}, is a theory and method that outlines the conditions under which agreement among individuals can indicate knowledge or accuracy. Many folk epistemological systems are based on this very relationship between consensus and truth; notable examples include the jury system, decentralized content moderation in online communities, and numerical ratings on review aggregation sites. The principle is even reflected in expressions such as ``50,000,000 Elvis fans can't be wrong.''

Researchers employing \textsc{cct} strive to measure the consensus from respondents' individual responses in cases where the researchers do not know the consensus ahead of time, nor which respondents have more or less knowledge. Indeed, the problem solved by \textsc{cct} is akin to determining the answer key to a test given to respondents while simultaneously grading those respondents with respect to the answer key \cite{batchelder1988test, oravecz2015hierarchical}. Additionally, like related Item Response Theory methods \cite{embretson2013item}, \textsc{cct} measures the difficulty of the questions. In \textsc{cct}, this is accomplished with cognitive-based response process models, with consensus answers and cognitive characteristics of the respondents estimated endogenously. Thus, \textsc{cct} is useful to researchers in situations where (a) the consensus knowledge of the respondents is unknown to the researcher ahead of time, (b) the researcher has access to a limited number of respondents who may or may not have equal access to this shared cultural knowledge, (c) the researcher can construct a relevant questionnaire but does not know which questions are more or less difficult, and (d) the researcher does not know much a priori about the characteristics of the respondents.

The \textsc{cct} framework additionally takes into account each respondent’s response biases, which are the shift and scale parameters of the function that maps their latent appraisal to a location on the response scale. An example of a scale bias is a tendency to mark most values either at the outer ends or in the middle section of the scale. An example of a shift bias is a tendency to mark values more frequently on one side of the scale. These parameters derive from a function often used to scale bias in probability estimation in the range (0, 1), called the Linear in Log Odds function \cite{fox1995ambiguity}. Thus, the scale bias parameter estimates the shrinkage or expansion of the latent appraisals, while the shift bias parameter estimates a shift to the left or right. 

The first \textsc{cct} model, the General Condorcet Model (GCM), was developed for binary data (true/false responses) and assumes that the consensus truth of each item is also a binary value \cite{romney1986culture}. The GCM has been widely used in the social and behavioral sciences \cite{weller2007cultural}. \cite{batchelder2012cultural} introduced an alternate assumption to the GCM to extend it to continuous truths. An extensive \textsc{cct} model for ordinal data was developed using a Gaussian appraisal model \cite{anders2015cultural}. In addition, \textsc{cct} models for continuous response data were developed to estimate and detect cultural consensuses, respondent knowledge, response biases, and item difficulty from continuous data \cite{anders2014cultural, batchelder1988test}.

Here, we describe the Continuous Response Model (CRM), developed by \cite{anders2014cultural} and that allows for multiple consensus truths, which serves as the basis for our extension to the model.

\subsection{Continuous Cultural Consensus Theory}
As a starting point, consider the Continuous Response Model (\textsc{crm}) \cite{anders2014cultural}, a cultural consensus model for continuous data derived from observations of the random response profile matrix $\textbf{X}_{ik} = (\emph{X}_{ik})_{N\times M}$ for $N$ respondents and $M$ items, where each respondent's response falls within $(0, 1)$ or a finite range that permits a linear transformation to $(0, 1)$. The \textsc{crm} links the random response variables in (0, 1) to the real line through the logit transform $\emph{X}^* = \textrm{logit}(\emph{X}_{ik})$. Therefore, each item has a consensus value in $(-\infty, \infty)$.

The \textsc{crm} is formalized and further explained by the following axioms:

\textbf{Axiom 1} (\emph{Cultural truths}). There is a collection of of $\emph{V} \geq 1$ latent cultural truths, 
$\{T_1, . . ., T_v, . . ., T_V \}$, where $T_V \in \ \prod_{k=1}^{M} (-\infty, \infty) \ $. Each participant \emph{i} responds according to only one cultural truth (corresponding consensus locations), as $T_{\Omega_i}$, where $\Omega_i \in \{1, . . . ., V \}$, and parameter $\Omega = (\Omega_i)_{1\times N}$ denotes the cultural membership for each respondent.

\textbf{Axiom 2} (\emph{Latent Appraisals}). It is assumed that each participant draws a latent appraisal, $Y_{ik}$, of each $T_{\Omega_{ik}}$, in which $Y_{ik} = T_{\Omega_{i}k} + \epsilon_{ik}$, The $\epsilon_{ik}$ error variables are distributed normal with mean 0 and standard deviation $\sigma_{ik}$.

\textbf{Axiom 3} (\emph{Conditional Independence)}. The $\epsilon_{ik}$ are mutually stochastically independent.

\textbf{Axiom 4} (\emph{Precision.}). There are cultural competency parameters $\textbf{E} = (E_i)_{1 \times N}$ with all $E_i > 0$, and item difficulty parameters specific to each cultural truth $\Lambda = (\lambda_{k})_{1\times\emph{M}}$, $\lambda_k > 0$ such that
\begin{equation}
    \sigma_{ik} = \lambda_k / \emph{E}_i.
\end{equation}
If all item difficulties are equal, then each $\lambda_k$ is set to 1.

\textbf{Axiom 5} (\emph{Response Bias}). There are two respondent bias parameters that act on each respondent's latent appraisals to arrive at the observed responses $\emph{X}_{ik}$. These include a scaling bias, $\textbf{A}=(\emph{a}_i)_{1\times N}, \emph{a}_i > 0$; and shifting bias $\textbf{B} = (\emph{b}_i)_{1\times N}, -\infty < \emph{b}_i < \infty$, such that
\begin{equation}
    X^*_{ik} = a_iY_{ik} + b_i.
\end{equation}

These axioms are developed to model the continuous responses of respondents that differ in cultural competency, $\emph{E}_i$, and response biases, $\emph{a}_i$ and $\emph{b}_i$, to items that have different shared latent truth values. The respondents have a latent appraisal with a mean at the item's consensus location, plus some error, which depends on their competence and the item difficulty. Axiom 1 locates the item truth values in the continuum. Axiom 2 defines the appraisal error is normally distributed with mean zero. Axiom 3 sets the appraisals to be conditionally independent given the respondents' cultural truth and the error standard deviations. Axiom 4 specifies the standard appraisal error that depends on the respondent's competence and item difficulty. Axiom 5 sets each respondent's response shift and scale biases.

\section{Deep Latent-Construct \textsc{CCT}}

\textsc{cct} represents the structure of culturally held beliefs as a lookup table, where keys correspond to questions and values to answers. In this representation, questions and answers lack defined internal structures and are connected only through correlations across respondents' answers. However, this formulation comes with several limitations. First, it treats each question-answer pair independently, preventing information from one question from informing our understanding of others. Second, the number of questions required to characterize a culture scales linearly with the number of culturally held beliefs. Finally, there is no way to benefit from existing structured knowledge bases that provide information about known entities and their relations.

We begin by extending \textsc{cct} to provide a more sophisticated representation of culturally held beliefs than a mere lookup table. In this extension, we define these beliefs as a \textit{latent construct}, a function that maps a question to a consensus answer through an intermediate representation \cite{van2020cultural}. We consider latent constructs that are structured as deep neural networks fine-tuned using a linear readout layer that is specific to a particular culture. Thus the latent construct as it applies to a particular culture is a combination of pre-trained embeddings and the regression weights associated with that culture.

Formally, then, we introduce two elements to the model described in Section 3.1: (1) pre-trained embeddings $\phi_k$ that represent a featurization of the entity $k$ under question, and (2) the culture-specific regression weights $\omega_{\Omega_i}$. The relationship between the regression weights of the latent construct and the embeddings is encapsulated in the embeddings is represented through the regression equation
\begin{align} \label{bayesian regression1}
\begin{split}
    T_{vk} = \phi_k\omega_{\Omega_i}^T \\
    Y_{ik} = T_{vk} + \epsilon_{ik},
\end{split}
\end{align}
where $\epsilon_{ik}$ is the error variable in Axiom 4 (Eq. 1). We then replace the consensus location described in Axiom 1 with a function that takes as input the embeddings and corresponding weights for each feature and outputs the consensus. 

When fitting the latent construct, we use Bayesian Ridge regression to regularize its weights. The prior for the coefficients, $\omega_{\Omega_i}$, is given by a spherical Gaussian:
\begin{equation}\label{weightpriors}
    p(\omega_{\Omega_i} \mid \zeta) = \text{Normal}(\omega_{\Omega_i} \mid 0, \zeta^{-1}\textbf{I}_p),
\end{equation}
with the prior over $\zeta$ assumed to be Gamma distributed, the conjugate prior for the precision of the Gaussian. 

\subsection{Infinite Deep-Latent Construct \textsc{CCT}}
In formulations that allow for the possibility of multiple cultures, \textsc{cct} analyzes eigenvalues obtained from the cross-participant correlation matrix to determine the number of cultures present \cite{anders2014cultural}. In the context of our Bayesian formulation of deep latent construct \textsc{CCT}, we employ a Bayesian non-parametric technique in which cultures (and their associated latent constructs) are drawn from a Dirichlet Process (DP) by way of the stick-breaking construction \cite{ishwaran2001gibbs}. The Dirichlet Process serves as a prior over discrete distributions and is particularly useful for mixture models, where the number of components is not known a priori and can grow with the data. The DP's attribute of being an infinite discrete prior has led to it being widely applied not only to mixture models \cite{griffiths2003hierarchical, neal2000markov}, but also to psychometric models as a prior over probabilities, facilitating flexible modeling of individual differences and underlying traits \cite{duncan2008nonparametric, miyazaki2009bayesian}. Using this method in the context of \textsc{CCT} provides an end-to-end probabilistic approach to learning the number of cultures needed to account for the observed data that simultaneously learns the culture assignments, latent constructs, and individual-level parameters. These parameters can be marginalized over the discrete cultural membership indicators $\boldsymbol{z}$ by using an efficient posterior inference algorithm (e.g., \textsc{adi}, \textsc{nuts}, \textsc{hmc}) for learning the joint posterior of the remaining model parameters.

Although the model assumes an unbounded number of cultures in theory, in practice, for a given set of respondents, only a finite number of cultures will have at least one respondent assigned to them. To make the model more computationally feasible and efficient, we truncate the Dirichlet Process by selecting an upper bound that caps the possible number of cultures. This approach strikes a balance between enabling the model to account for a variety of cultures while keeping resource consumption, such as processing time and memory, in check.

\subsection{Hierarchical specification of the extended \textsc{CCT}}

In this section, we provide a hierarchical specification of the generative model that underpins Infinite Deep Latent Construct (i\textsc{dlc-cct}), where population distributions are specified for the parameters using hyperparameters \cite{lee2011cognitive}. These hyperparameters are estimated from their own distributions and can represent the central tendency across items or respondents, which may be unique to each dataset. The hierarchical structure of the generative model is as follows:
\begin{align*}
\omega_{\Omega_i} \sim \text{Normal}(0, \zeta^{-1}) && \text{Coefficient weights} \\
T_{vk} = \phi_k\omega_{\Omega_i}^T && \text{Latent construct item location}\\
\text{log}(\lambda_{vk}) \sim \text{Normal}(\mu_{vk}, \tau_{vk}) && \text{Item difficulty} \\
\text{log}(E_{i}) \sim \text{Normal}(\alpha_{E_{\Omega_i}}, \kappa_{E_{\Omega_i}}) && \text{Respondent competency} \\
\text{log}(a_i) \sim \text{Normal}(\mu_{a_{i}}, \tau_{a_{i}}) && \text{Respondent scaling bias} \\
b_i \sim \text{Normal}(\mu_{b_{i}}, \tau_{b_{_i}}) && \text{Respondent shifting bias}\\
\Omega_i \sim \text{Categorical}(\pi) && \text{Group membership} \\
\pi \sim \text{stickbreaking}(\beta) && \text{Pr. of group membership} \\
\beta \sim \text{Beta}(1, \alpha) && \text{Group sparsity}
\end{align*}

The item consensus $T_{vk}$ is derived by applying Bayesian Ridge Regression to the pretrained embeddings. The location of the item consensus is calculated by taking the dot product of the relevant two row matrices. The other model parameters, $E_i$, $\lambda_i$, and $a_i$, which are each located on the positive half-line, are log-transformed to the real line and also assumed to be sampled from a normal population-level distribution. The shift bias, $b_i$, is located on the real line, paramaterized with a mean and precision (inverse variance). Note that the respondents' competence parameter remains singly indexed by \emph{i}, through an indexing technique in which their distribution is specified by their group membership $\Omega_i$. Culture assignments are derived via a stick-breaking prior, and this allows for varying probabilities of being in any of \emph{V} groups; note that \emph{V} is unknown priori and needs to be estimated from observed data. The sparsity of cultures, $\beta$, is sampled from a beta distribution. The variables $\Omega_i$ and $\pi$ are removed for the single-truth variant of our model.

\section{Method}
We applied the i\textsc{dlc-cct} and \textsc{dlc-cct} to a diverse range of datasets that included judgments of the healthiness of food, humor, leadership effectiveness, social attributes of faces, and risk. These datasets were chosen because they include respondents' individual responses to questions (items) that are related to their shared knowledge or beliefs, and because the nature of the domain is such that a consensus contributes to an intersubjective truth. For our i\textsc{dlc-cct} model, we rescaled these ratings to fall within the range of 0 to 1. Excluded were datasets with only aggregate responses, or where the nature of the domain is fully objective or subjective. 

Models were implemented in NumPyro \cite{phan2019composable} with the \textsc{jax} backend \cite{bradbury2020jax}. The model components were integrated into a single likelihood function and a set of prior distributions. Inference was performed using a Gibbs Sampler \cite{liu1996peskun} combined with the No-U-Turn Sampler (\textsc{nuts}) \cite{hoffman2014no}, a standard Markov Chain Monte Carlo sampling algorithm, as implemented in NumPyro. We used two chains with 10,000 warm-up samples and 10,000 draw samples, thereby obtaining 20,000 posterior samples for each model. We ensured that the posterior had converged by ensuring there were no divergent transitions and monitoring chain diagnostics such as $R_\textrm{hat}$. The code for the models and datasets are available on GitHub at [Redacted for blind submission]. 

In the next section, we elucidate the phenomena associated with each dataset, along with a thorough description of the datasets themselves.

\section{Concept and Data}
\subsection{Sense of humor}

Everyone appreciates humor. In the workplace, it offers benefits such as team bonding, employee motivation, idea generation, and frustration diffusion through venting \cite{holmes2007making, pryor2010workplace}. However, it is important to recognize that humor can also have downsides, such as causing distractions, harming credibility, or offending others in diverse work environments \cite{decker2001relationships}. The differences in how humor is enacted and understood arise from individual differences, such as gender, beliefs, and mental state \cite{duncan1990humor}. Therefore, organizations may find themselves in a complex position, responsible for considering individual differences in perception of humor to avoid negative consequences in the workplace \cite{lyttle2007judicious}.

Another stream of research focuses on the use of humor in marketing materials, such as commercial and print ads, as well as entertainment products like TV shows and web content. A common belief is that making someone laugh can lead to a purchase. Advertising and marketing research, indeed, has demonstrated that humor can enhance ad likability, partly due to its ability to evoke positive emotions \cite{cline2003does}. Humorous ads positively impact viewers' moods and elicit more positive feelings than non-humorous ads \cite{madden1988attitude}. These positive emotions may lead to a favorable attitude towards the ad \cite{alden2000effects, eisend2011humor}, increased liking of the source \cite{zhang1996responses}, and stronger persuasion \cite{walter2018priest}. While studies indicate the positive impacts of humor in marketing and advertising, the effectiveness of humor depends on the targeted audience's idiosyncratic features \cite{barta2023influencer, zhang1996effect}. Hence, incorporating humor tailored to the target audience in marketing and advertising could be vital for companies to positively influence consumers and avoid negative consequences \cite{beard2008advertising}. 

We employed the i\textsc{dlc-cct} on the Jester dataset \cite{goldberg2001eigentaste}, commonly used in recommendation system tasks. The dataset consists of 140 jokes rated by 59,132 internet users on a scale of {-}10 (extremely unfunny) to 10 (extremely funny). To avoid computational intensity that could be caused by the high number of respondents, we selected the ratings of a random subset of 5,000 users from the dataset as input for our model. Rating distribution across items is sparse due to the disparity in the number of ratings made for different items. Ratings of 112 jokes were used for training and the remaining 28 were held-out as a validation set, an 80-20 split.

We obtained embeddings for each joke from RoBERTa, a pre-trained language model \cite{liu2019roberta} that extracts features from textual data and generates a latent vector in a 768-dimensional space. These latent vectors serve as pre-trained embeddings in the model.

\subsection{First impression of faces}

Though warned not to judge a book by its cover, people nonetheless attribute a wide variety of traits (e.g., trustworthiness, dominance, or smartness) to strangers based on their facial appearance. These judgments are formed quickly and have considerable impact on human behavior in domains as diverse as loan approval \cite{chen2022s}, politics \cite{olivola2010elected}, law \cite{flowe2011examination}, business \cite{olivola2014many}, and social decisions \cite{mueller1996facial, olivola2014many}. For example, \cite{olivola2014many} showed that people form impressions of aspiring leaders from their faces, which in turn predict their success in reaching prestigious leadership positions.

Some trait judgments have high levels of inter-rater agreement within particular cultures and are shared by observers in multiple cultures \cite{sutherland2018facial, mcarthur1987cross}. Such agreement among the groups is a so-called consensus impression that is shared by many individuals or communities. For example, many individuals in the U.K. and the U.S. attribute naivety and trustworthiness to faces with large eyes and round features \cite{mcarthur1987cross}. The same participants tend to judge short, squat faces to be more aggressive than faces that are tall and thin \cite{geniole2014facial}. On the other hand, face-trait judgments are found to be idiosyncratic and differ between individuals living within the same culture \cite{sutherland2018facial}. It suggests that one’s local environment that is surrounded around family and community likely drives the observed individual differences. Similarly, \cite{stolier2020trait} showed that perceivers learn conceptual knowledge of how traits correlate that shape trait inferences across faces, personal knowledge, and stereotypes in one’s environment. These results show the importance of individual differences in first impression formation relative to cultural variability.

We applied the iDLC-\textsc{cct} to a large dataset of people’s first impressions of faces \cite{peterson2022deep}. The dataset contains over 1 million judgments of 34 trait inferences for 1,000 face images. Each face is rated by 30 unique participants for each trait. In this dataset, the face images are generated using a synthetic photorealistic image generator, StyleGAN2 \cite{karras2020analyzing}. The generator network component of StyleGAN2 models the distribution of face images conditioned on a 512-dimensional, unit-variance, multivariate normal latent variable. This vector is the pretrained embedding used for our modeling. For each trait, we used the ratings of 800 faces for training and the remainder (200 faces) for validation, an 80--20 split. 

\subsection{Risk perception}

Research on risk perception investigates people’s views when asked to describe and assess potentially dangerous activities and technologies. Such research supports risk analysis and decision-making in society, such as enhancing techniques for obtaining risk-related opinions, predicting public reactions to hazards, and refining the exchange of risk information among the public, technical experts, and policymakers \cite{slovic1987perception, kai1979prospect}. Risk is fundamentally a subjective notion \cite{slovic1987perception}, and discerning how individuals perceive risk is essential for understanding the relationship between risk and decision-making processes at the individual, group, and organizational levels \cite{slovic1987perception, slovic2013perception}. Various factors have been found to explain how people perceive a hazard, such as the risk’s characteristics \cite{fischhoff1978safe}, perceived benefits \cite{wolfenbarger2000ecological}, knowledge \cite{evans1995relationship}, personal norms \cite{de2010morality}, and affective associations \cite{finucane2000affect}. These factors could be unobserved variables that should be considered when examining the public's perception of a hazard. 

We applied the iDLC-\textsc{cct} to people’s risk perceptions of technological hazards, activities, and participant-generated risk sources \cite{bhatia2019predicting}. The dataset on risk perception of technological hazards consists of 125 technologies of varying risk levels. The items in this dataset were based on Slovic’s experiment \cite{slovic1987perception}. The dataset on risk perception of daily activities consists of 125 activities of varying risk levels. The items in this dataset were also based on Slovic’s experiment \cite{slovic1987perception}. The dataset on participant-generated risk sources consists of the 200 risk sources most frequently listed by participants, without any category limitations. We used to word2vec \cite{mikolov2013distributed} word representations to obtain an embedding for each risk source, which is in a 300-dimensional space. This vector is the pretrained embedding used for our modeling. For risk sources of technologies and activities, we used the ratings of 100 sources for training and the remainder (25 sources) for validation, an 80--20 split. For participant-generated risk sources, we used the ratings of 160 risk sources for training and the remainder (40 sources) for validation, an 80--20 split.

\subsection{Leadership perception}

Identifying the factors that predict leader selection is crucial, as a leader's influence on their organization's success impacts the well-being of its members and those affected by the organization's output \cite{jago1982leadership}. As a result, organizations and their members should be highly motivated to recognize and select effective leaders within their domain, ideally based on objective indicators of leadership quality. Leadership researchers have emphasized the importance of studying how followers perceive their leaders \cite{craig1998perceived, giessner2008license, phillips1981causal}. Although researchers’ opinions may vary regarding the characteristics of effective leaders, individuals often possess implicit leadership theories that define the cognitive structures and criteria distinguishing leaders from non-leaders, as well as effective leaders from ineffective ones \cite{lord1984test}. While skilled leaders may not always be acknowledged as such, the perception of leadership ability is essential for an individual's success in a leadership role \cite{carnes2015matters, olivola2010elected}.

We applied the iDLC-\textsc{cct} model to a dataset containing people's judgments of leadership effectiveness \cite{bhatia2022predicting}. This dataset includes assessments from 210 participants regarding 293 individuals, consisting of both leaders and non-leaders. Participants rated leadership effectiveness on a scale of 0 (extremely ineffective) to 100 (extremely effective). We utilized word2vec word representations \cite{mikolov2013distributed} to generate an embedding for each individual, projecting the word vectors into a 300-dimensional space. This vector is the pretrained embedding used for our modeling. We used the ratings of 235 individuals for training and the remainder (58 individuals) for validation, an 80--20 split.

\subsection{Food healthiness}

People hold diverse opinions regarding what constitutes a healthy diet \cite{capewell2011rapid, gurkan2022food}. These opinions are shaped by various factors, including personal beliefs, associations, attitudes, and knowledge regarding the healthiness of food \cite{messer1984anthropological}. Due to the complex interplay of psychological, cognitive, and social factors influencing consumer choices, there is no definitive answer to what makes food appear healthy \cite{lobstein2009defining}. Even experts cannot reach a consensus on the definition of a healthy diet \cite{quealy2016sushi}. Consequently, understanding people's health beliefs and perceptions of healthy food is crucial for creating culturally and socially appropriate behavior change interventions \cite{barrera2013cultural}.

Choosing healthy food is a complex behavior that involves cultural (e.g., customs, social norms), psychological (e.g., body image), and social factors (e.g., price availability, ethical concerns) \cite{pearcey2018comparative}. Prior research on healthy food perceptions across diverse groups has highlighted the unique ways individuals define healthy food \cite{banna2016cross}. Cross-cultural studies indicate that some cultures share similar views on healthy food choices and nutritional intake \cite{akamatsu2005interpretations}, while others differ \cite{banna2016cross}. Factors such as gender and education level also impact judgments about the healthiness of food \cite{turrell2006socio}. Beyond individual characteristics, front-of-package labeling is considered to have a significant influence on people's evaluations of food healthiness \cite{orquin2014brunswik}. However, judgments based on front-of-package labeling are also affected by individual features \cite{schuldt2013does}.

We applied the iDLC-\textsc{cct} model to a dataset of laypeople and experts’ judgments of food healthiness \cite{gandhi2022computational}. The lay people dataset contains judgments of 149 lay and 19 expert participants of a diverse set of 172 foods. Participants were asked to judge the healthiness of food on a scale ranging from  {-}100 (extremely unhealthy) to {+}100 (extremely healthy). We used word2vec word representations \cite{mikolov2013distributed} to obtain an embedding for each food, projecting the word vectors into a 300-dimensional space. This vector is the pretrained embedding used for our modeling. We used the ratings of 235 food items for training and the remainder (34 food items) for validation, an 80--20 split.

\section{Results}
Across all the datasets, we consistently found that the \textsc{dlc-cct} outperformed the original \textsc{cct} model, showing a reduction in the average RMSE by 0.12 and an increase in $R^2$ by 0.22 (Table 1). Because the original \textsc{cct} lacks a mechanism by which to meaningfully generalize across items based on their content, it relies only on the posterior mean of the consensus parameter's prior when predicting held-out items. In contrast, as indicated in Table 1, \textsc{dlc-cct} significantly outperforms the original \textsc{cct} by generalizing to held-out items, demonstrating the benefits of integrating a deep latent construct into cultural consensus analysis.

Likewise, we found that the i\textsc{dlc-cct} (with its multiple consensuses) outperforms the \textsc{dlc-cct} (with only a single consensus).  The \textsc{dlc-cct} yielded average RMSE and $R^2$ values of 0.22 and 0.33, respectively. In comparison, the i\textsc{dlc-cct} performed better, reflected in an average RMSE of 0.20 and an $R^2$ of 0.41. This represents a reduction in RMSE by 0.02 and an increase in $R^2$ by 0.08. Comparison between the two reveals the advantages of incorporating heterogeneity in consensus beliefs across subsets of respondents. 

\begin{table}
\caption{\label{model-comparison} Model comparison.}

\begin{threeparttable}
\fontsize{9pt}{9pt}\selectfont
\centering
\setlength{\tabcolsep}{1pt} 

\begin{tblr}{
    width = \linewidth,
    cell{1}{3} = {c=2}{c},
    cell{1}{5} = {c=2}{c},
    cell{1}{7} = {c=4}{c},
    cell{5}{1} = {r=2}{},
    cell{7}{1} = {r=3}{},
    colspec = {X[l] X[l] X[c,0.5] X[c,0.5] X[c,0.5] X[c,0.5] X[c,0.5] X[c,0.5] X[c,0.5] X[c,0.5]}, 
    vline{5,7} = {2-42}{},
    hline{2-3} = {3-10}{},
    hline{4-5,7,43,10} = {-}{},
}
                                                            &                               & \textbf{CCT} &     & \textbf{DLC-\textsc{CCT}} &      & \textbf{iDLC-\textsc{CCT}} &      &                 &              \\
                                                            &                               & $R^2$        & RMSE & $R^2$        & RMSE &     $R^2$     & RMSE & Culture entropy & No. cultures \\
\textbf{Leadership}                                        & \textit{Leadership}           & .08    & 0.29 & .31    & .24 & .35     & .24 & 2.23            & 7            \\
\textbf{Humor}                                             & \textit{Humor perception}     & .00    & .30 & .27    & .22 & .27     & .22 & 2.69            & 12           \\
\textbf{Healthy food}                                      & \textit{Lay people}           & .00    & .30 & .52    & .20 & .56     & .19 & .40            & 2            \\
                                                            & \textit{Experts}              & .00    & .31 & .43    & .21 & .43     & .21 & .74            & 2            \\
\textbf{Risk sources}                                     & \textit{Technology hazards}   & .00    & .32 & .33    & .26 & .36     & .24 & .51            & 4            \\
                                                            & \textit{Daily activities}     & .04    & .29 & .26    & .26 & .32     & .24 & .67            & 2            \\
                                                            & \textit{Participant-generated risks} & .00    & .31 & .47    & .23 & .50     & .22 & 1.44            & 3            \\
\textbf{Impressions of faces}                            & \textit{Dominant}             & .00    & .38 & .33    & .22 & .41     & .20 & 1.46            & 7            \\
                                                            & \textit{Trustworthy}          & .00    & .40 & .29    & .20 & .34     & .19 & 2.11            & 8            \\
                                                            & \textit{Smart}                & .00    & .33 & .29    & .18 & .36     & .17 & 2.17            & 5            \\
                                                            & \textit{Attractive}           & .00    & .39 & .41    & .19 & .45     & .18 & 2.29            & 6            \\
                                                            & \textit{Outgoing}             & .00    & .38 & .36    & .18 & .46     & .17 & 1.97            & 5            \\
                                                            & \textit{Age}                  & .00    & .26 & .40    & .12 & .42     & .12 & .03            & 2            \\
                                                            & \textit{Fat}                  & .00    & .30 & .37    & .15 & .44     & .13 & 1.56            & 5            \\
                                                            & \textit{Familiar}             & .00    & .35 & .47    & .20 & .48     & .20 & 0.59            & 3            \\
                                                            & \textit{Gender}               & .00    & .43 & .58    & .22 & .67     & .19 & 1.37            & 5            \\
                                                            & \textit{Typical}              & .00    & .41 & .27    & .20 & .30     & .19 & 2.55            & 8            \\
                                                            & \textit{Happy}                & .00    & .35 & .51    & .16 & .60     & .15 & 1.33            & 4            \\
                                                            & \textit{Dorky}                & .00    & .42 & .19    & .25 & .30     & .23 & 2.08            & 6            \\
                                                            & \textit{Long hair}            & .00    & .35 & .26    & .21 & .55     & .17 & 1.44            & 7            \\
                                                            & \textit{Skin color}            & .00    & .28 & .59    & .13 & .66     & .12 & 0.83            & 5            \\
                                                            & \textit{Smug}            & .00    & .35 & .27    & .24 & .41     & .22 & 2.21            & 7            \\
                                                            & \textit{Groomed}            & .00    & .42 & .28    & .20 & .36     & .19 & 1.54            & 5            \\
                                                            & \textit{Cute}            & .00    & .41 & .47    & .20 & .53     & .19 & 1.89            & 5            \\
                                                            & \textit{Alert}            & .00    & .40 & .26    & .20 & .36     & .18 & 2.16            & 6            \\
                                                            & \textit{Hair color}            & .00    & .44 & .60    & .18 & .64     & .17 & .48            & 5            \\
                                                            & \textit{Privileged}            & .00    & .35 & .31    & .19 & .39     & .17 & 2.07            & 5            \\
                                                            & \textit{Liberal}            & .00    & .26 & .16    & .23 & .20     & .22 & 1.79            & 5            \\
                                                            & \textit{Asian}            & .00    & .38 & .19    & .28 & .39     & .27 & .15            & 6            \\
                                                            & \textit{Middle-Eastern}            & .00    & .39 & .30    & .25 & .34     & .24 & 1.03            & 6            \\
                                                            & \textit{Hispanic}            & .00    & .40 & .16    & .29 & .17     & .29 & .38            & 4            \\
                                                            & \textit{Polynesian}            & .00    & .37 & .29    & .25 & .37     & .24 & 1.49            & 5            \\
                                                            & \textit{Native-American}            & .00    & .33 & .26    & .25 & .33     & .23 & .89            & 4            \\
                                                            & \textit{Black}            & .00    & .45 & .25    & .19 & .43     & .17 & .24            & 5            \\
                                                            & \textit{White}            & .00    & .52 & .38    & .28 & .60     & .23 & .86            & 4            \\
                                                            & \textit{Looks like me}            & .00    & .29 & .35    & .19 & .49     & .17 & 1.57            & 3            \\
                                                            & \textit{Electable}            & .00    & .39 & .18    & .26 & .24     & .25 & .77            & 3            \\
                                                            & \textit{Gay}            & .00    & .40 & .07    & .25 & .09     & .25 & 1.77            & 4            \\
                                                            & \textit{Religious}            & .00    & .42 & .14    & .22 & .22     & .20 & 2.21            & 6            \\
                                                            & \textit{Outdoors}            & .00    & .38 & .44    & .27 & .47     & .27 & 1.00            & 3 \\
                                                            
\end{tblr}
    \begin{tablenotes}
      \scriptsize
      \item \emph{Notes.} Performance ($R^2$) and predictive accuracy (RMSE) of single-truth (\textsc{dlc-cct}) and multi-truth (i\textsc{dlc-cct}) models. No. of cultures and entropy of cluster assignment shown for i\textsc{dlc-cct}.
    \end{tablenotes}
  \end{threeparttable}
\end{table}

The sixth column of Table 1 presents the posterior mode of the number of instantiated cultures --- those with at least one respondent assigned to them. Although this count provides insights into the magnitude of heterogeneity in cultural consensus, it does not convey the distribution's uniformity (or lack thereof) concerning cultural assignments. An entropy-based metric fills this gap by estimating the uncertainty in the assignment of a randomly selected respondent under the modal posterior cultural distribution. Lower entropy occurs when there are a few dominant clusters, while higher entropy occurs when there is a more balanced distribution of cultures. The posterior over models parameters for the i\textsc{dlc-cct} featured an average cultural entropy of 1.39 and an average cultural count of 4.82 instantiated cultures (i.e., those with at least one respondent assigned to them in the modal posterior assignment), demonstrating both a multiplicity of consensus as well as a non-uniform distribution of assigned respondents per culture. Crucially, as highlighted in the sixth column of Table 1, despite varying in the extent of heterogeneity, none of the 40 datasets was best fit by the single-culture model --- all displayed heterogeneity in consensus beliefs.

\begin{figure}[htp]
    \centering

    \includegraphics[width=2.5cm]{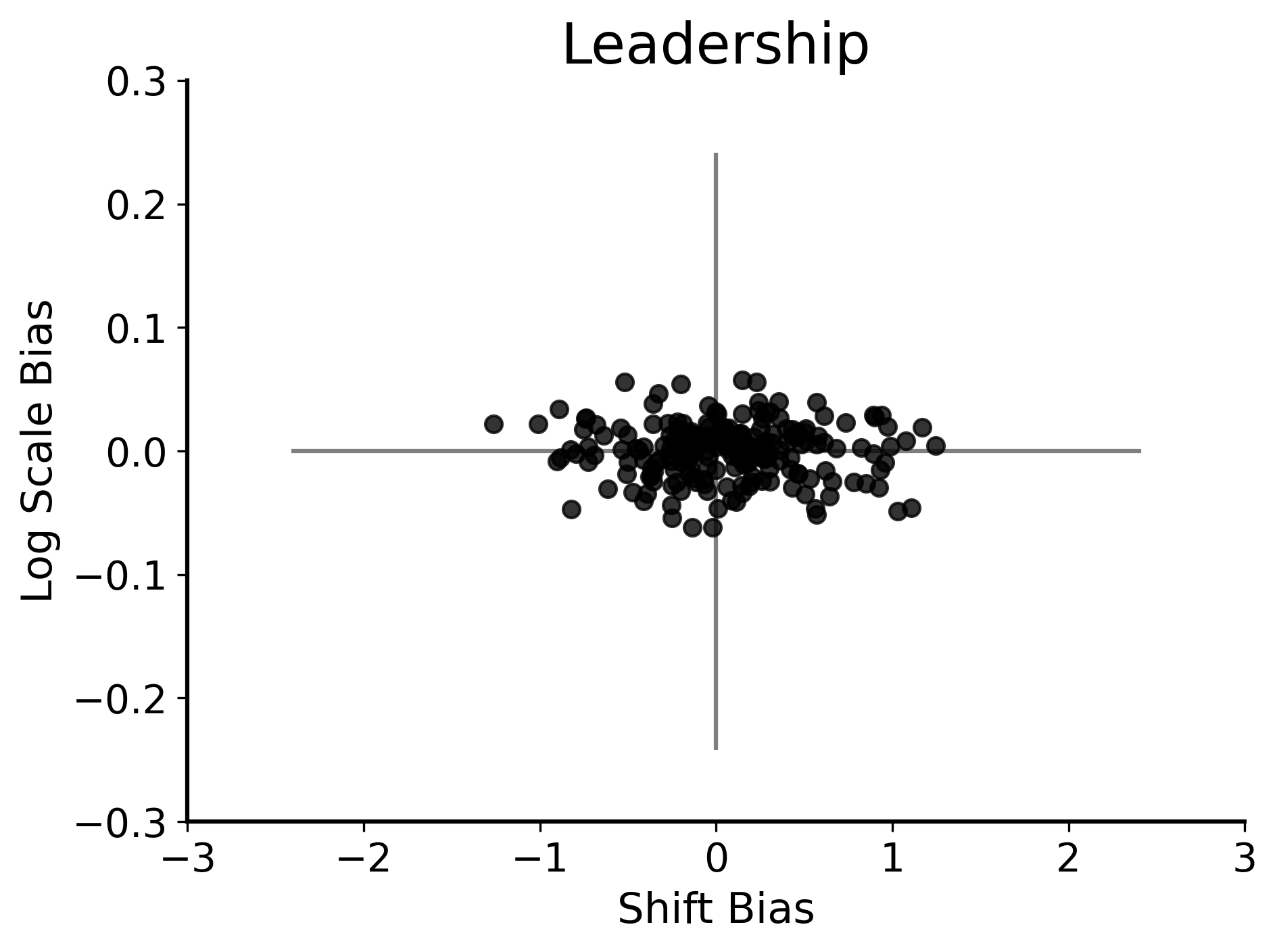}\hspace{-0.17cm}
    \includegraphics[width=2.5cm]{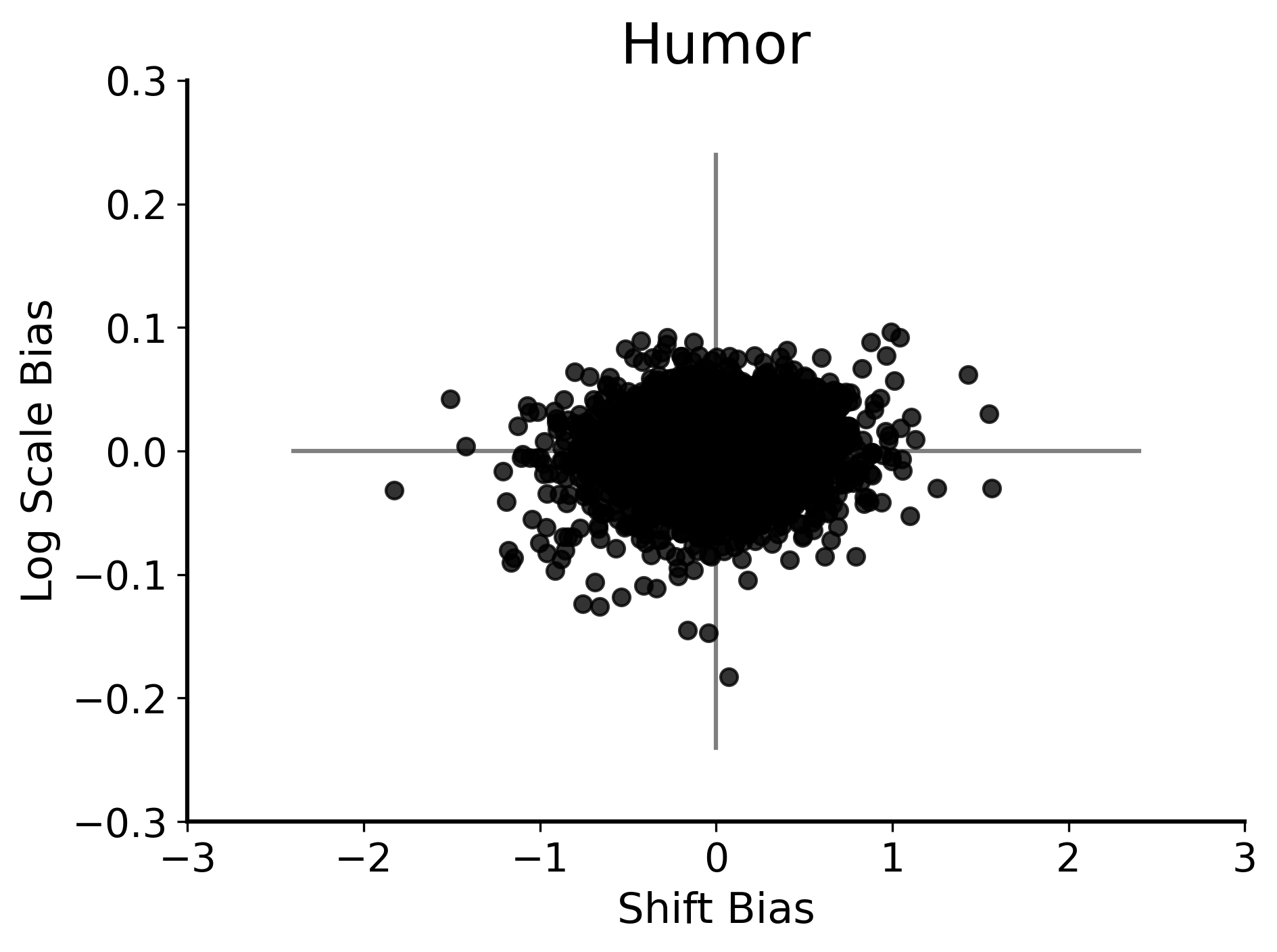}\hspace{-0.17cm}
    \includegraphics[width=2.5cm]{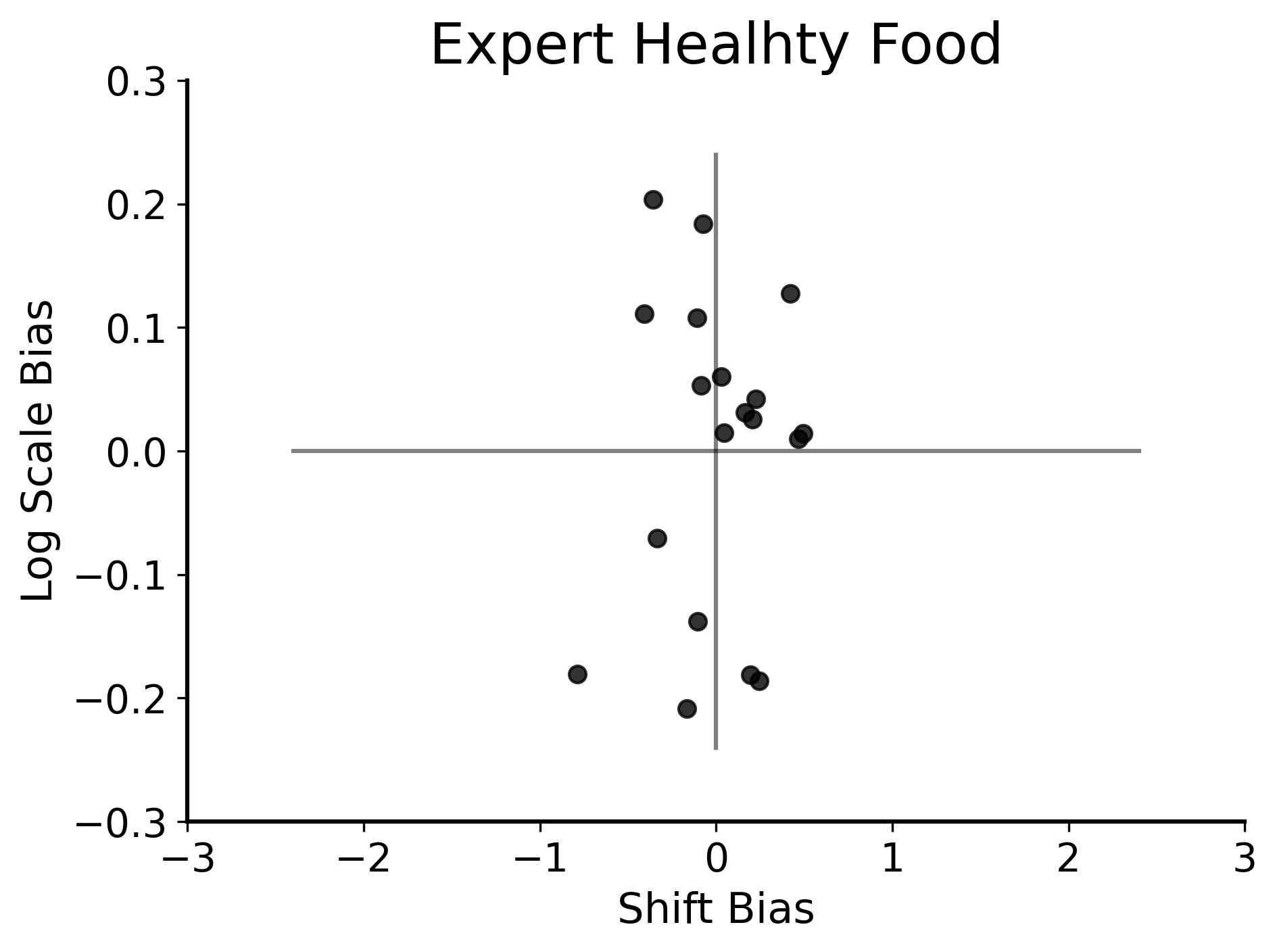}\hspace{-0.17cm}
    \includegraphics[width=2.5cm]{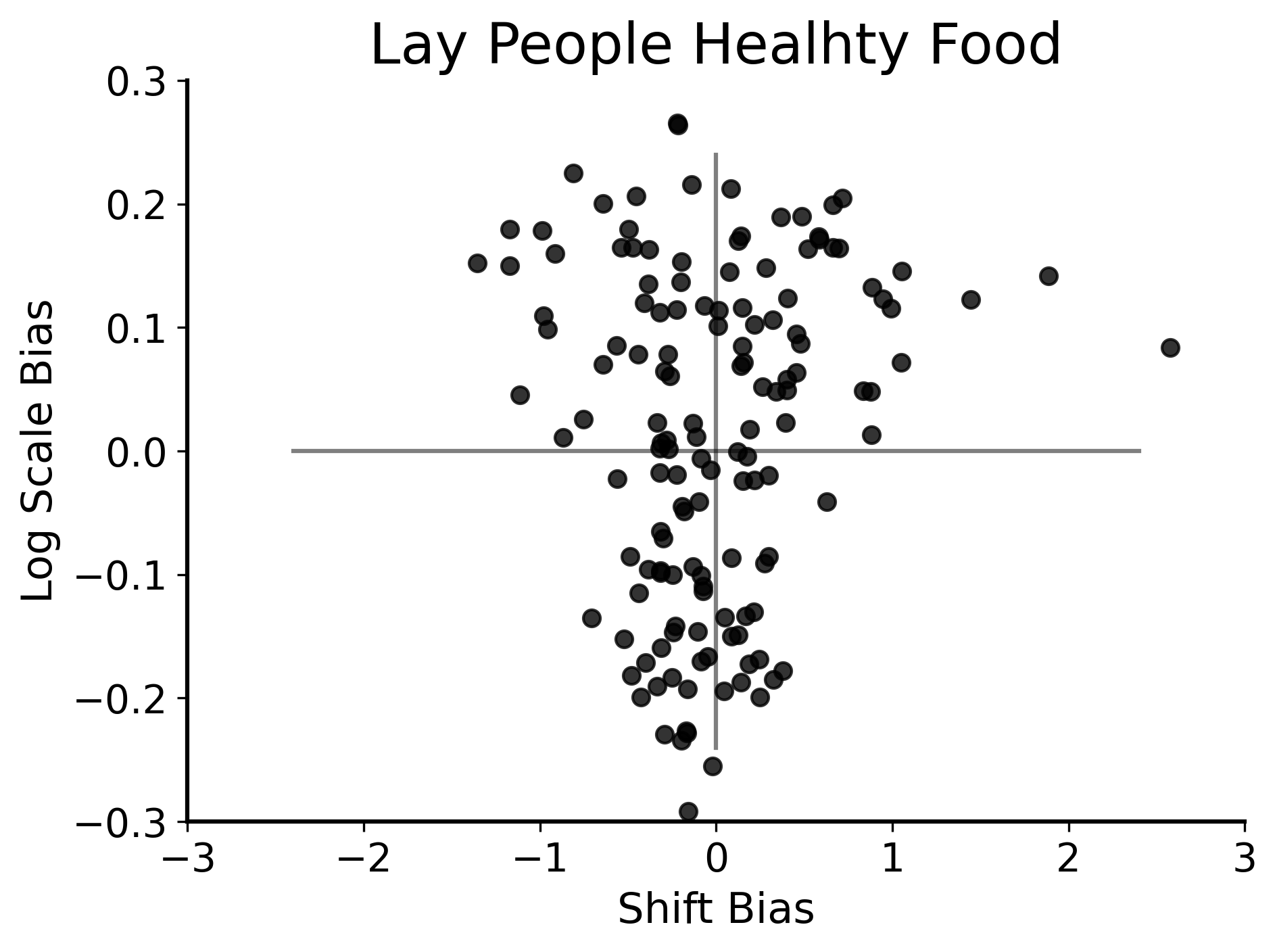}\hspace{-0.17cm}
    \includegraphics[width=2.5cm]{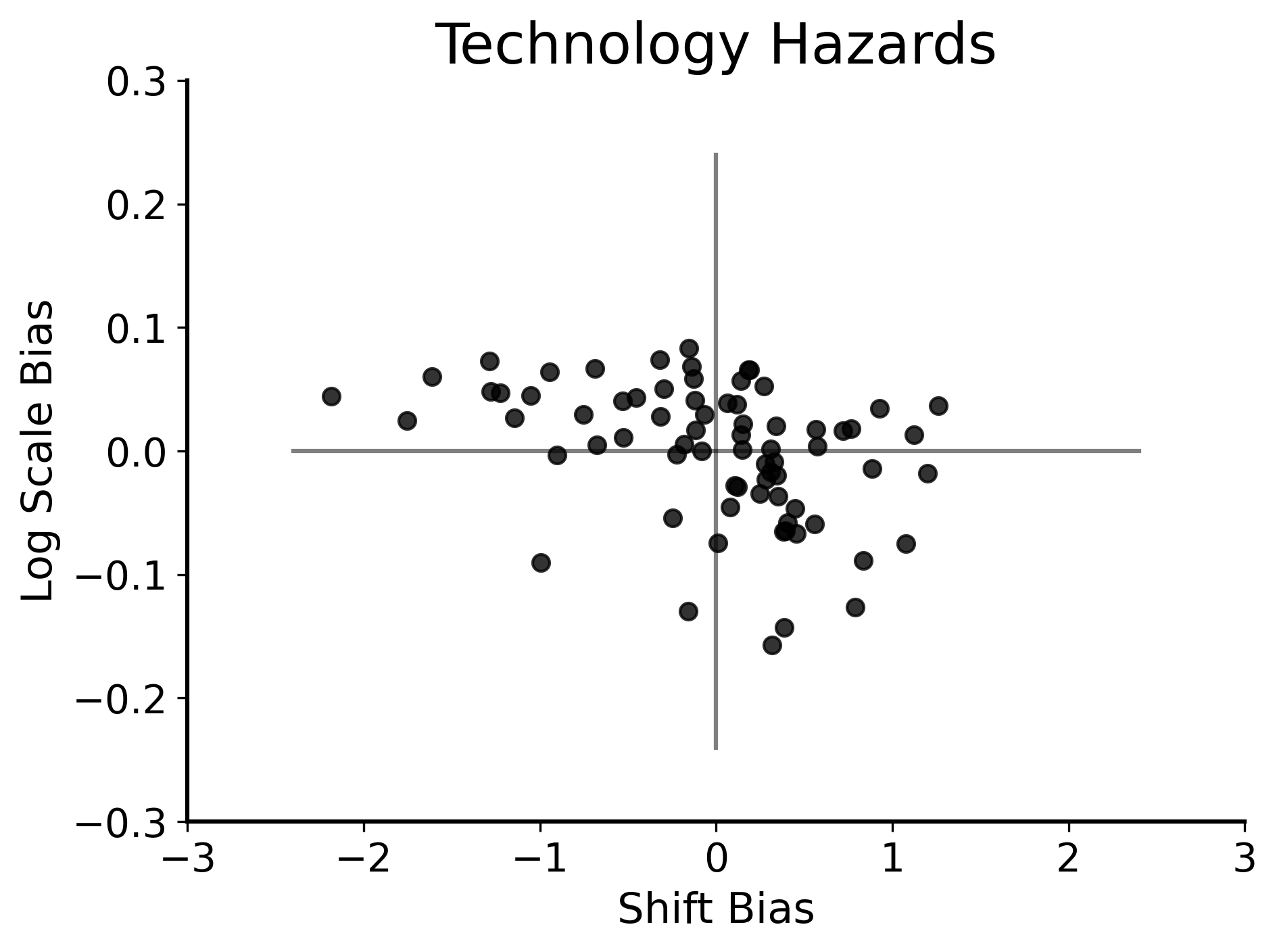}\hspace{-0.17cm}
    \\
    \includegraphics[width=2.5cm]{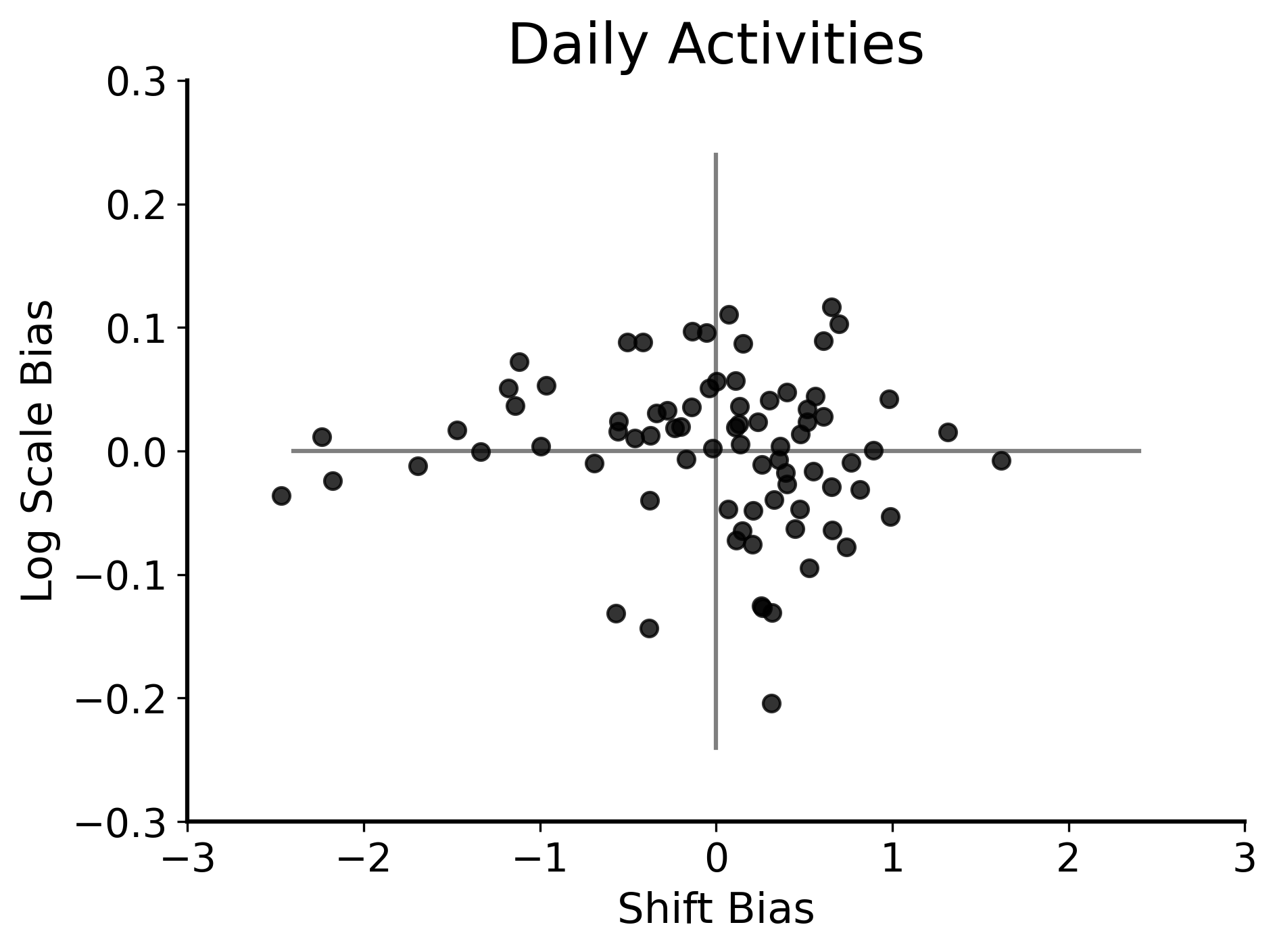}\hspace{-0.17cm}
    \includegraphics[width=2.5cm]{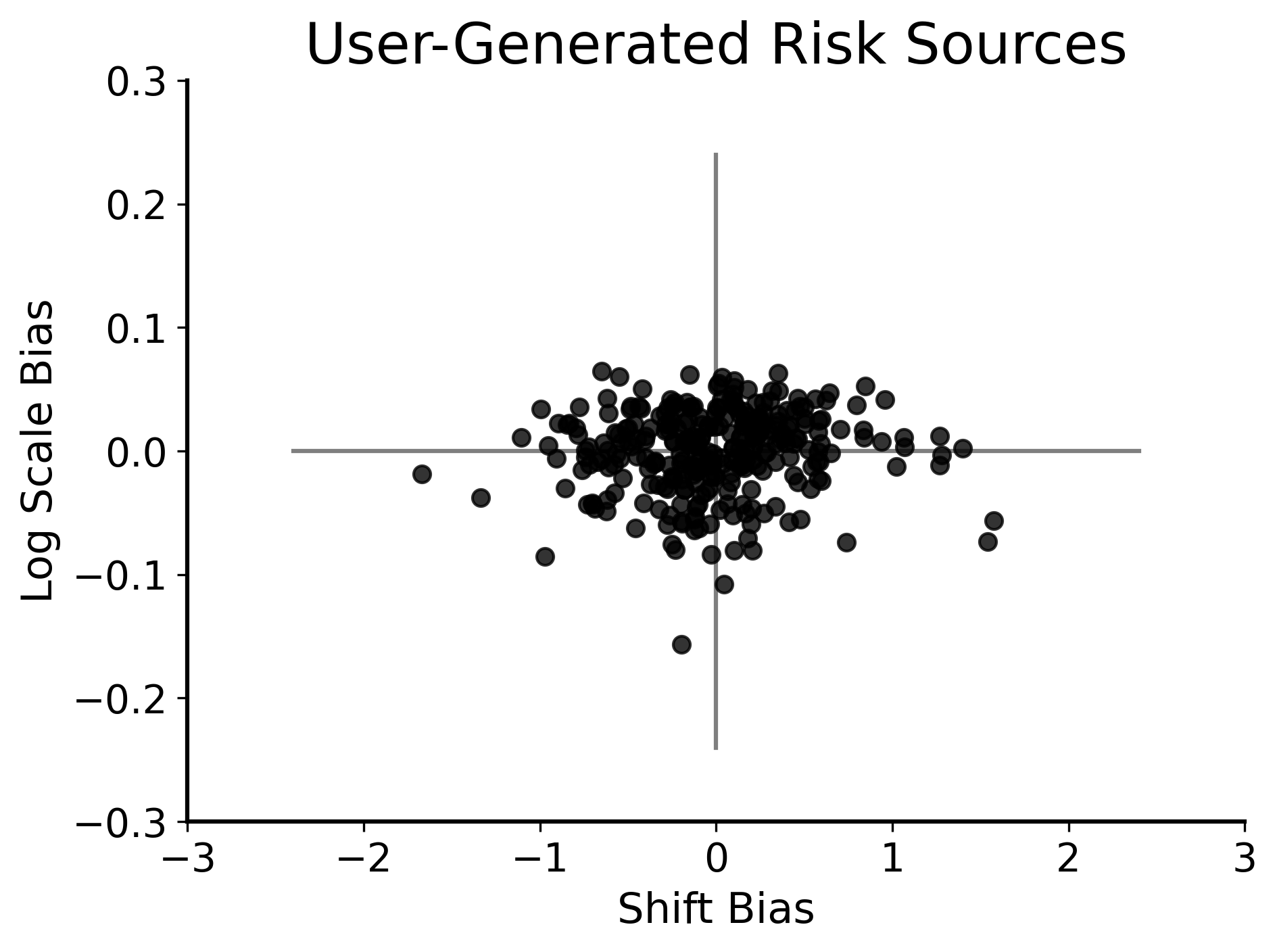}\hspace{-0.17cm}
    \includegraphics[width=2.5cm]{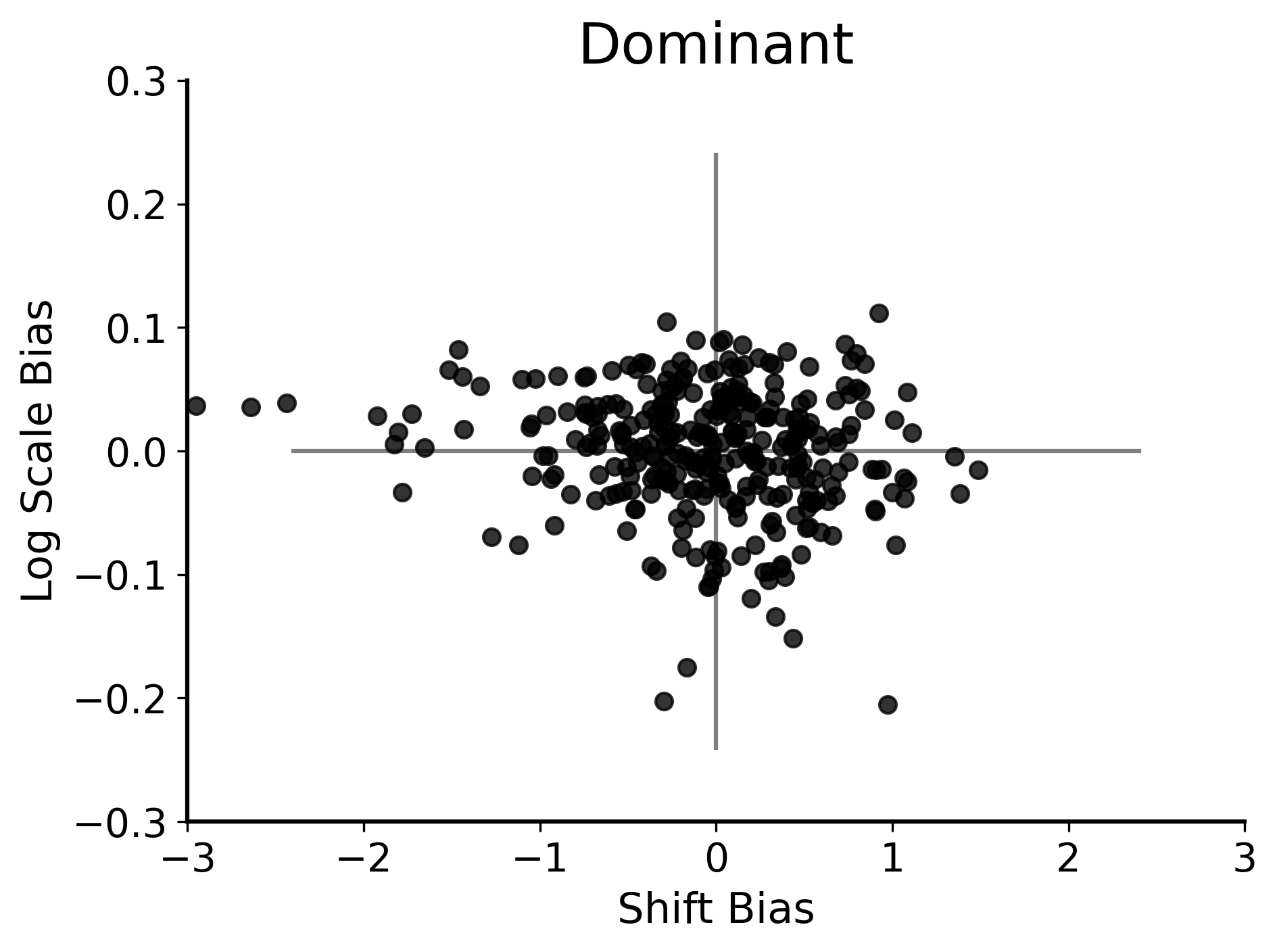}\hspace{-0.17cm}
    \includegraphics[width=2.5cm]{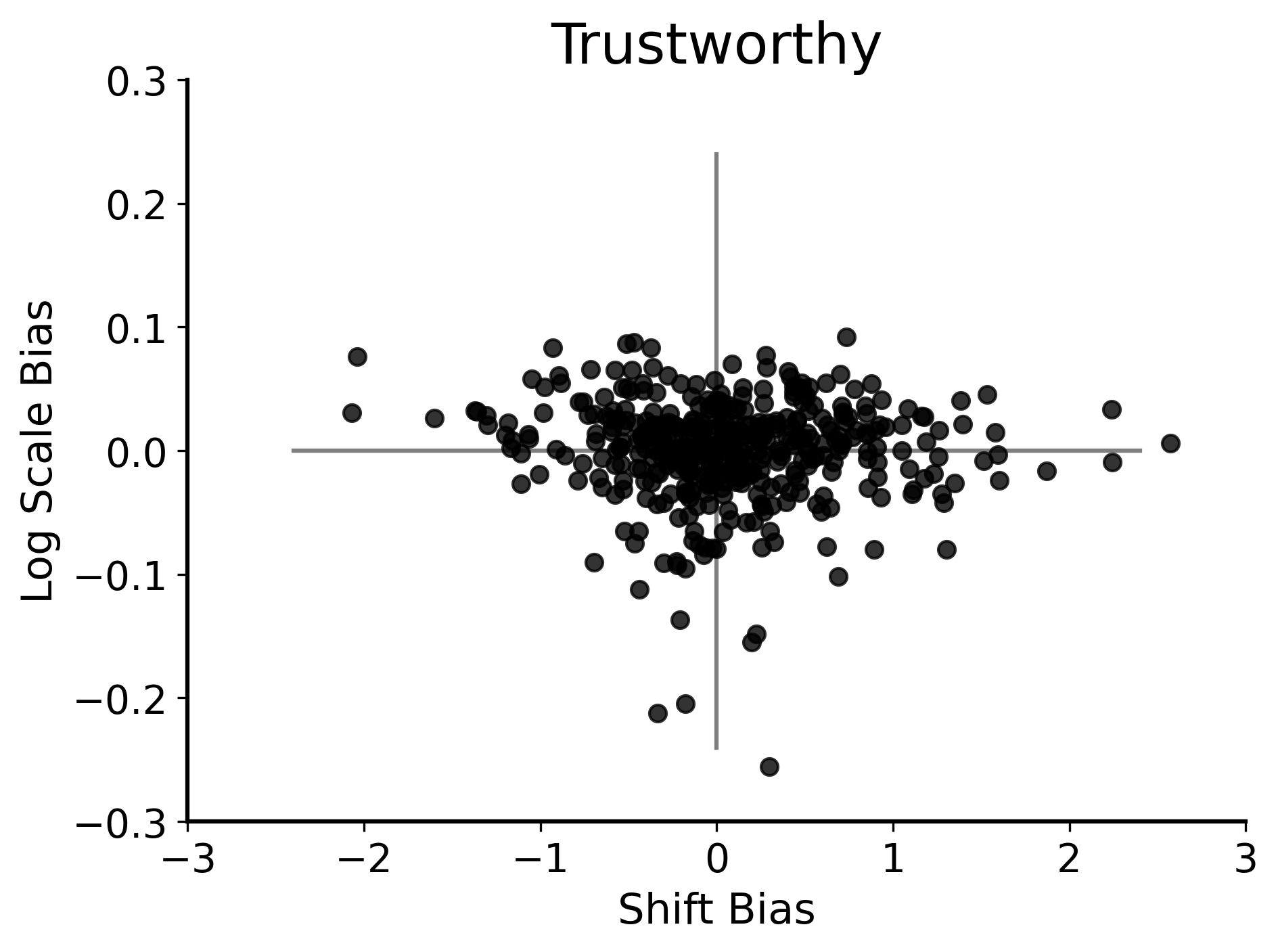}\hspace{-0.17cm}
    \includegraphics[width=2.5cm]{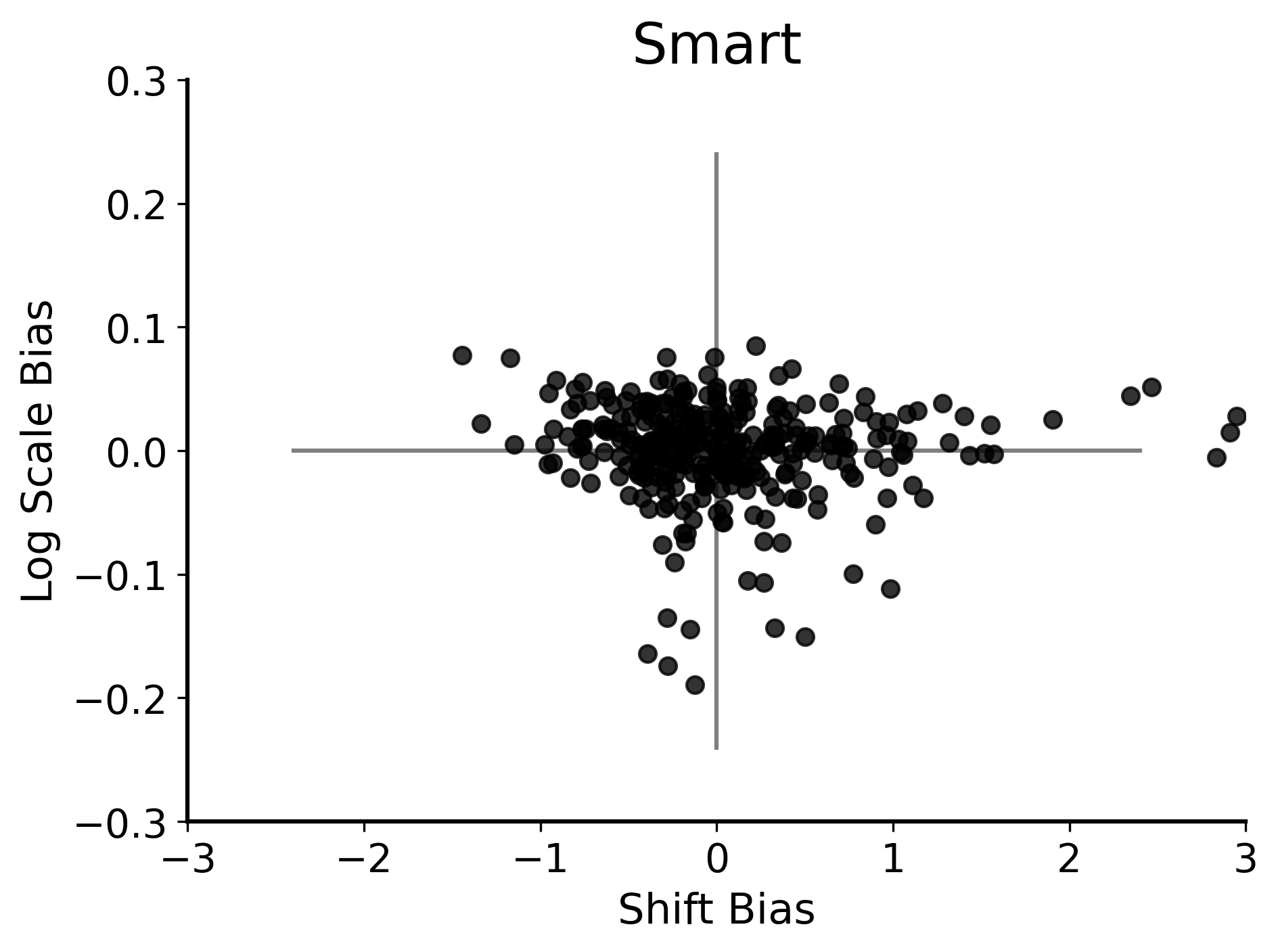}\hspace{-0.17cm}
    \\
    \includegraphics[width=2.5cm]{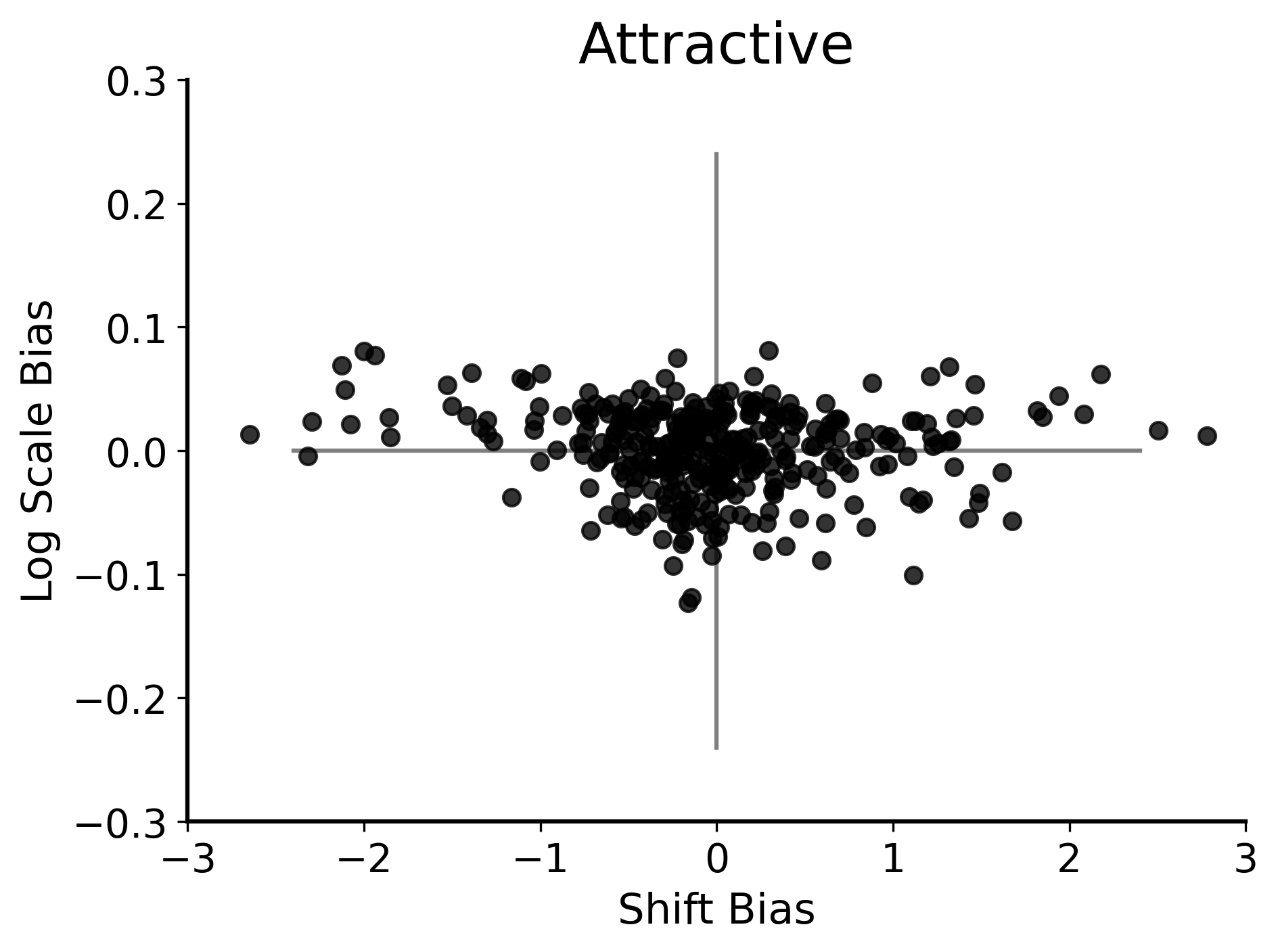}\hspace{-0.17cm}
    \includegraphics[width=2.5cm]{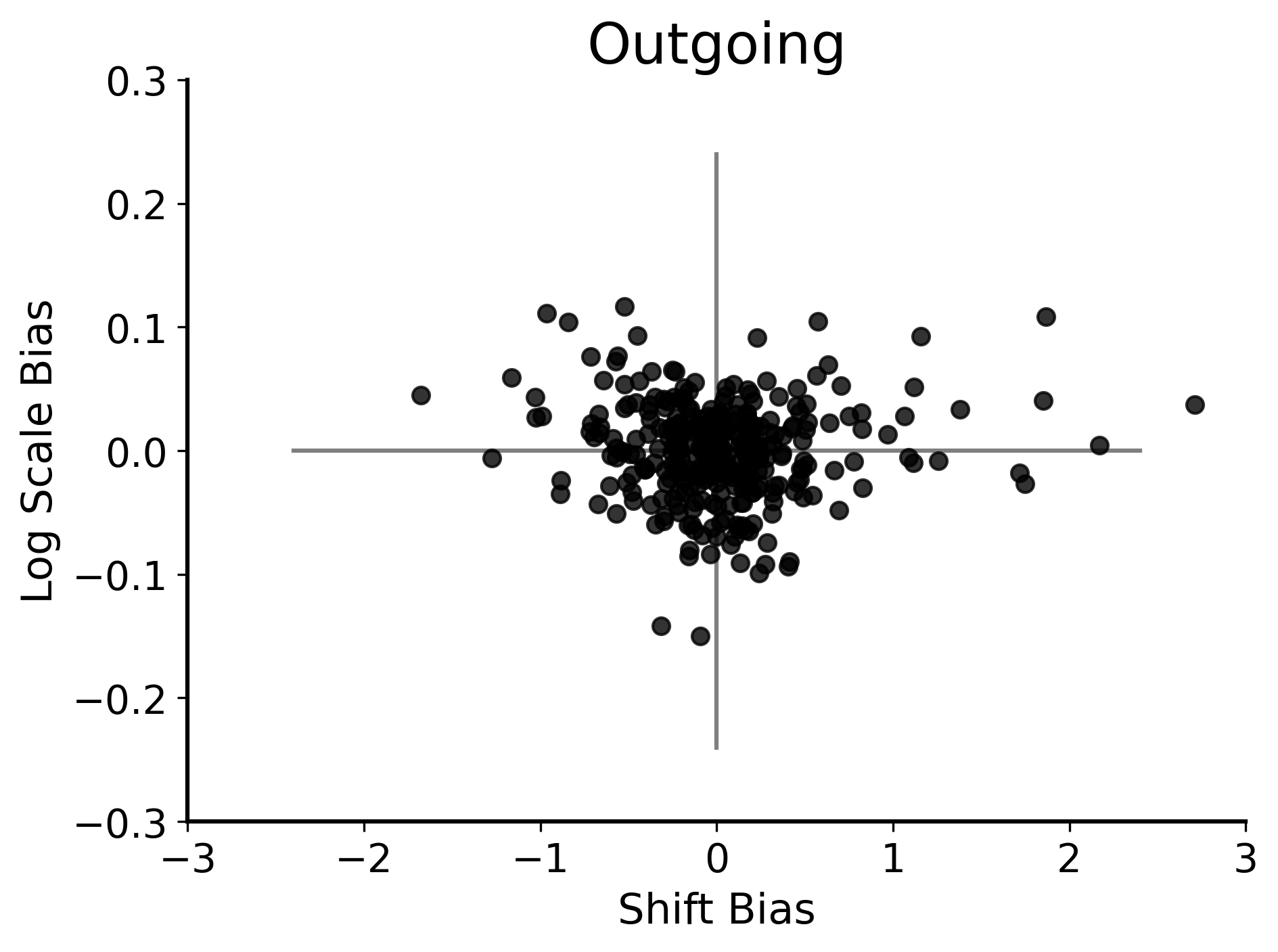}\hspace{-0.17cm}
    \includegraphics[width=2.5cm]{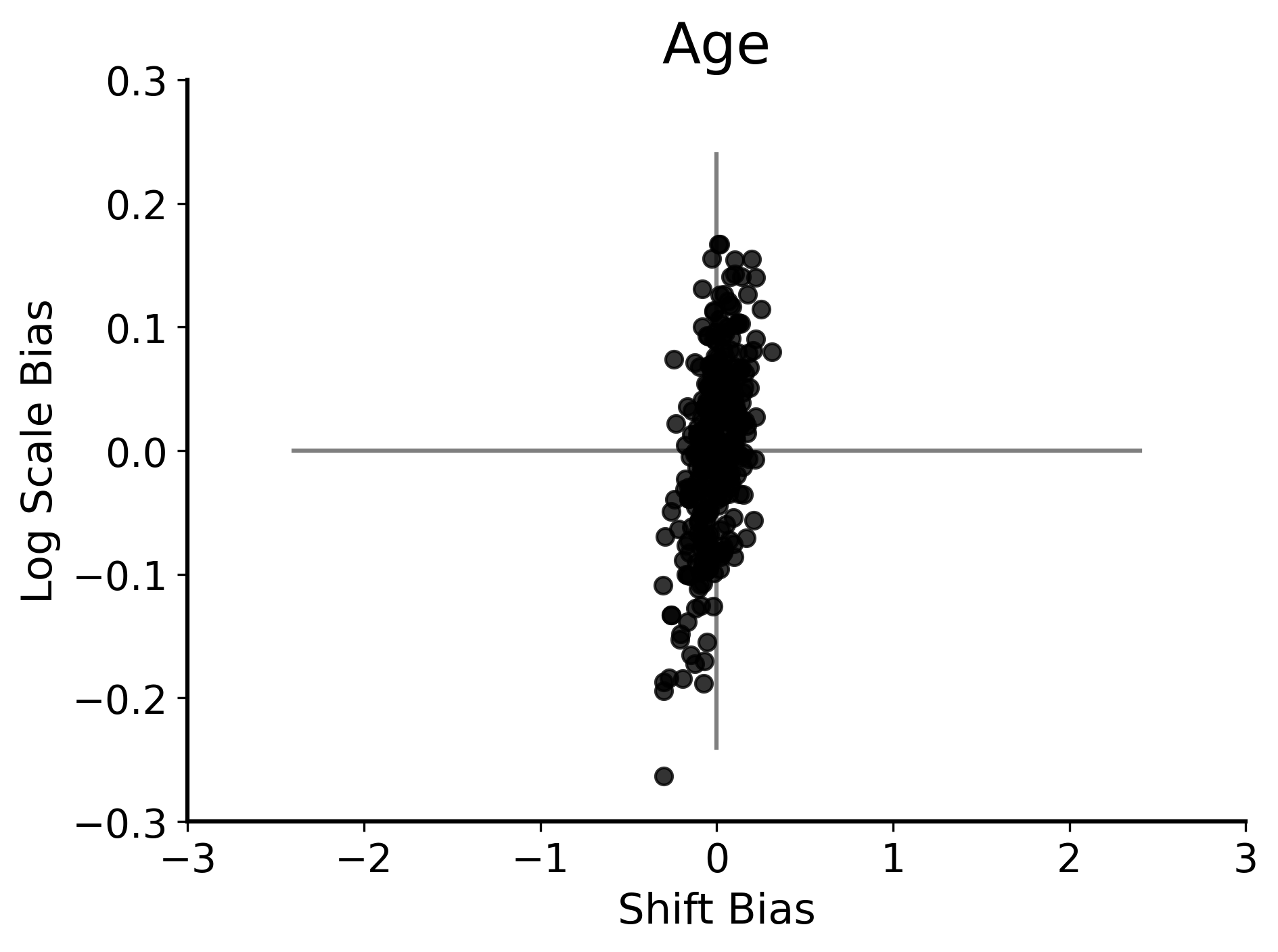}\hspace{-0.17cm}
    \includegraphics[width=2.5cm]{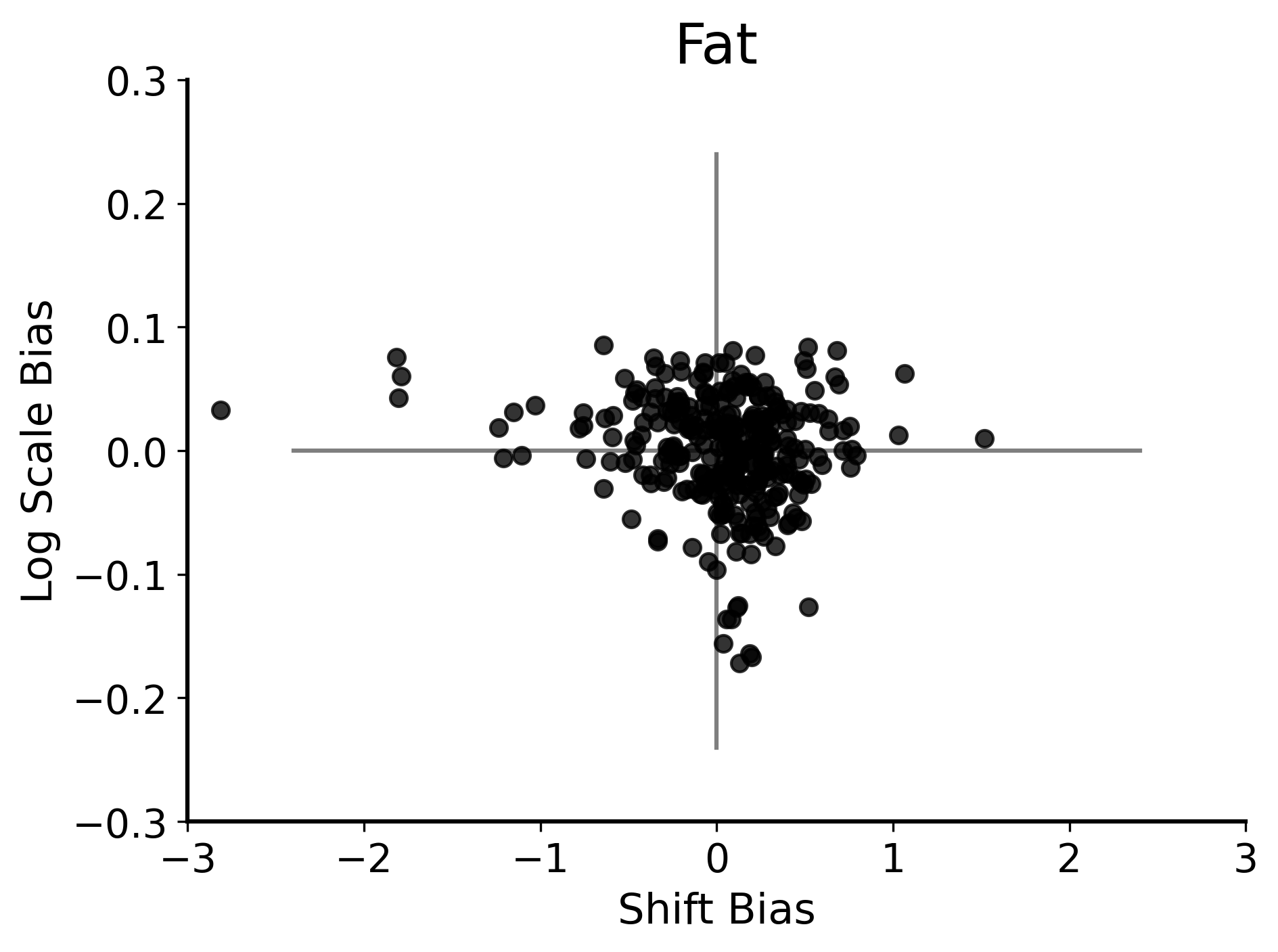}\hspace{-0.17cm}
    \includegraphics[width=2.5cm]{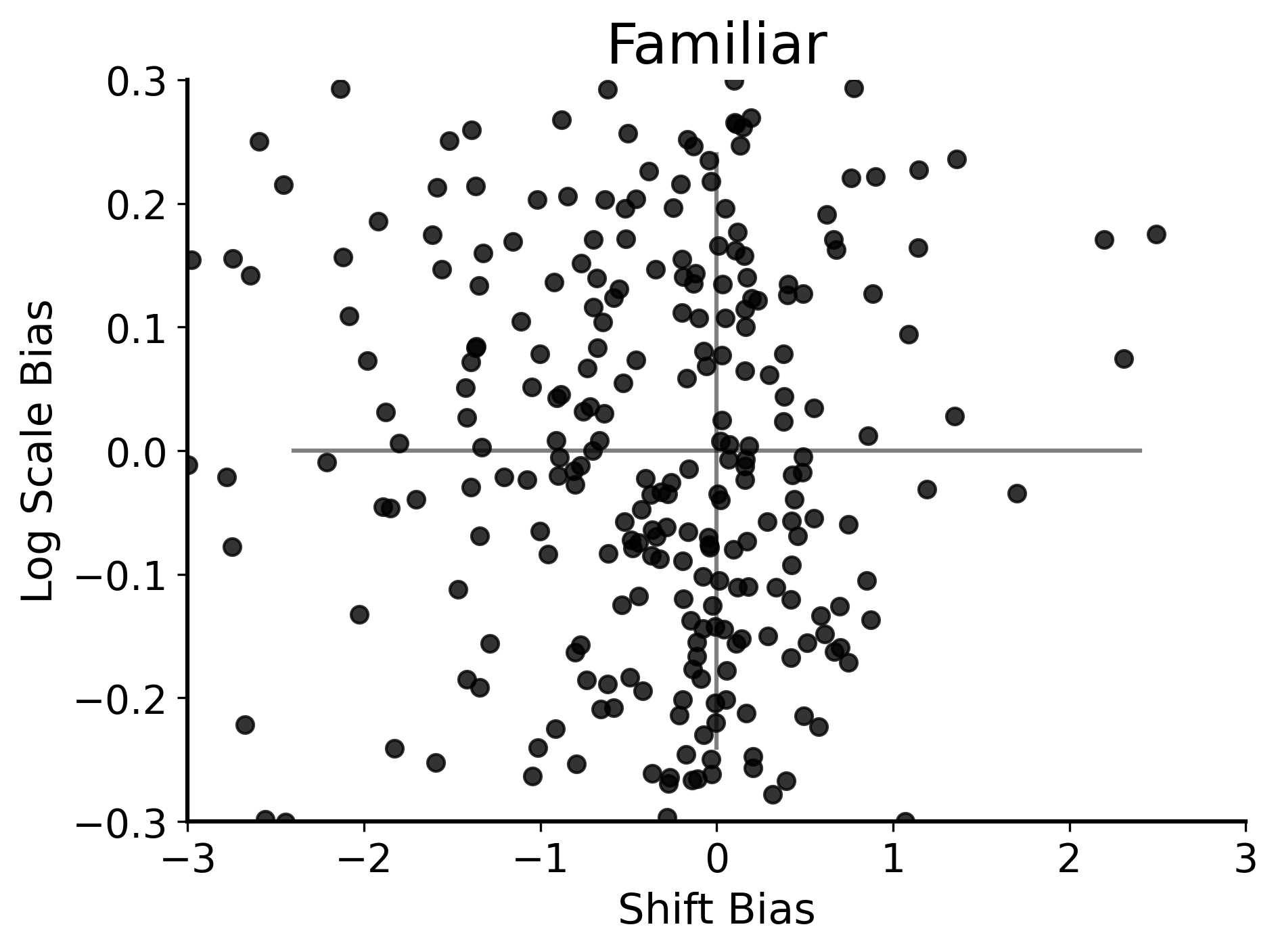}\hspace{-0.17cm}
    \\
    \includegraphics[width=2.5cm]{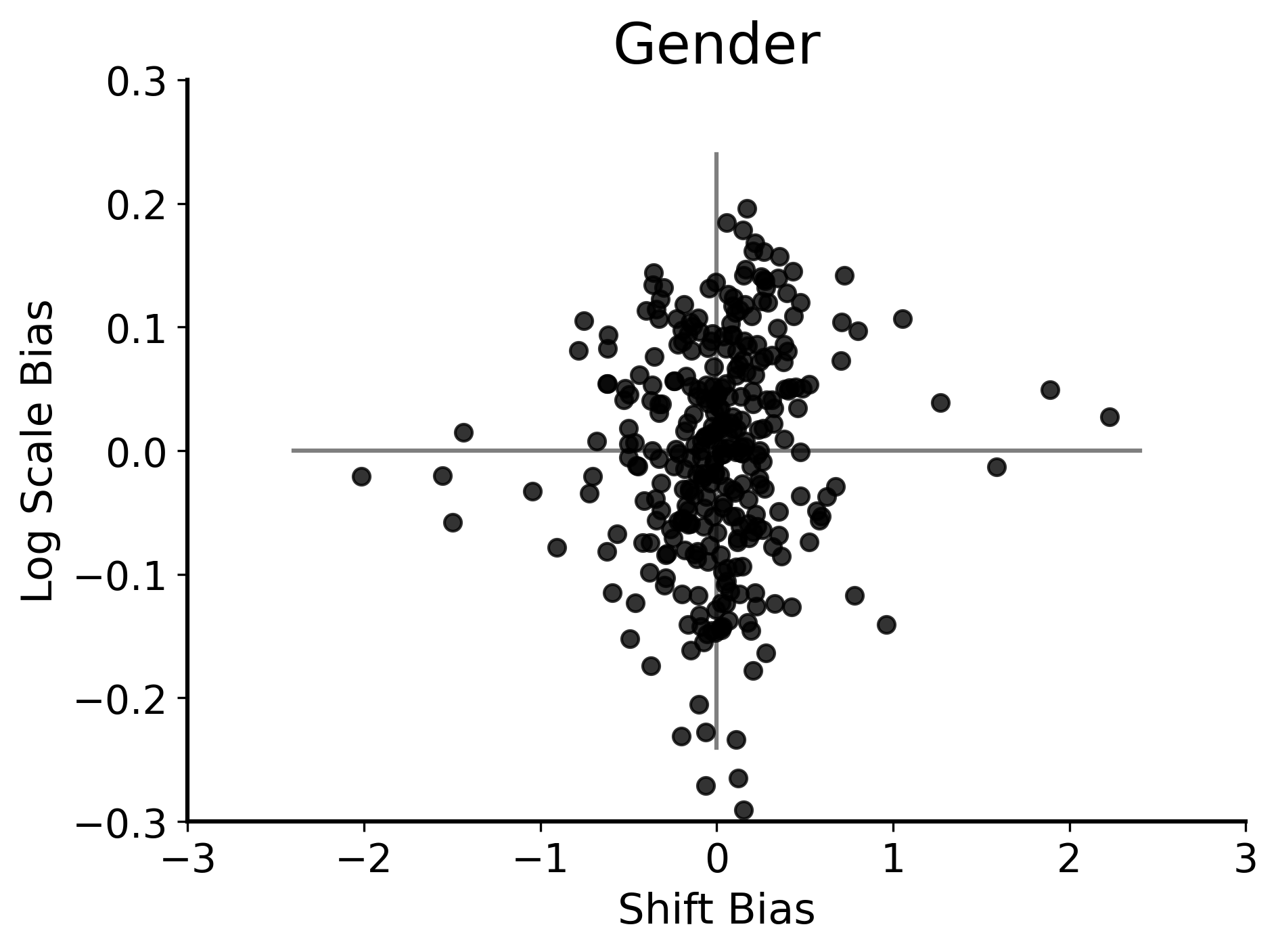}\hspace{-0.17cm}
    \includegraphics[width=2.5cm]{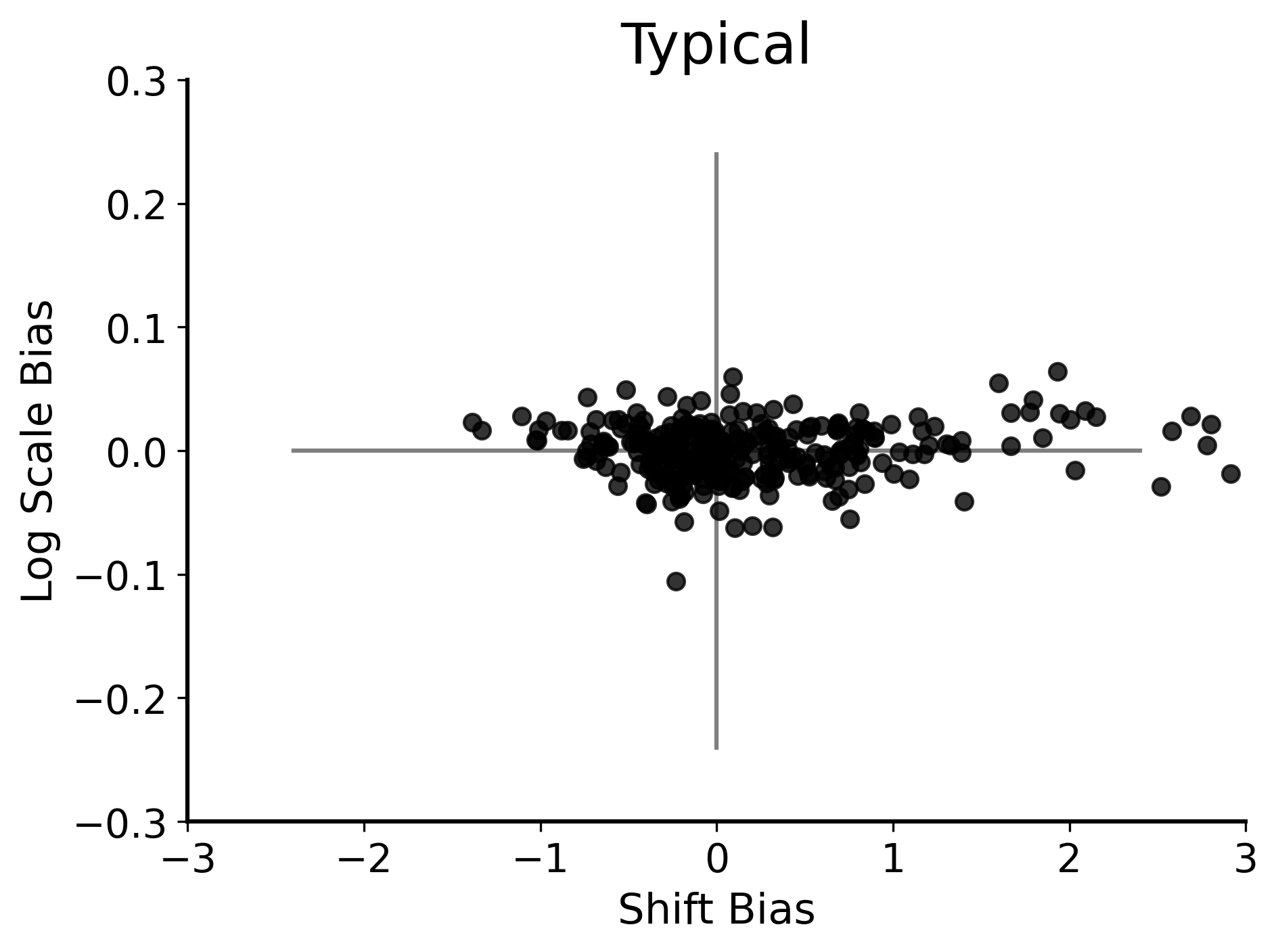}\hspace{-0.17cm}
    \includegraphics[width=2.5cm]{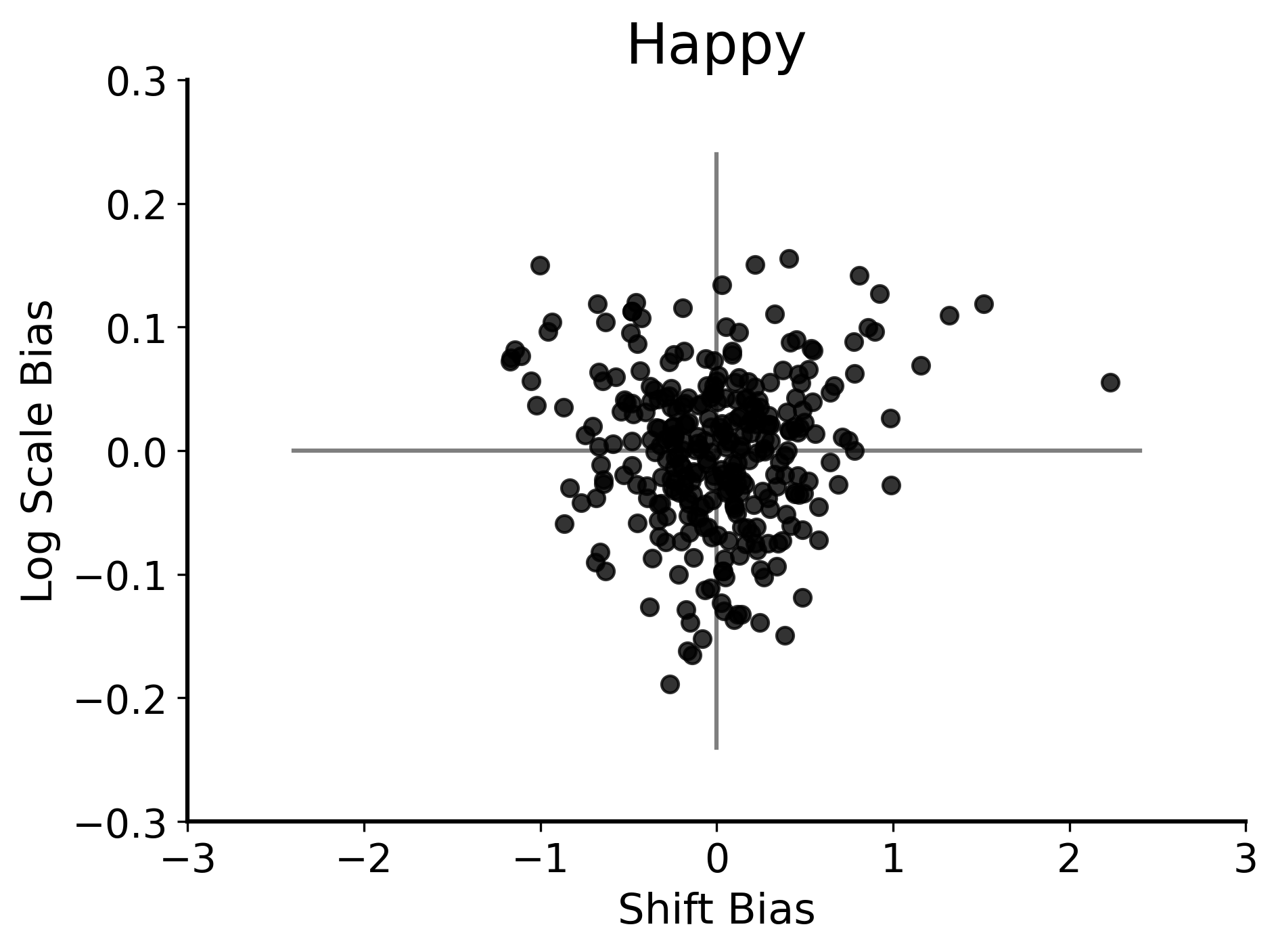}\hspace{-0.17cm}
    \includegraphics[width=2.5cm]{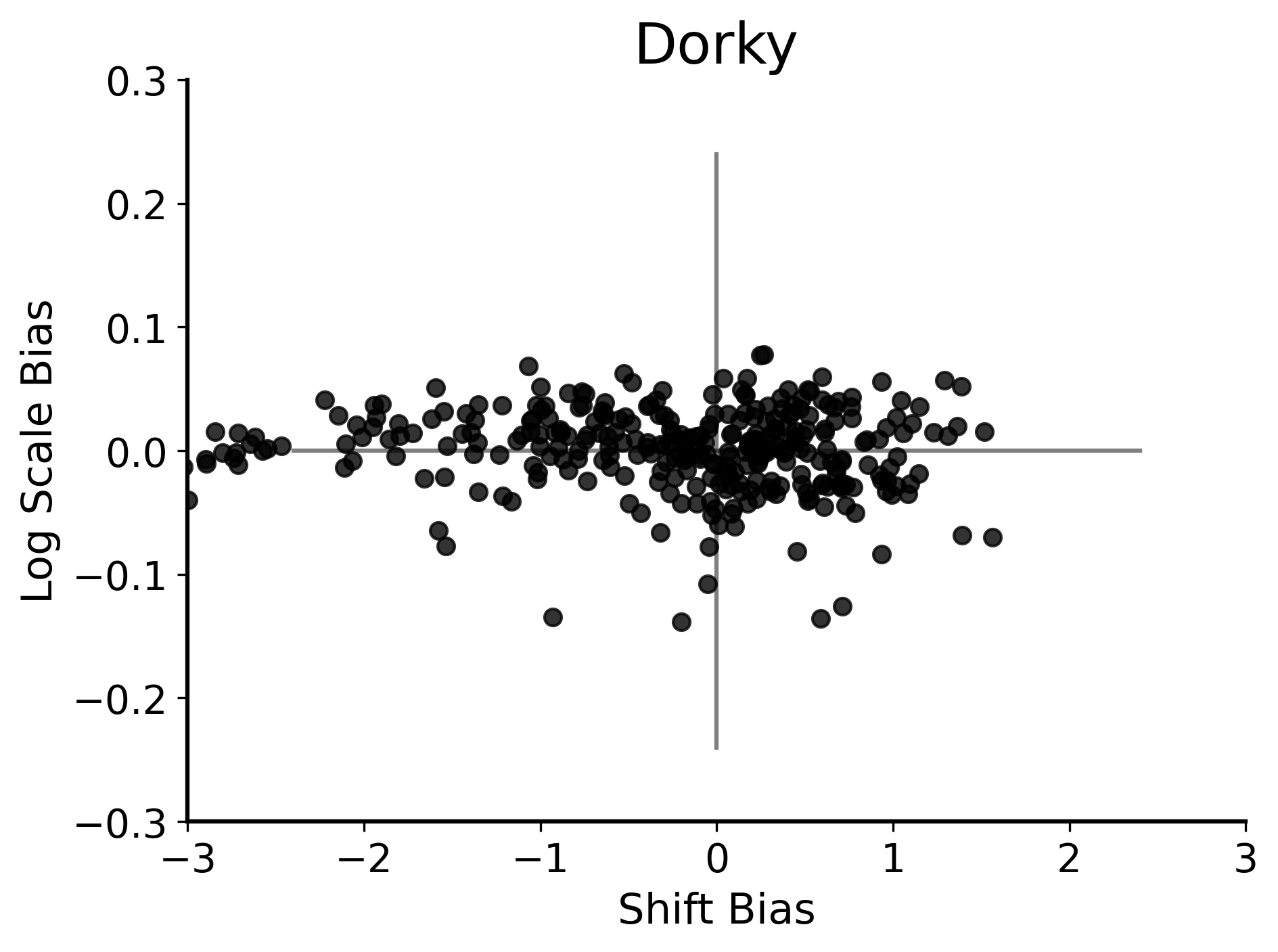}\hspace{-0.17cm}
    \includegraphics[width=2.5cm]{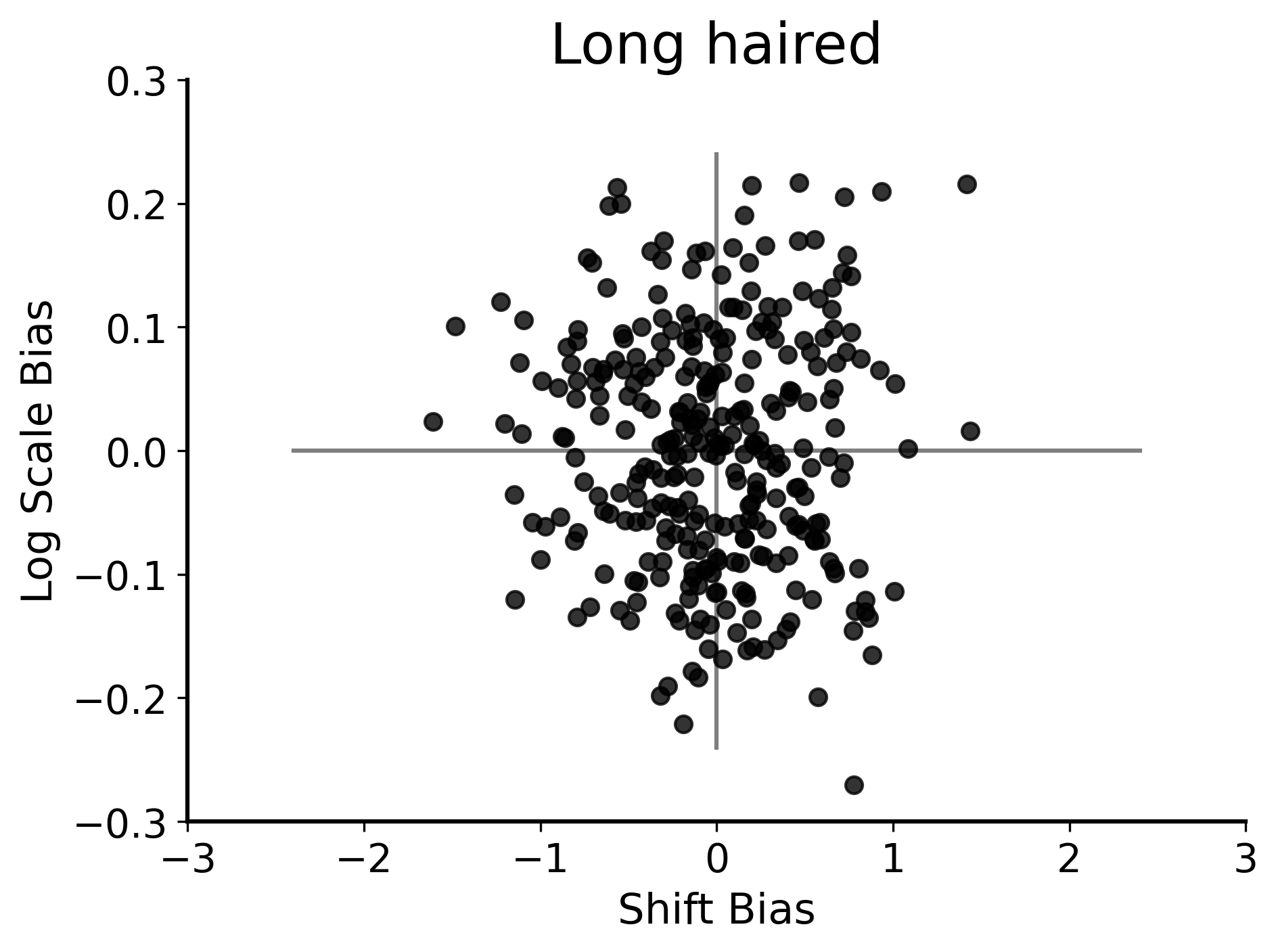}\hspace{-0.17cm}
    \\
    \includegraphics[width=2.5cm]{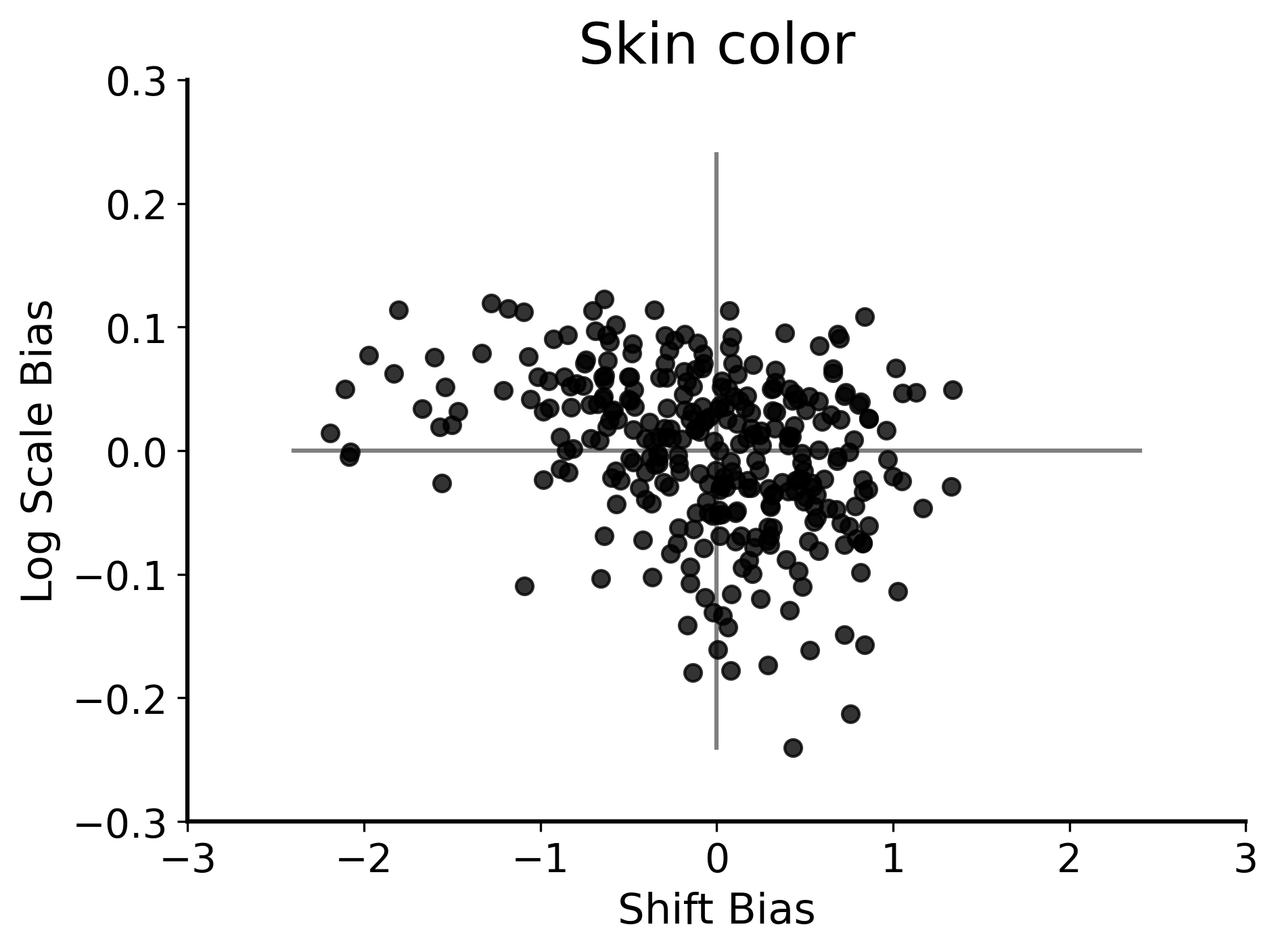}\hspace{-0.17cm}
    \includegraphics[width=2.5cm]{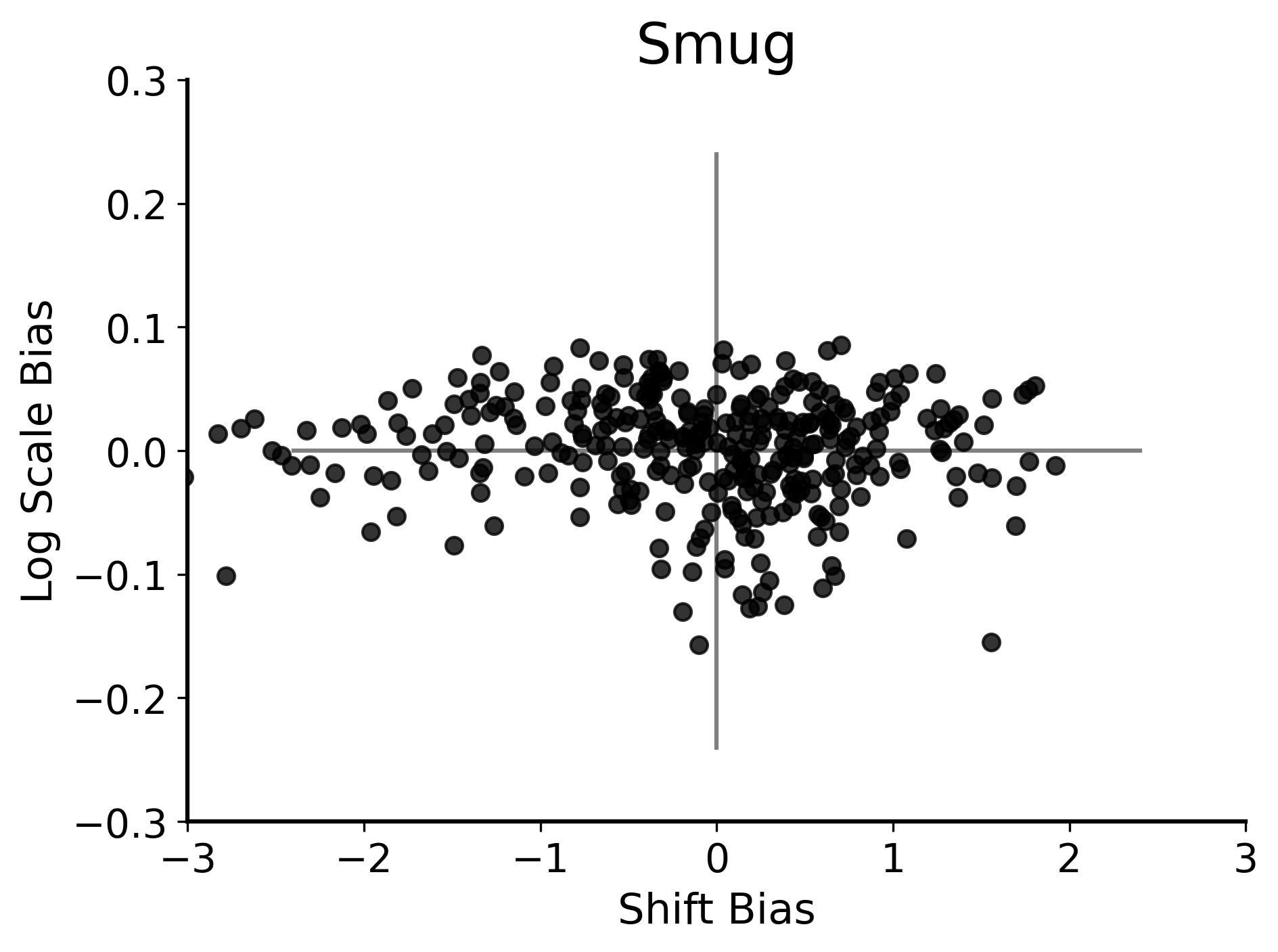}\hspace{-0.17cm}
    \includegraphics[width=2.5cm]{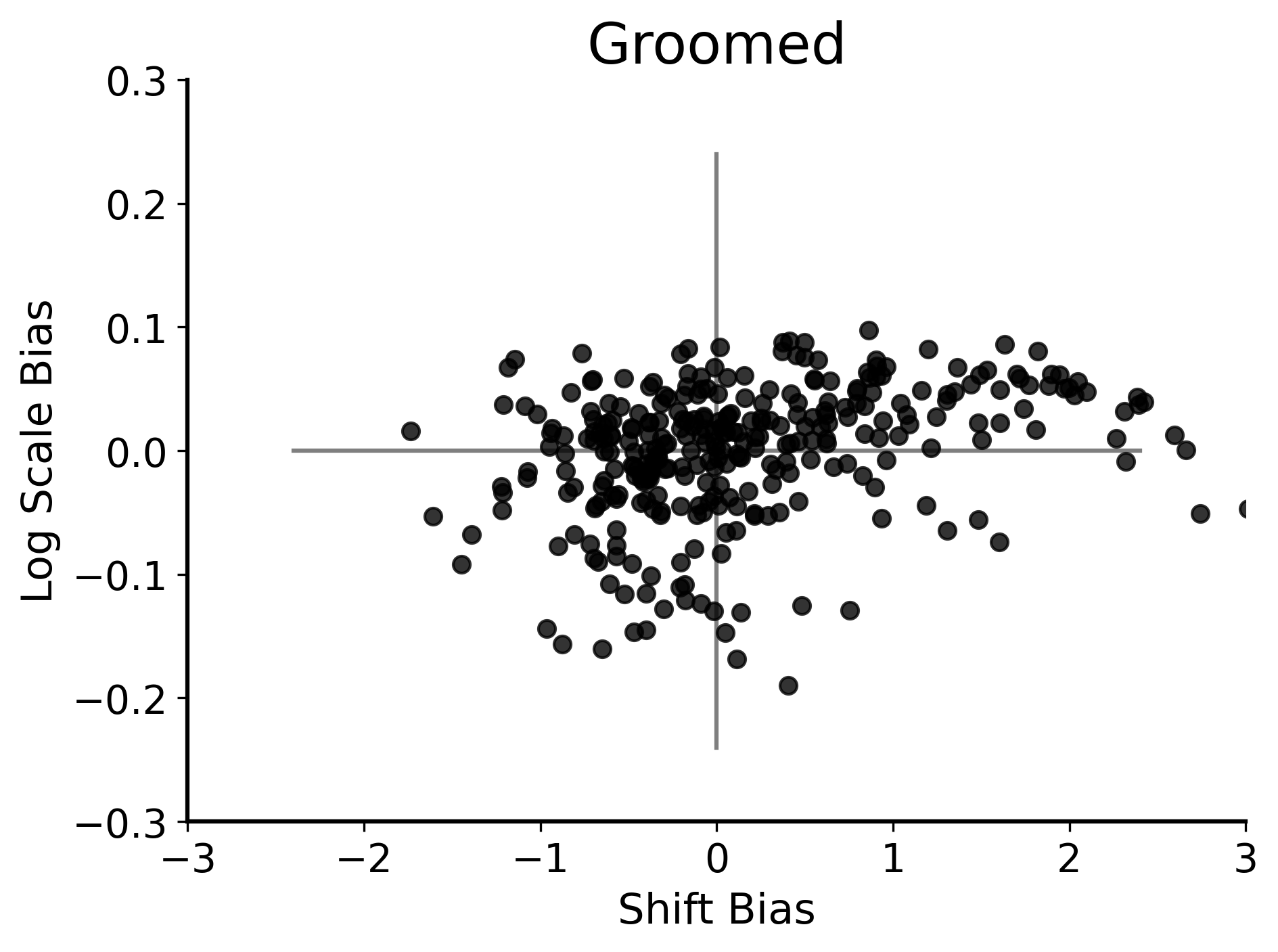}\hspace{-0.17cm}
    \includegraphics[width=2.5cm]{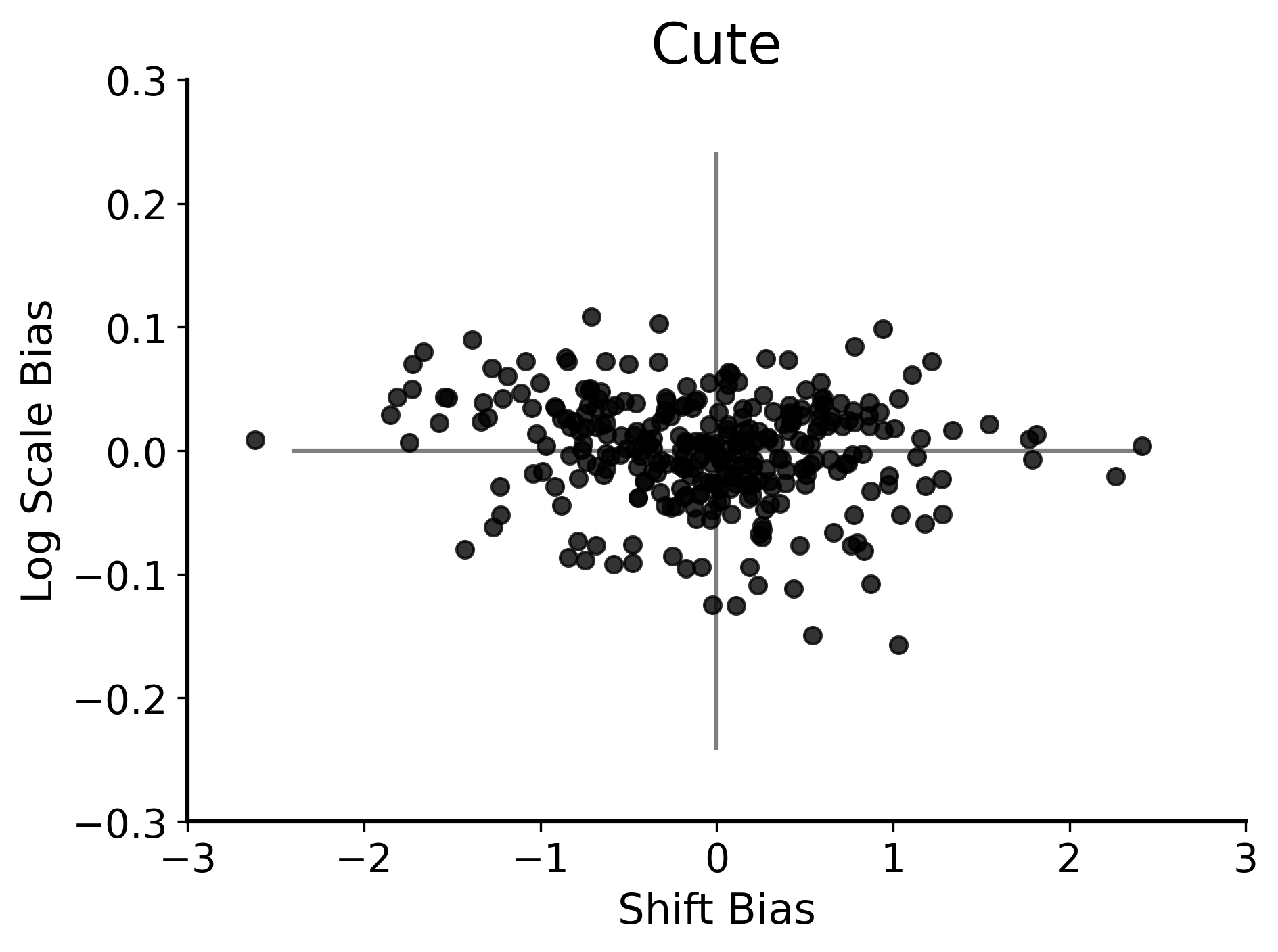}\hspace{-0.17cm}
    \includegraphics[width=2.5cm]{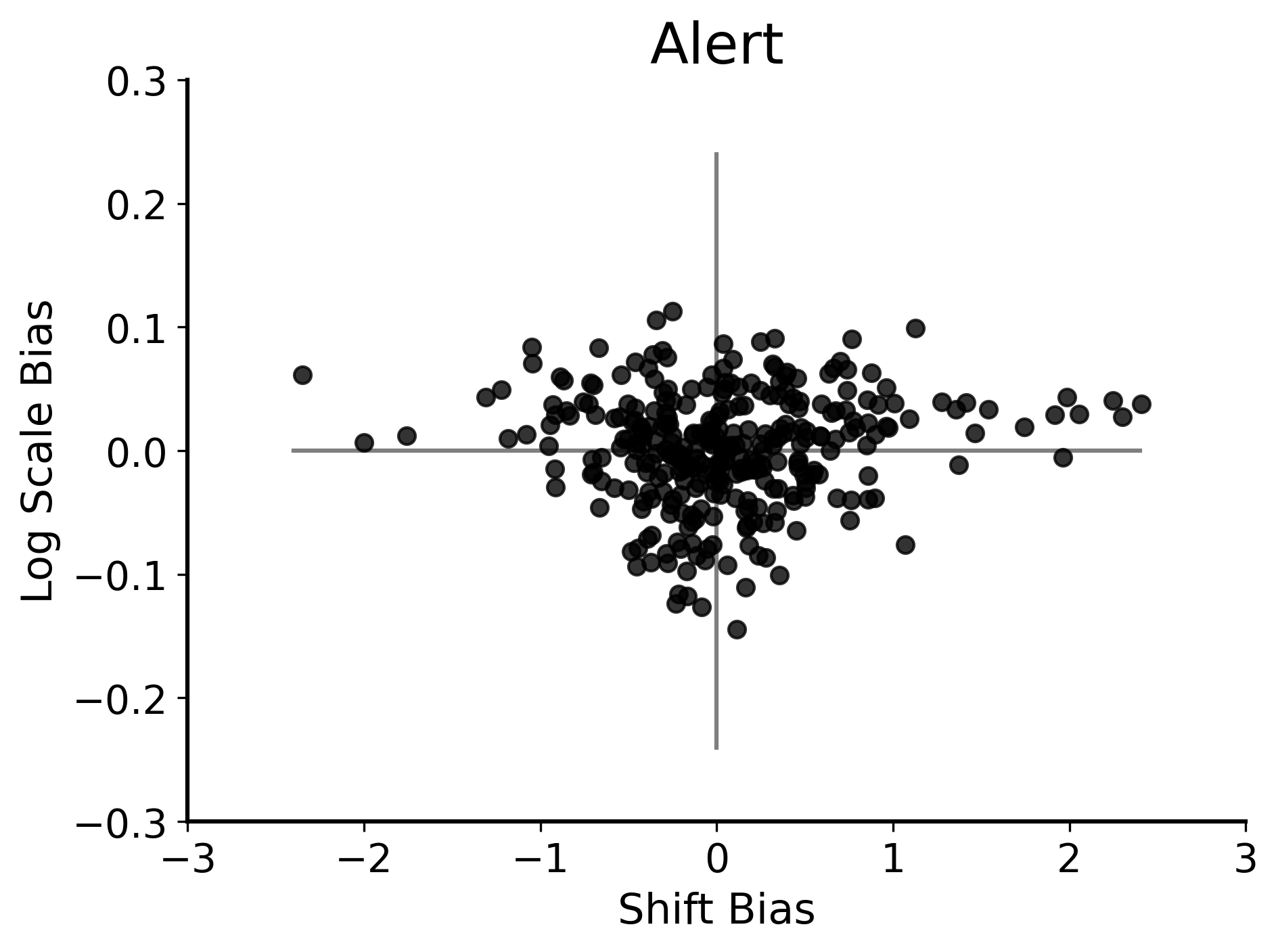}\hspace{-0.17cm}
    \\
    \includegraphics[width=2.5cm]{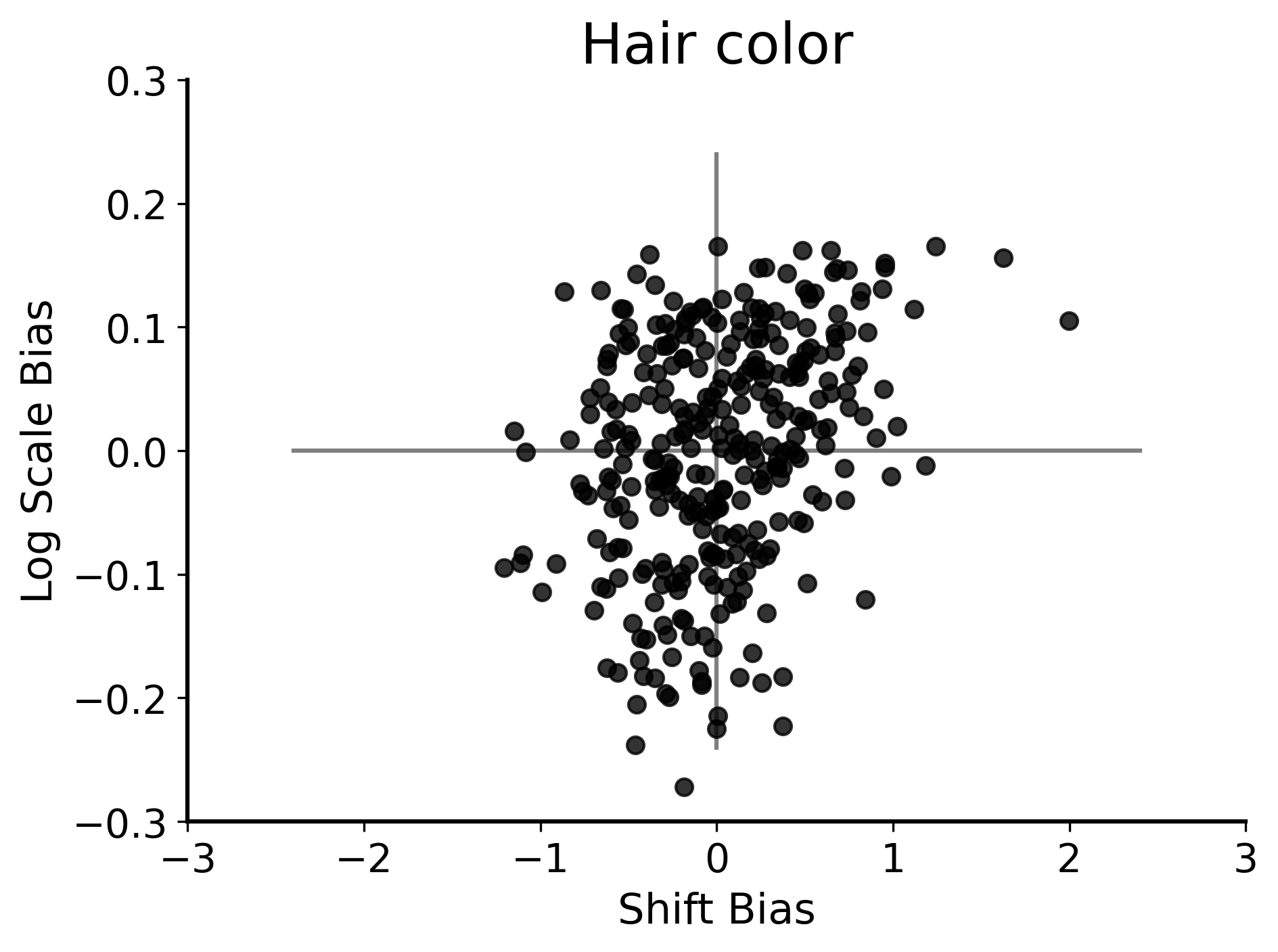}\hspace{-0.17cm}
    \includegraphics[width=2.5cm]{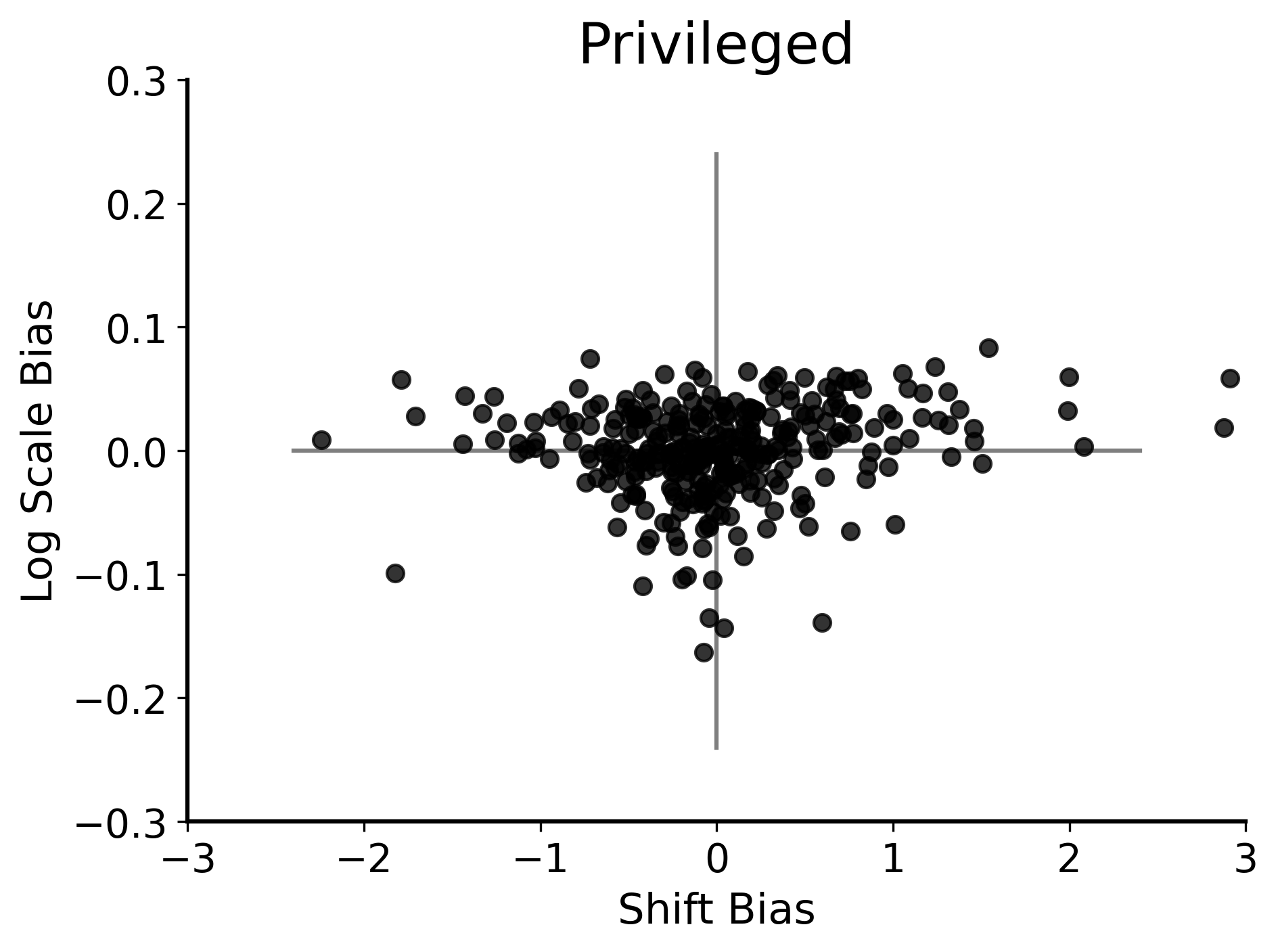}\hspace{-0.17cm}
    \includegraphics[width=2.5cm]{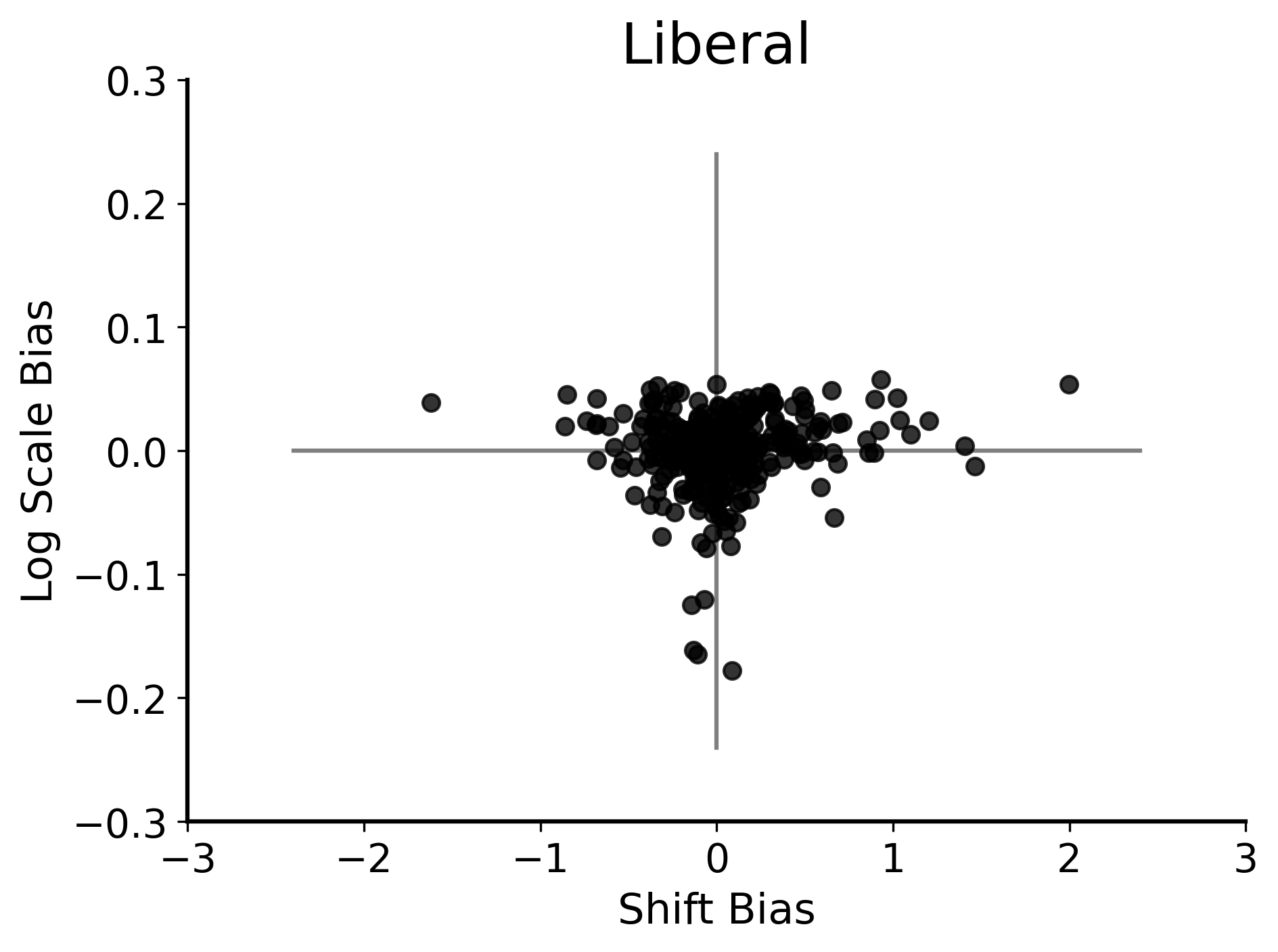}\hspace{-0.17cm}
    \includegraphics[width=2.5cm]{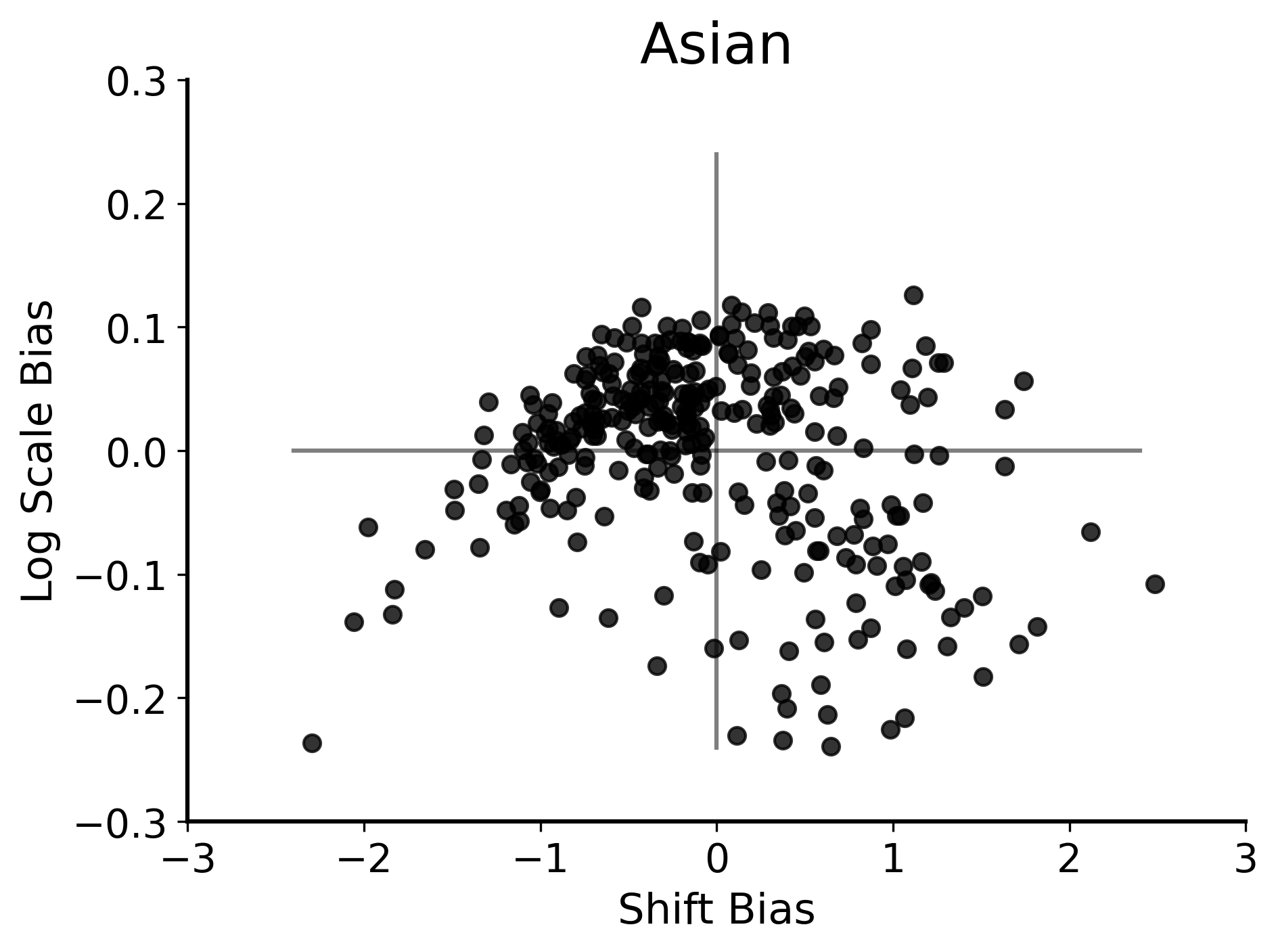}\hspace{-0.17cm}
    \includegraphics[width=2.5cm]{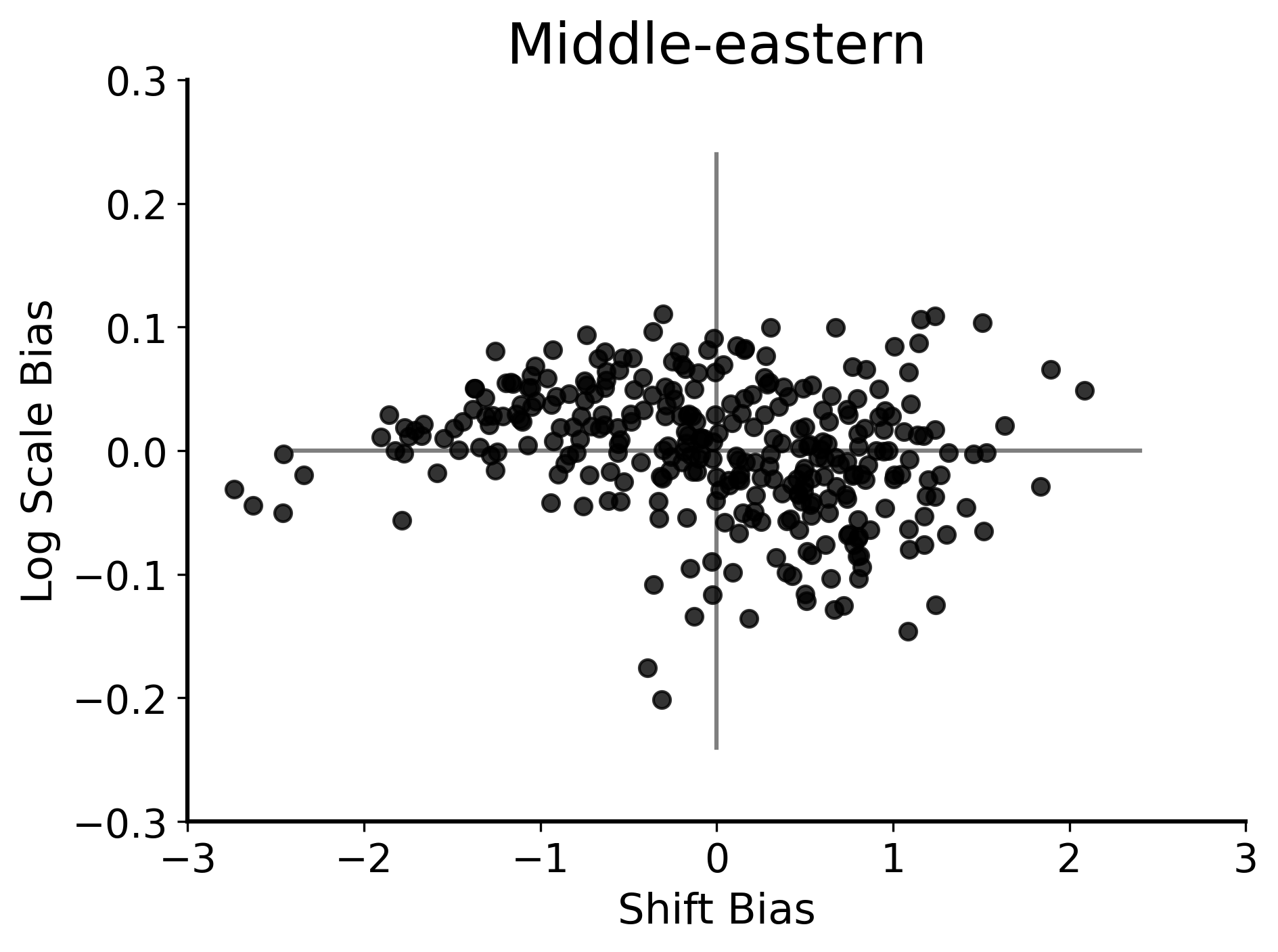}\hspace{-0.17cm}
    \\
    \includegraphics[width=2.5cm]{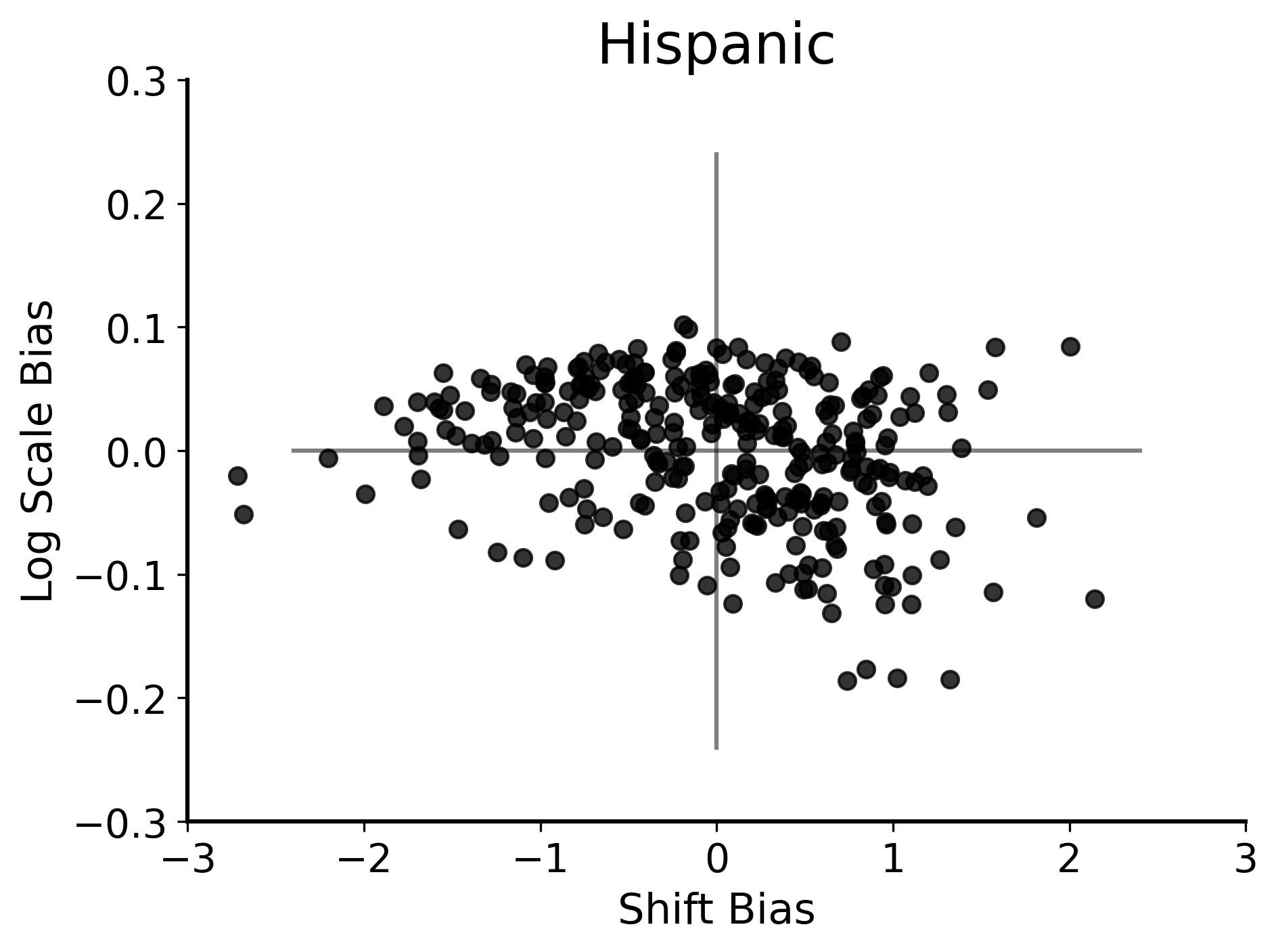}\hspace{-0.17cm}
    \includegraphics[width=2.5cm]{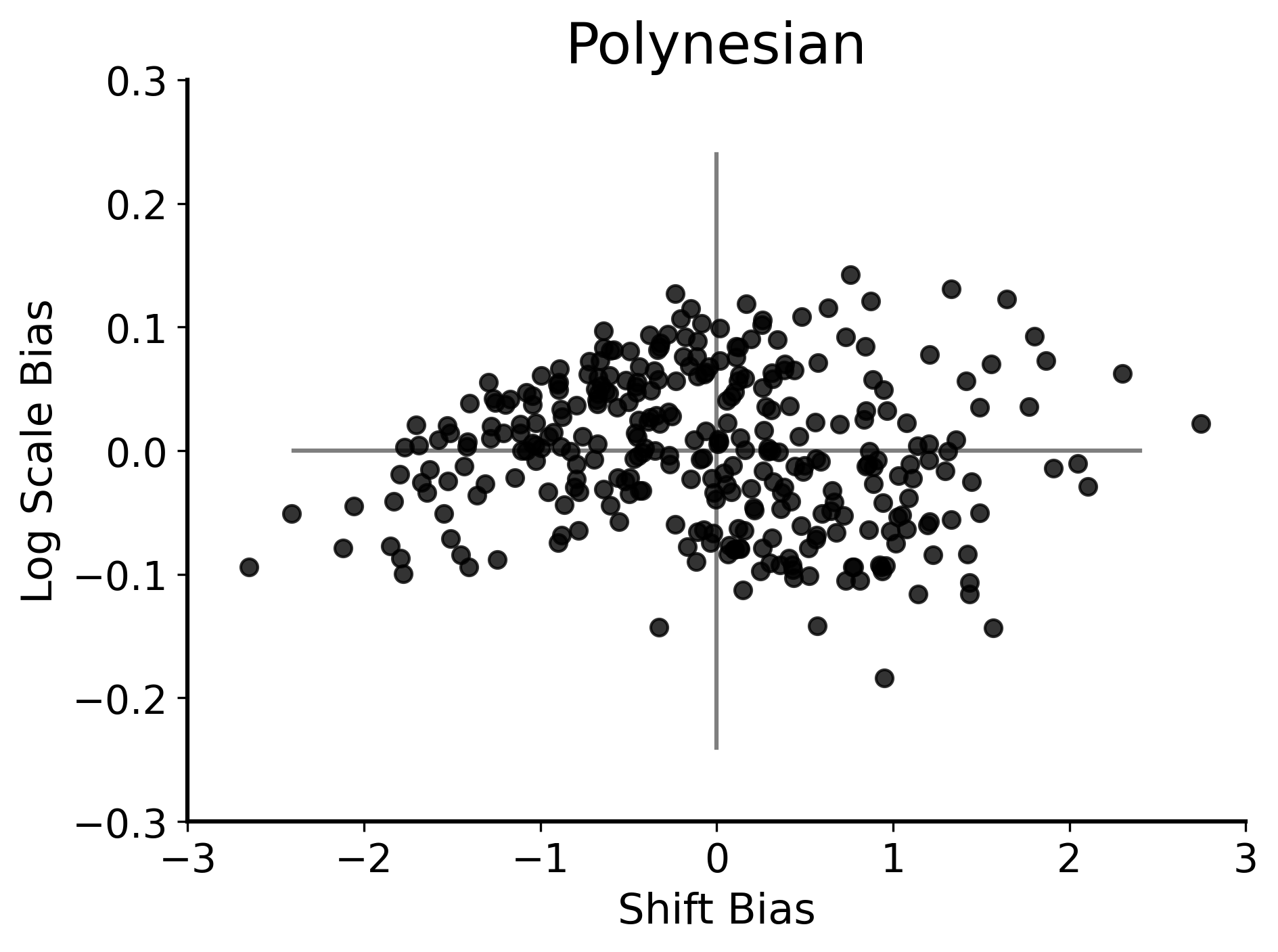}\hspace{-0.17cm}
    \includegraphics[width=2.5cm]{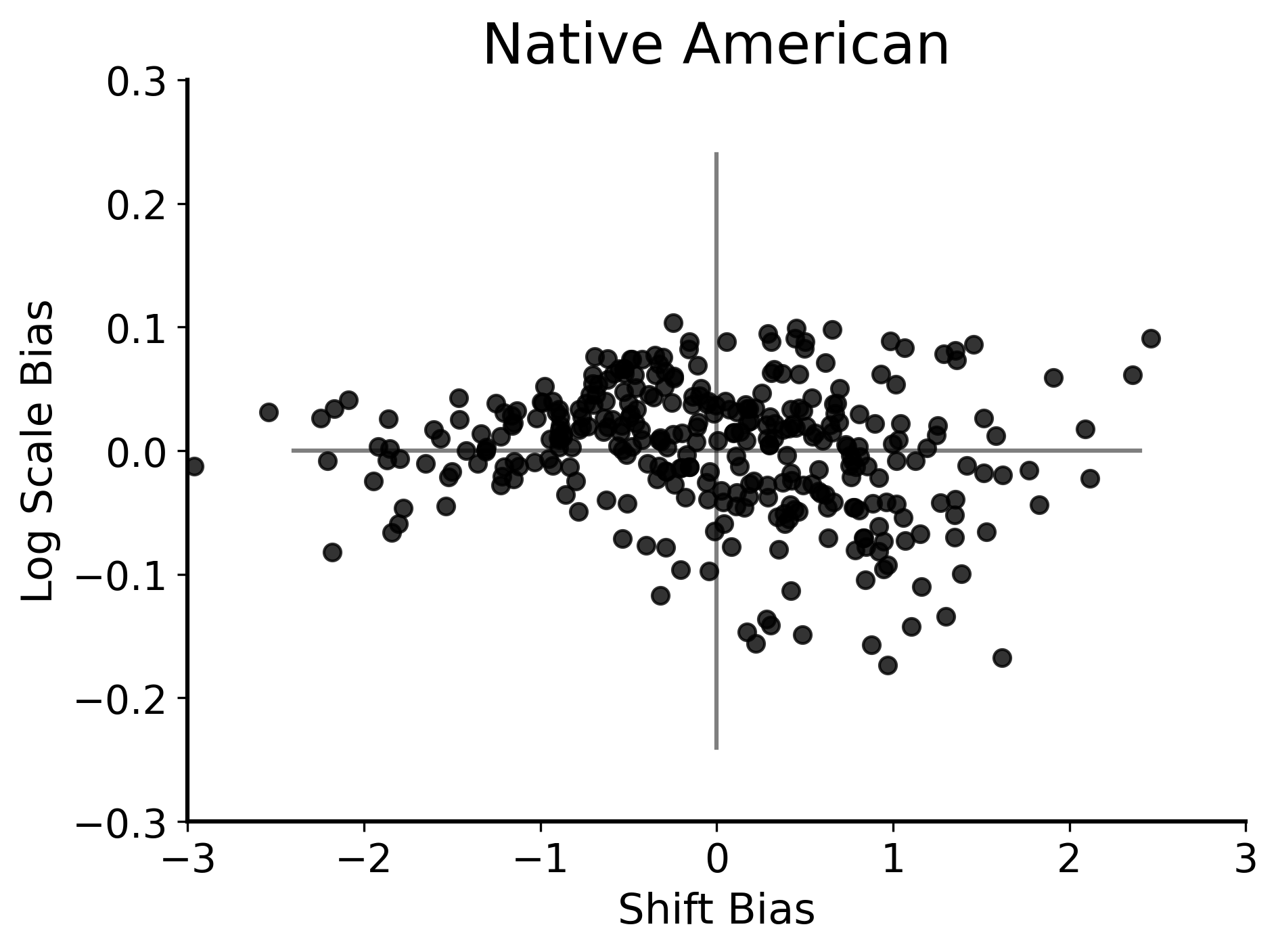}\hspace{-0.17cm}
    \includegraphics[width=2.5cm]{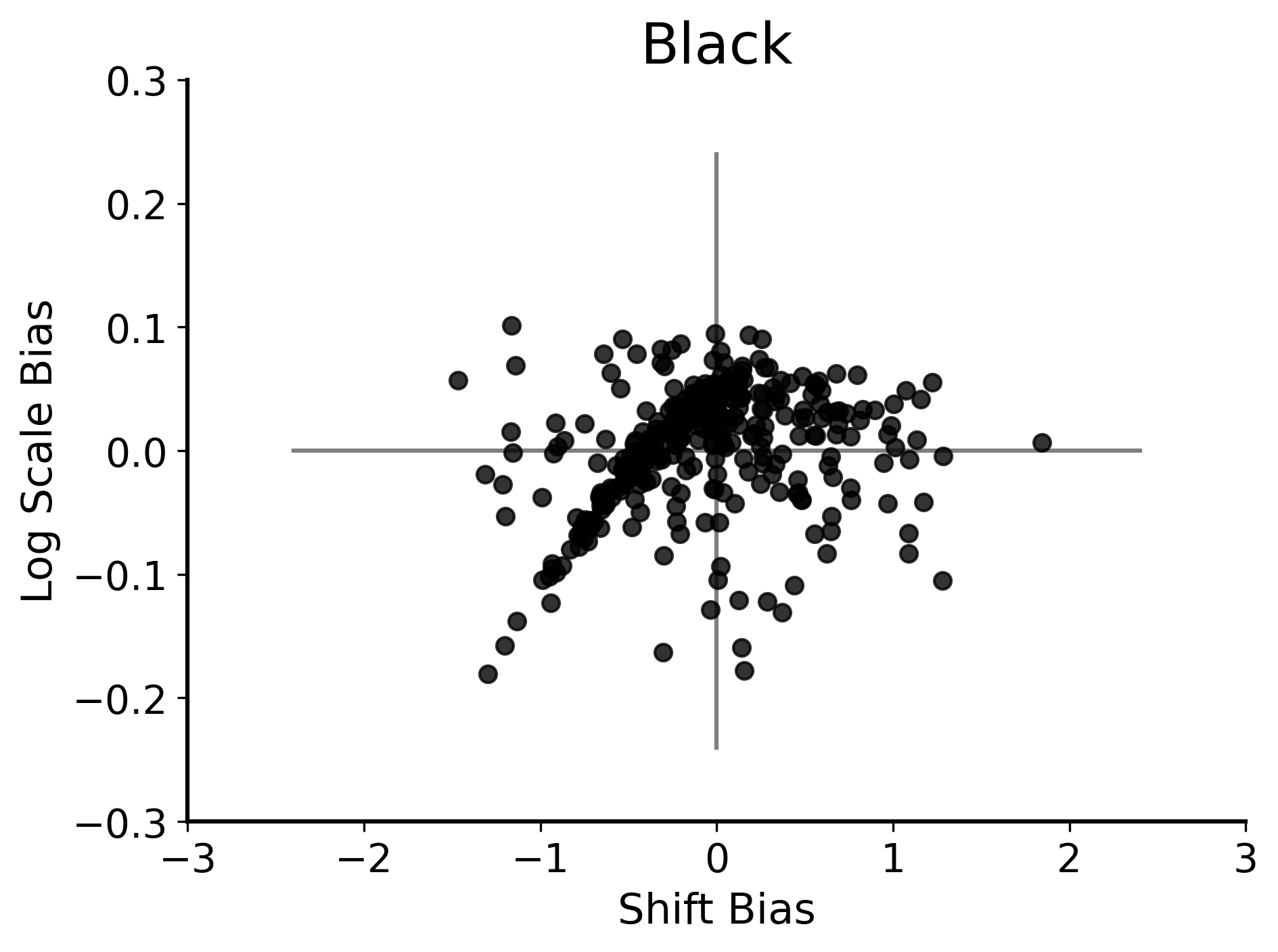}\hspace{-0.17cm}
    \includegraphics[width=2.5cm]{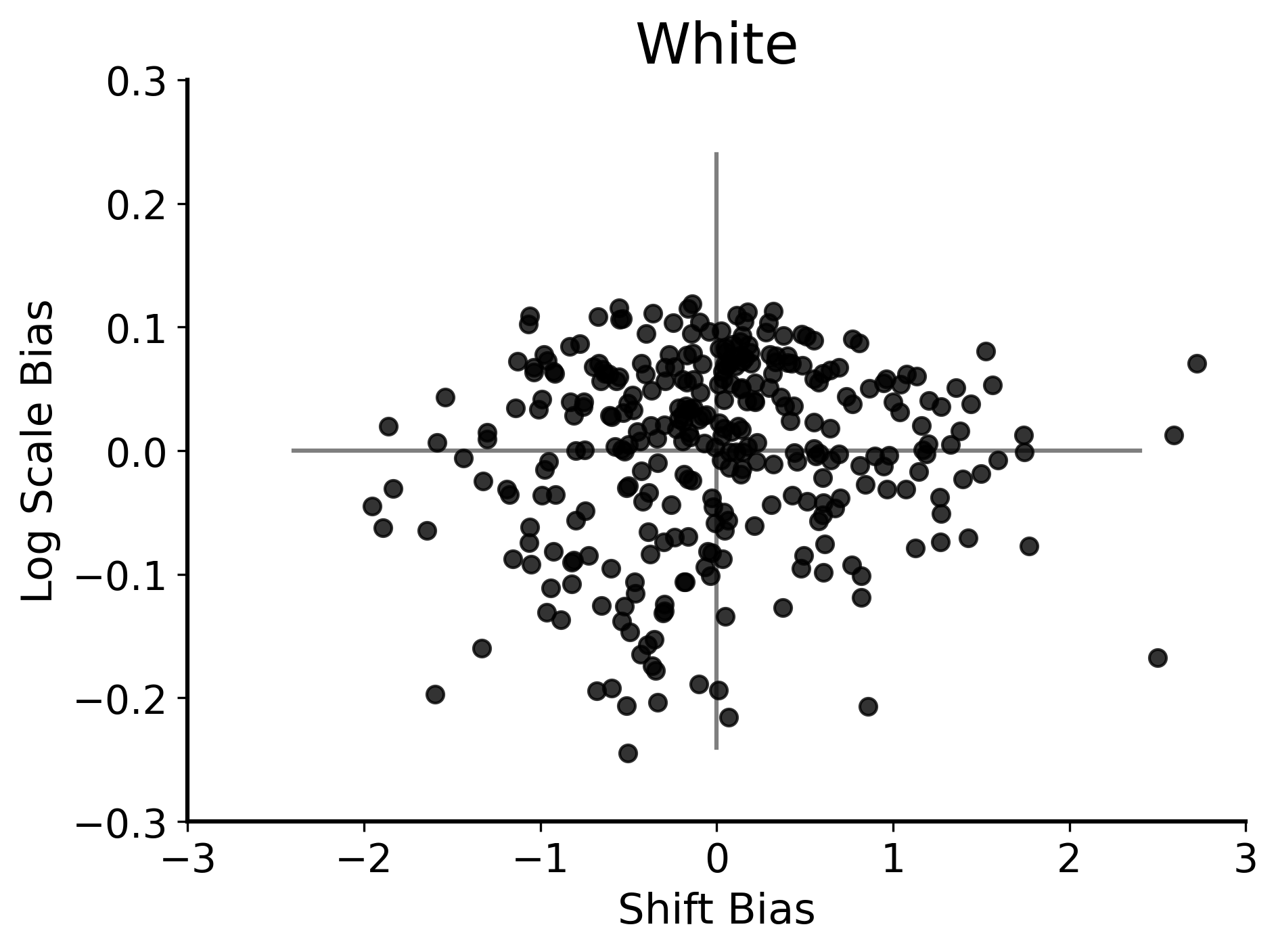}\hspace{-0.17cm}
    \\
    \includegraphics[width=2.5cm]{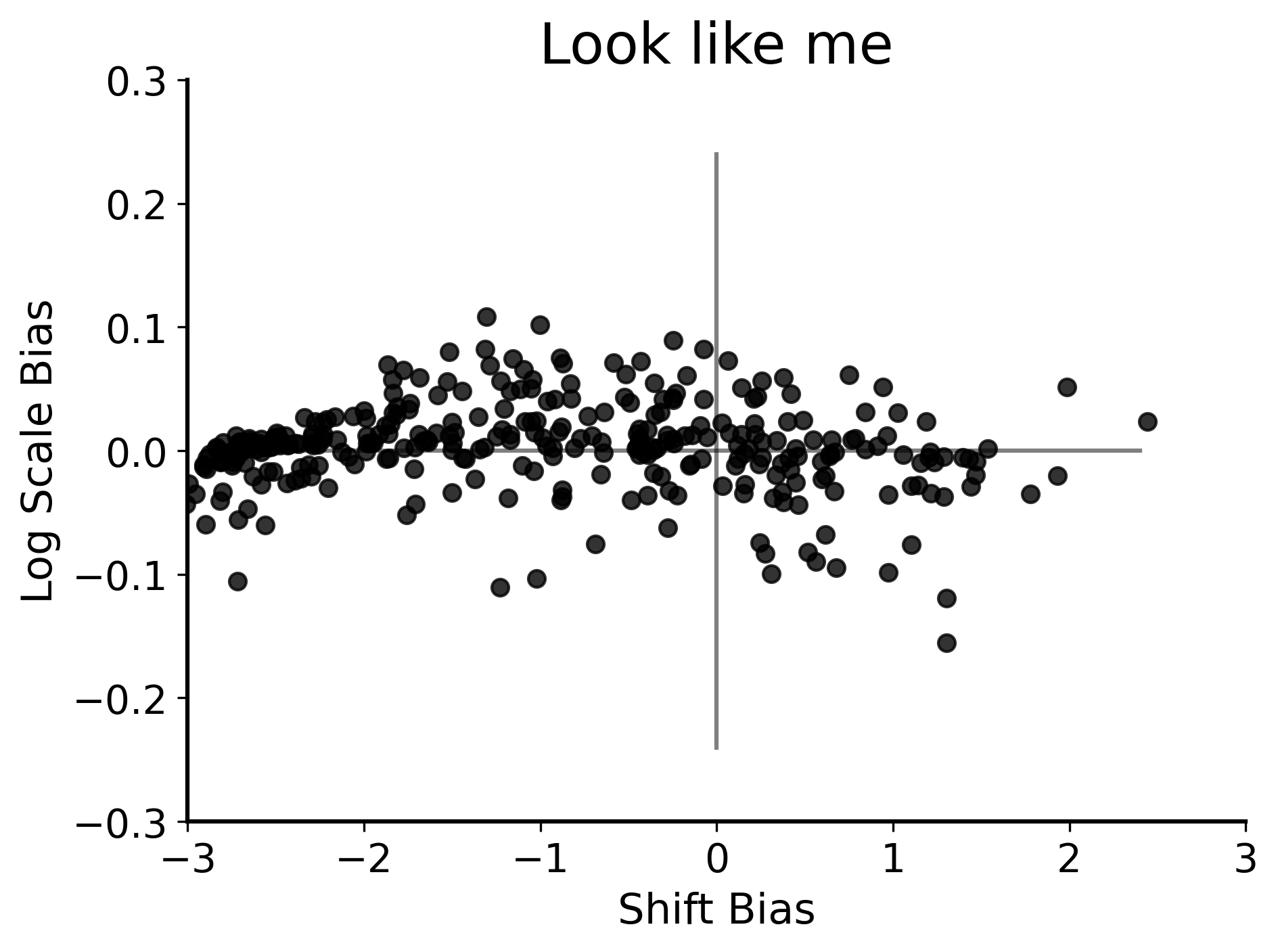}\hspace{-0.17cm}
    \includegraphics[width=2.5cm]{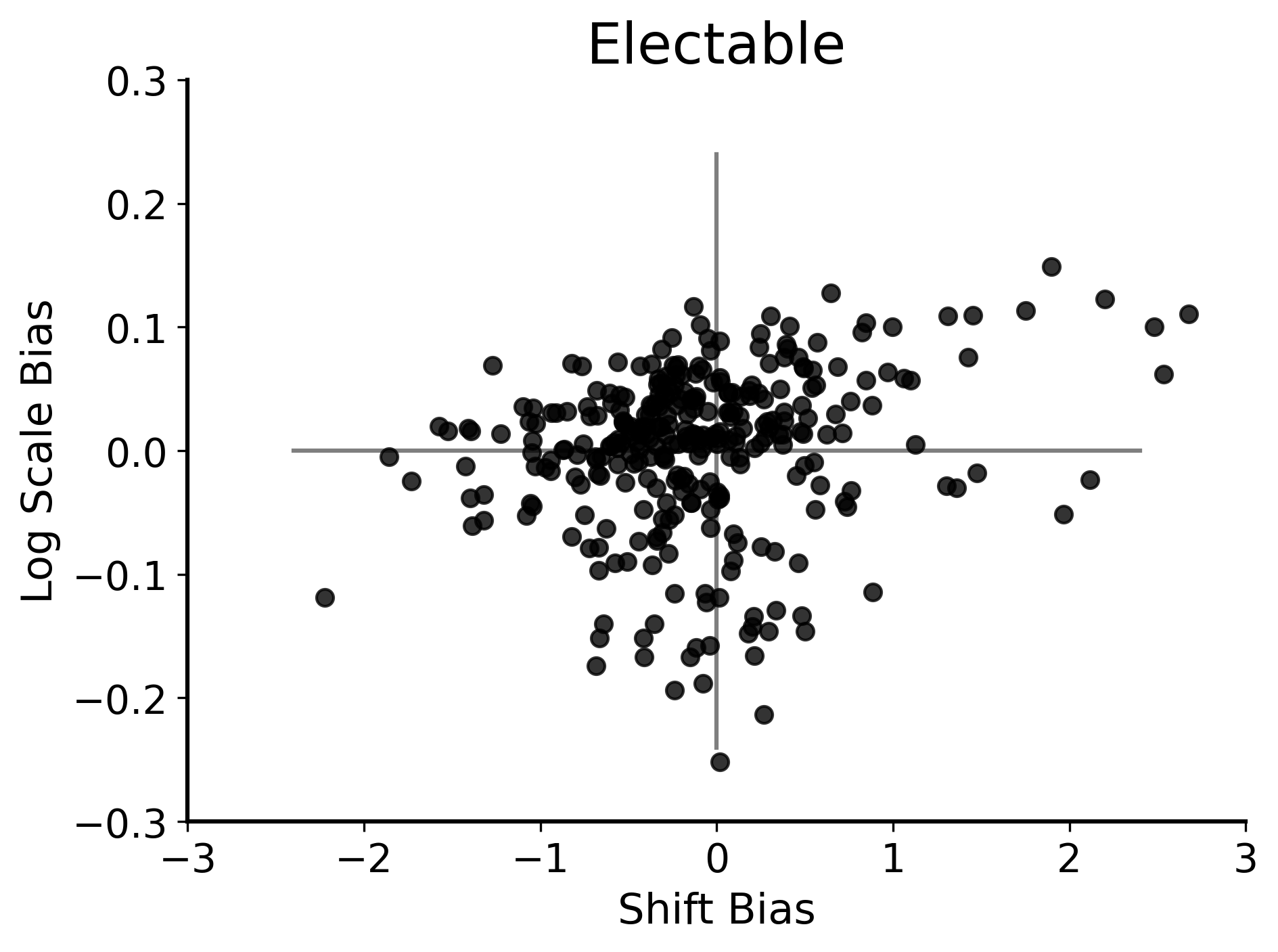}\hspace{-0.17cm}
    \includegraphics[width=2.5cm]{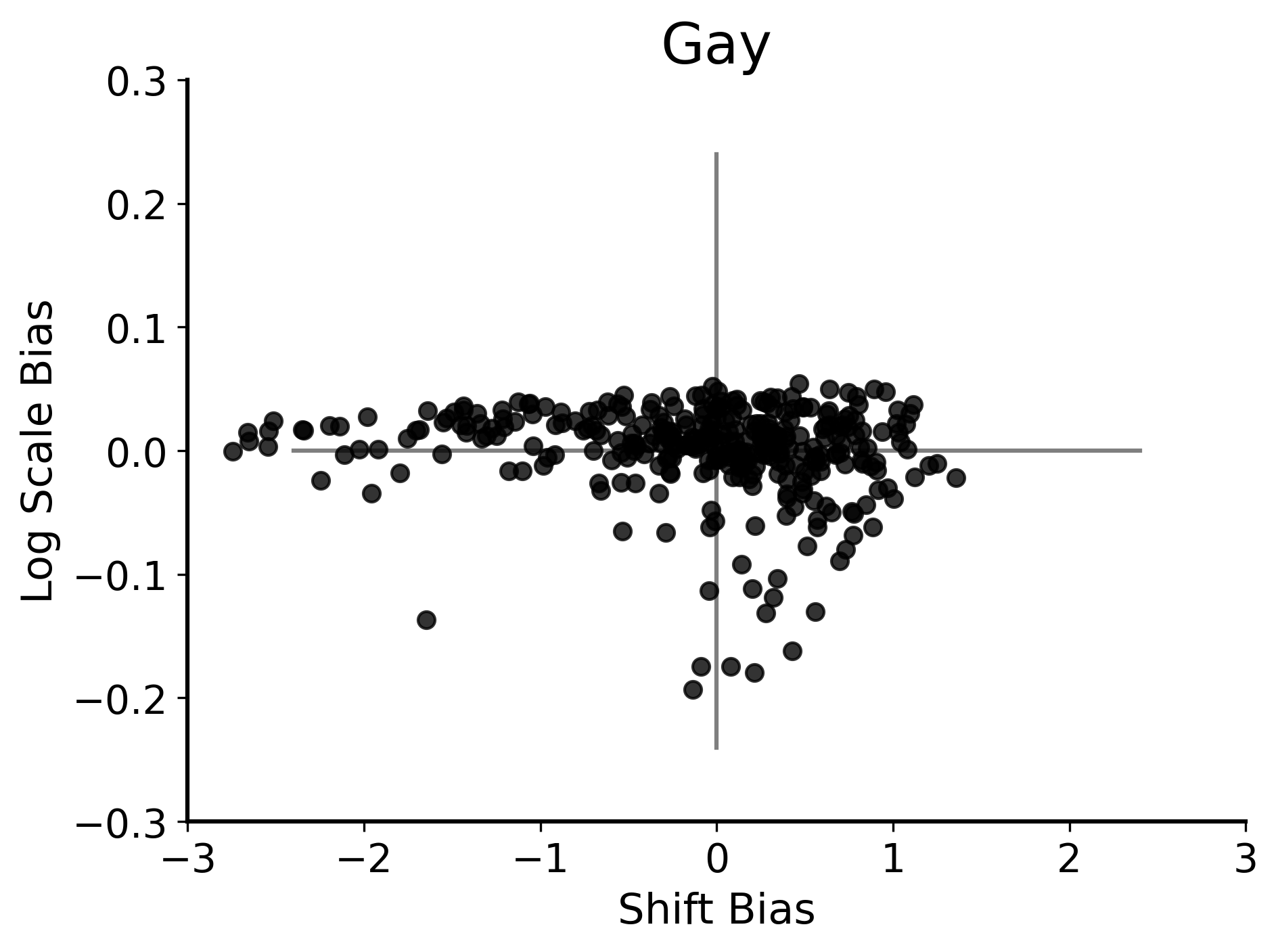}\hspace{-0.17cm}
    \includegraphics[width=2.5cm]{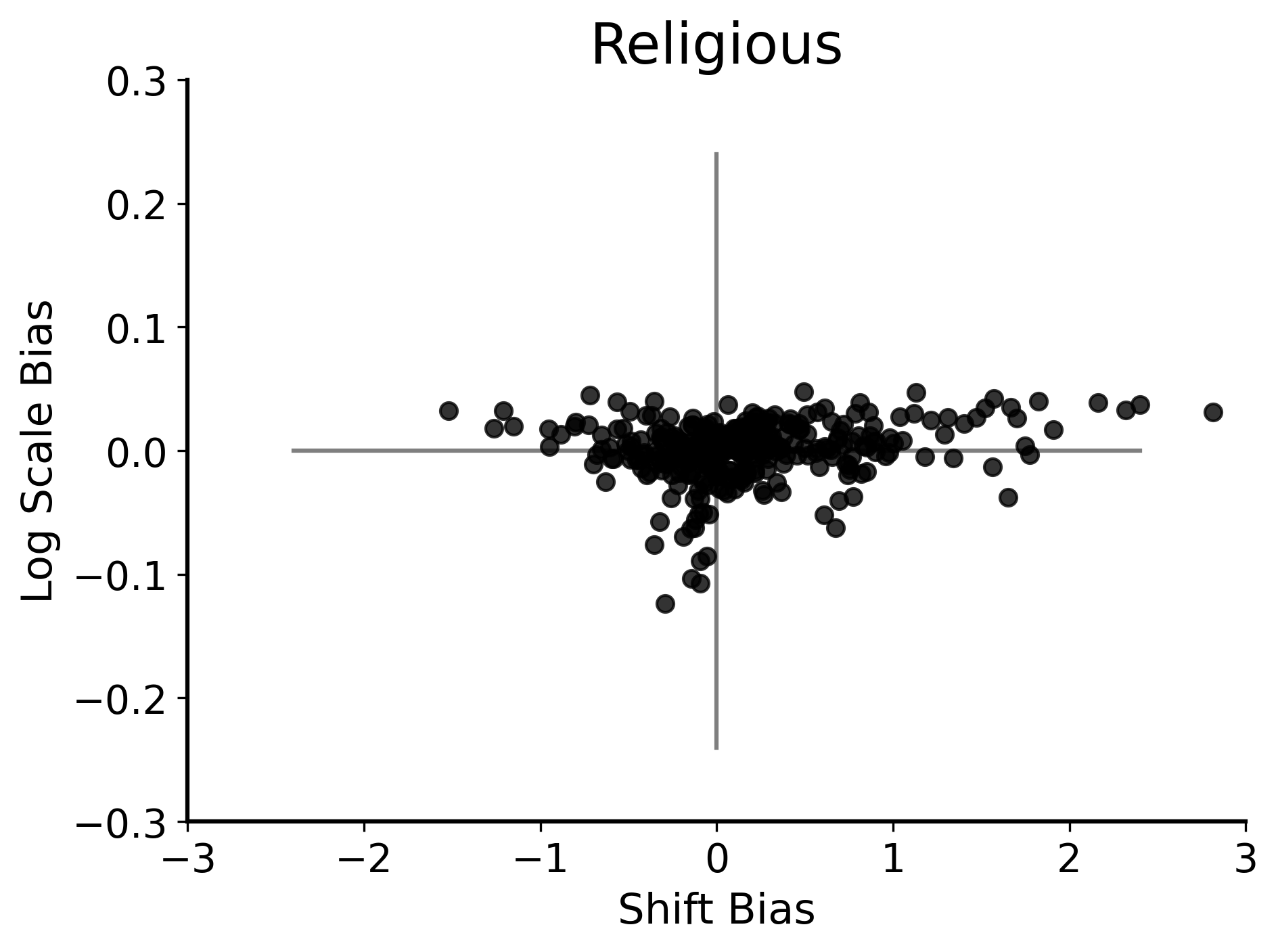}\hspace{-0.17cm}
    \includegraphics[width=2.5cm]{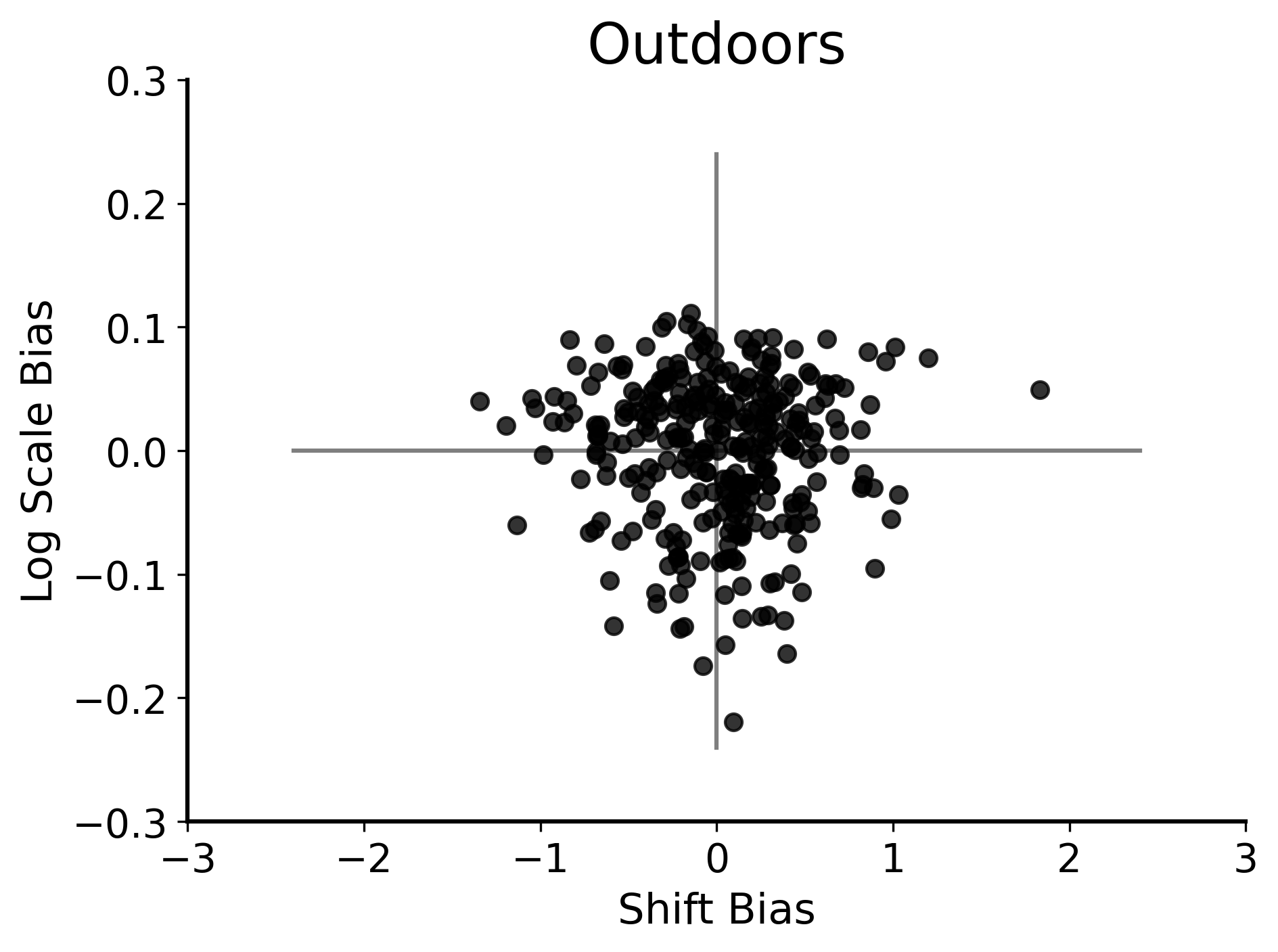}\hspace{-0.17cm}

    \caption{Individual response biases plotted against each other for each participant.}
    \label{fig:your_label}
\end{figure}

To ascertain if there was a dominant culture within a dataset, we measured the proportion of datasets that assigned at least 50\% of respondents to a single culture --- these datasets represent a ``majority'' culture. Out of the 40 datasets, 23 exhibited a majority culture. The datasets with the weakest majority (less than 66\%) included participant-generated-risks, long hair, groomed, alert, liberal, electable, outdoors, fat, and happy. On the other hand, those with the strongest majority (more than 66\%) were technology hazards, daily activities, lay and expert opinions on healthy food, age, familiarity, skin color, hair color, and ethnic backgrounds such as Asian, Middle-Eastern, Hispanic, Native-American, and White. A comparison of the first and third columns of Table 1 reveals that the i\textsc{dlc-cct} improved the fit for 38 out of the 40 concepts evaluated. The predictive accuracy showed an improvement for 32 of the 40 concepts, as evidenced by the comparison between the second and fourth columns in Table 1. These enhancements are more pronounced for the datasets with significant heterogeneity in consensus beliefs among participants because the single-truth model overlooks such variation by assuming that all respondents operate under a single consensus view.


Figure 1 illustrates respondents' response biases. Understanding these biases helps to account for individual differences in responses that cannot be attributed to cultural competency, nor culture assignment. By definition, the average shift bias was 0.0 and the average log scale bias was 0.0. Across participants, the standard deviation of the shift bias was 0.65, while for scale bias, it was 0.08. Many datasets exhibit roughly similar variations in shift and scale biases, but some show notable differences. For instance, in the first impression dataset, the ``age'' trait  has little variation in shift bias (\emph{SD} = 0.10), whereas the religious trait shows opposite pattern, with little variation in scale bias (\emph{SD} = 0.02) but considerable variation in shift bias (\emph{SD} = 1.43).

\subsection{Model Ablation Experiments}

Understanding the contribution and significance of different components within i\textsc{dlc-cct} is important for gaining insights into the model's behavior and understanding where the observed heterogeneity in consensus views arise. In this section, we focus on model ablation experiments, a systematic approach aimed at evaluating the impact of individual components or features on the model’s overall performance \cite{lecun1989optimal}. By selectively removing or neutralizing certain elements of the model, we can isolate the effect that each component has on the outcome. This is instrumental in identifying which aspects of the model are critical to its predictive power, and which may be redundant or even detrimental. Such experiments are especially useful for model interpretation, fine-tuning, and guiding future developments. We used the leadership perception dataset for the parameter ablation experiments. The ablation experiments for other datasets can be found in the Appendix.

\begin{table}
\centering
\begin{tabular}{lccccc}
\hline
\textbf{Model} & \textbf{$R^2$} & \textbf{RMSE} & \textbf{Cultural Entropy} & \textbf{No. Cultures} \\
\hline
i\textsc{dlc-cct}         & .35 & .24 & 2.23 & 6 \\
\hline
\textsc{dlc-cct}             & .31 & .24 & 0.0 & 1 \\
i\textsc{dlc}             & .22 & .26 & 4.00 & 8 \\
\textsc{cct}             & .08 & .29 & 1.07 & 5 \\
Bayesian Ridge Regression & .13 & .28 & .00  & 1 \\
Ablation shift            & .25 & .27 & .84  & 2 \\
Ablation scale            & .31 & .25 & .00    & 1 \\
Ablation competence       & .23 & .26 & 2.60 & 10 \\
Ablation item difficulty  & .24 & .25 & 3.50   & 8 \\
\hline
\end{tabular}
\caption{\label{demo-table} i\textsc{DLC-CCT} Model Ablation.}
\end{table}

In our first ablation experiment, we examined the significance and contribution of the deep latent constructs to the model's performance. To achieve this, we randomly shuffled the embedding feature values across items, disrupting their structure while preserving their marginal distribution. This process eliminates any meaningful relationships that the features might have captured. The i\textsc{dlc-cct} performed poorly when the deep features' structure was disrupted ($R^2$ = 0.0). The results indicate that the content of the pre-trained embeddings is critical for the model's performance.

To examine the significance of each individual and item-level parameters on the predictive performances of i\textsc{dlc-cct}, we conducted parameter ablation experiments. Table 2 presents the results of these ablation experiments, detailing which parameters were held constant to assess their contribution to prediction accuracy. The results reveal that each parameter in i-\textsc{dlc-cct} enhances predictive accuracy. Bayesian Ridge Regression yields the poorest performance, whereas i\textsc{dlc-cct}, when incorporating all the model components, emerges as the most effective. Setting the scale bias to a constant had the least impact on predictive power, and interestingly, it caused the model to output a single consensus belief, indicating a singular cultural cluster. The scale parameter specifically accounts for magnitude preferences, such as a tendency to mark most values at the outer ends or the middle section of the scale. Therefore, taking into account respondents' magnitude preferences may be instrumental in discerning group-level differences.

\subsection{Robustness to Sparse Sampling}
Robustness to sparse sampling refers to a model's ability to sustain performance when faced with limited data. This analysis is especially valuable in situations where data collection and labeling are resource-intensive or challenging. To assess the robustness of our models against sparse sampling, we trained them using incremental data quantities of 5\%, starting from 10\% and increasing up to 80\% of the total data.  In the analyses presented here, we considered the leadership perception dataset. The robustness to sparse sampling experiments for other datasets can be found in the Appendix. Fig. 2 illustrates our robustness analysis, revealing that both the i\textsc{dlc-cct} and \textsc{dlc-cct} models display considerable resilience to sparse sampling. Impressively, even with a significantly reduced training set, constituting merely 10\% of the items, both models could still achieve some level of generalizability (i.e., i\textsc{dlc-cct: $R^2$ = .17; \textsc{dlc-cct: $R^2$ = .15}}). As the quantity of training data increased, the models' performance also improved incrementally. Notably, the models reached parity with the performance achieved with 80\% of the training data while requiring only 65\% of the data to do so.

\renewcommand{\thefigure}{2}
\begin{figure}[h]
\centering
\includegraphics[width=6.5cm, height=6.5cm]{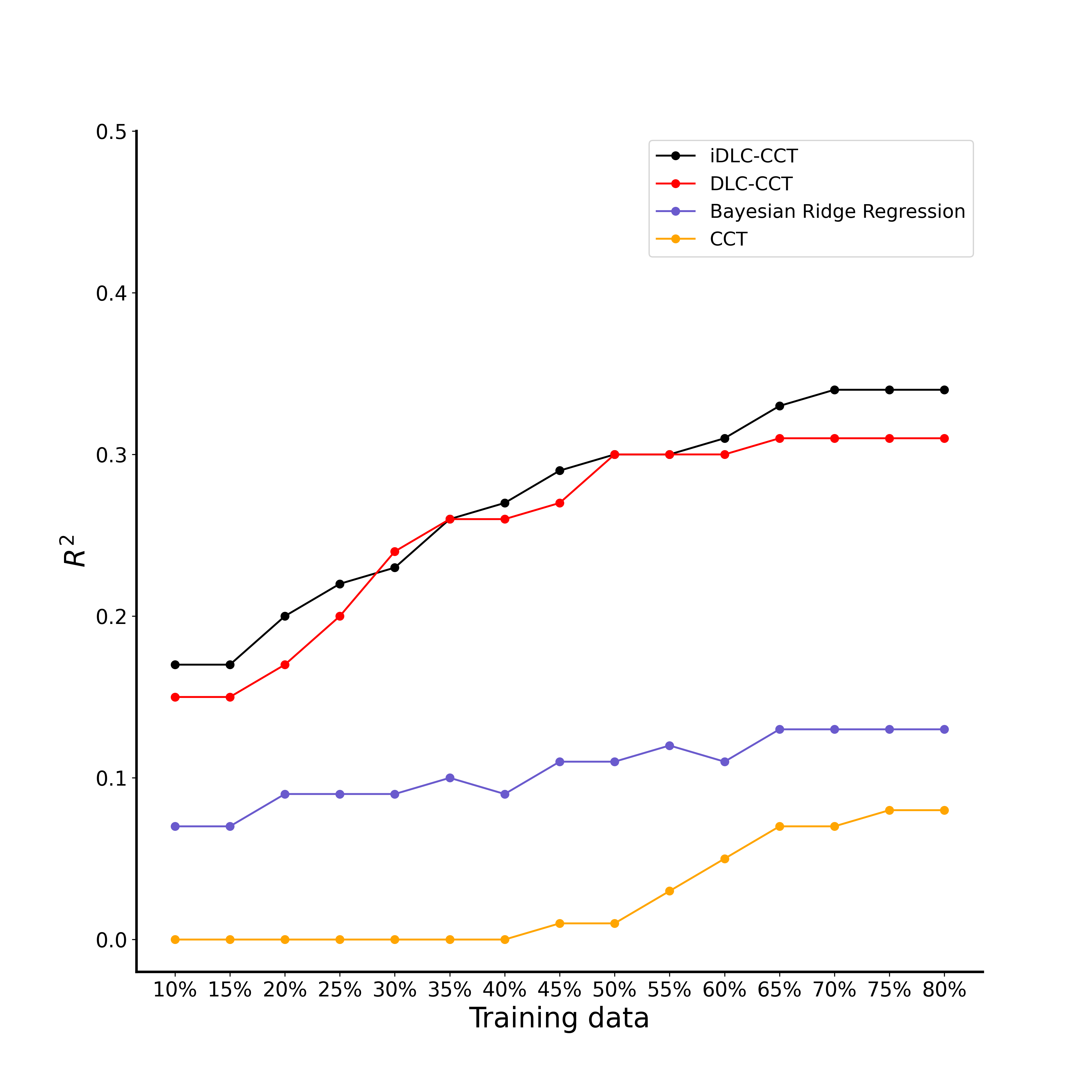}
\caption{Robustness analysis of the i\textsc{DLC-CCT} and \textsc{DLC-CCT} models under sparse sampling conditions. The models were trained with incrementally increasing quantities of data (from 10\% to 80\%) and tested for performance ($R^2$).}
\end{figure}

Compared to i\textsc{dlc-cct} and \textsc{dlc-cct}, the Bayesian Ridge Regression model did not demonstrate a reasonable level of performance under these conditions. This suggests that considering individual and group-level differences is a significant factor in harnessing collective intelligence effectively. Our findings emphasize the value of utilizing models such as i\textsc{dlc-cct} and \textsc{dlc-cct}, which are designed to account for these variances, especially in scenarios where data might be sparse or difficult to gather. The original \textsc{cct} demonstrates a slight improvement in its predictions when there is a larger training dataset because it can better estimate the posterior of the consensus parameter's prior with more data when predicting held-out items.

\subsection{The Discrepancy Between Demographic Features vs. Identified Cultures}

Researchers have used demographic features to explain variations in people's perceptions of concepts and entities \cite{rieger2015risk}. While some studies suggest these features can be indicative, others assert they cannot be solely relied upon \cite{sutherland2020individual}. Although demographic features can sometimes explain these variations, they are not always available for organizations and researchers to incorporate into their computational models. Moreover, in the interconnected society we live in, people's perceptions are influenced by a complex network of factors that extend beyond basic demographic attributes \cite{tsui1989beyond}. A significant part of this complexity is contributed by the easily accessible and diverse array of viewpoints on concepts found on social media, online platforms, and professional and social networks \cite{dubois2018echo}. These platforms not only allow individuals to form unique understandings but also expose them to a variety of perspectives \cite{cinelli2021echo}. 

We continued to use the leadership perception dataset to investigate whether the cultures identified by the i\textsc{dlc-cct} align with demographic features. Fig. 3 shows that demographic features alone do not provide clear insight into the hetereogeneity in perceptions of leadership that are captured by the i\textsc{dlc-cct}. This insight illuminates the multifaceted and nuanced nature of leadership perceptions in our interconnected society, highlighting that these perceptions are not just products of individual demographic attributes, but also of the shared knowledge and perspectives within our interconnected social fabric.

\renewcommand{\thefigure}{3}
\begin{figure}[h]
\centering
\includegraphics[width=8.5cm, height=6.5cm]{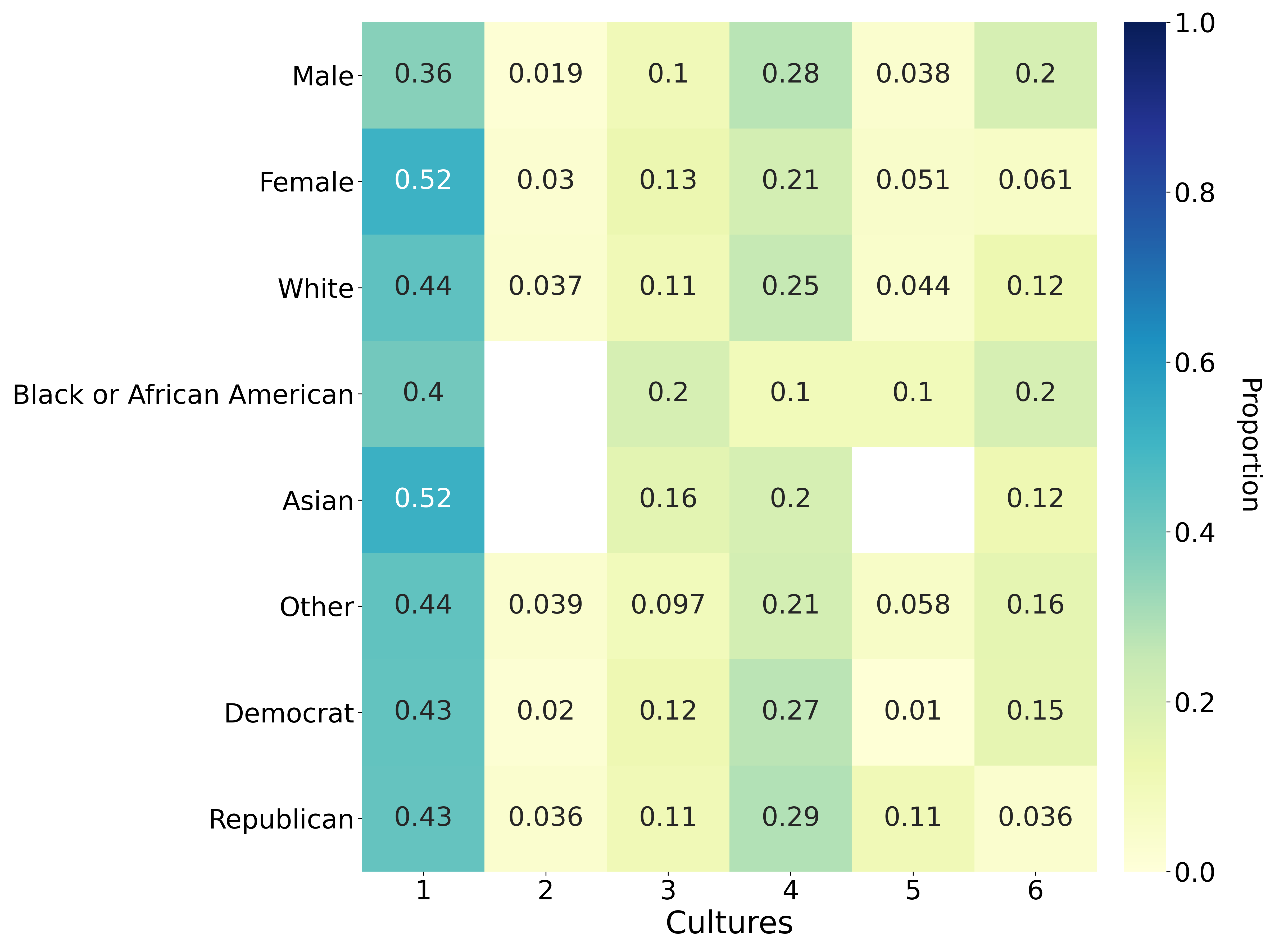}
\caption{Demographic Distribution of Cultural Assignments}
\end{figure}

\section{Discussion}
In this paper, we developed i\textsc{dlc-cct}, an extension of Cultural Consensus Theory that integrates a latent construct to map between pre-trained embeddings of an entity and the consensus belief among one or more respondent subsets concerning those entities. The model effectively aggregates beliefs from individuals, including experts and non-experts, to estimate consensus while identifying idiosyncratic and group-level differences in cultural constructs. By incorporating features from deep neural networks into \textsc{cct}, we can estimate cultural consensus for any entity using pre-trained networks or other available embeddings. We argue that i\textsc{dlc-cct} is a robust foundation for assessing group consensus levels by leveraging the underlying structure and inter-relatedness of beliefs and a foundation for consensus-aware technologies. Our findings reveal that considering group-level consensus variations enhances predictive accuracy and effectively harnesses collective intelligence for decision-making and collaboration within organizations. 

In the subsections that follow, we discuss the scalability of the i\textsc{dlc-cct} model (and multiple-consensus models more generally), describe its possible applications to consensus building and as a computational foundation for consensus-aware information technologies, and consider variants of the model that relax some of its core assumptions.

\subsection{The Small-Variance Asymptotic Approximation}

An ongoing challenge in machine learning involves creating algorithms that can scale to extremely large datasets. Although probabilistic methods --- and Bayesian models in particular --- offer flexibility, the absence of scalable inference techniques may hinder their effectiveness with certain data. For instance, in the context of clustering, $k$-means algorithms are frequently preferred in large-scale scenarios over probabilistic methods like Gaussian mixtures or Dirichlet process (DP) mixtures \cite{teh2010dirichlet}. This is because $k$-means is relatively easy to implement and can manage larger datasets.

Applying an asymptotic analysis to the variance or covariance of distributions within a model has been used to create connections between probabilistic and non-probabilistic models. Examples of this include forming connections between probabilistic and standard PCA by allowing the covariance of the data likelihood in probabilistic PCA to approach zero \cite{tipping1999probabilistic, roweis1997algorithms}. Similarly, the $k$-means algorithm can be obtained as a limit of the EM algorithm when the covariances of the Gaussians associated with each cluster decrease towards zero \cite{kulis2011revisiting}. Not only do small-variance asymptotics offer a conceptual link between different approaches, they can also provide practical alternatives to probabilistic models when dealing with large datasets, as non-probabilistic models often have better scalability. While still emerging, the use of such techniques to derive scalable algorithms from complex probabilistic models offers a promising direction for the development of scalable learning algorithms.

A simple procedure (Algorithm 1) enables the conversion of any existing single-culture \textsc{cct} to an i\textsc{cct} via an algorithm derived from a small-variance asymptotic analysis of the Dirichlet Process \cite{jiang2012small}. One of the primary advantages of the conversion is that it opens the door to novel scalable algorithms for multiple cultures.

We begin with a description of the small-variance asymptotic approximation to the Infinite Gaussian Mixture Model, which alternates through three stages: cluster assignment, cluster updating, and cluster instantiation \cite{kulis2011revisiting}. In cluster assignment, each point is assigned to the cluster that is nearest by Euclidean distance. Initially, all points belong to the same (and only) cluster. In the cluster updating phase, the cluster center is adjusted to the mean of the points assigned to it. During cluster instantiation, the algorithm identifies the point farthest from the center of its assigned cluster, and if that distance exceeds $\lambda$, it establishes a new cluster. The center of the new cluster is positioned at that point, which is then assigned to it. The algorithm repeats until no point is further than $\lambda$ from its cluster center. The parameter $\lambda$ thus acts as a concentration parameter, which, along with the complexity of the data, will determine the number of clusters that eventually appear, much like the concentration hyperparameter of a Dirichlet process that controls the number of clusters in an infinite Gaussian mixture model.

\begin{algorithm}
\caption{Small-Variance Asymptotic Approximation to the Infinite Cultural Consensus Model}
\begin{algorithmic}[1]
    \State \textbf{Input}: Individuals $x_1, \dots, x_n$, $\lambda > 0$
    \State \textbf{Initialization}: $C = \emptyset$, $\textsc{dlc-cct}$
    
    \While{$\forall c \in C, \forall x_i \in c : CC(x_i, c) < \lambda$}     
        \State \Comment{\textbf{Culture Assignment}} 
        \For{each $x_i$}
            \State  $\text{CC}(x_i, C)= {\text{CC}(x_i,c_1),\text{CC}(x_i,c_2),...,\text{CC}(x_i,c_n))}$ 
            \State $c_{\text{max}} = \arg\max_{c} CC(x_i, c)$
            \If{$CC(x_i, c_{\text{max}}) \geq \lambda$}
                \State Assign $x_i$ to $c_{\text{max}}$
            \Else 
                \State $c_{new} = {\text{logit}(x_{i1}), \text{logit}(x_{i2}), ..., \text{logit}(x_{im})}$
                \State Assign $x_i$ to $c_{\text{new}}$
                \State Add $c_{\text{new}}$ to $C$
            \EndIf
        \EndFor
        
        \State \Comment{\textbf{Culture Updating}}
        \For{each $c$}
            \State $C_c = \textsc{dlc-cct} (X_c)$
        \EndFor
        
        \State \Comment{\textbf{Culture Instantiation}}
        \For{each $c$}
            \State $x_{\text{min}} = \arg\min_{x_i \in c} \text{CC}(x_i, c)$
            \If{$\text{CC}(x_{\text{min}}, c) < \lambda$}
                \State $c_{\text{new}} = \text{logit}(x_{\text{min}})$
                \State Assign $x_{\text{min}}$ to $c_{\text{new}}$
                \State Add $c_{\text{new}}$ to $C$
            \EndIf
        \EndFor
    \EndWhile
    
    \State \textbf{Output}: $C$
\end{algorithmic}
\end{algorithm}

By analogy, the small-variance asymptotic approximation to the Infinite Cultural Consensus Model likewise proceeds through three steps: culture assignment, culture updating, and culture instantiation. In culture assignment, each respondent is assigned to the culture for which they exhibit the highest cultural competence. Since the continuous version of cultural consensus theory models ratings as a logistic transformation of the latent appraisal, the assigned culture will be the one whose cultural consensus values correlate highest with the participant's logit-transformed responses. Initially, all respondents belong to the same (and only) culture. In culture updating, consensus values for each culture are determined by fitting a single-truth cultural consensus model to the data from respondents assigned to that culture. And during culture instantiation, the algorithm examines the participant with the lowest cultural competence for their assigned culture. If that respondent's cultural competence is less than $\lambda$, a new culture is created with consensus values set to the logit-transformed responses of the respondent, who is then assigned to it.

The small-variance asymptotic approximation facilitates the conversion of existing single-truth cultural consensus models to the infinite cultural consensus model. There are two preconditions. First, the model must provide a means to compute the cultural consensus, which is precisely the function of existing single-truth cultural consensus methods and thus presents no issue. Second, the model must enable one to identify the culture for which a respondent exhibits the highest cultural competence. The readiness of this capability under existing methods is varied. As previously discussed, for the continuous model, it involves calculating the linear correlation between the respondent's logit-transformed responses and the consensus for each culture, then identifying the culture with the highest correlation. When these conditions are met, the single-truth consensus models can be scaled to an efficient multiple-truth consensus model.

\subsection{Applications and Considerations}

Next, we consider potential applications of the i\textsc{dlc-cct}. One such application is as a computational foundation for consensus-aware information technologies. Our findings demonstrate the potential of methods to detect and characterize heterogenity in consensus beliefs. Building on these methods, one could integrate i\textsc{dlc-cct} into information systems to enable interventions that help to detect and understand a lack of consensus within a purported group or to build consensus among the members of a team to foster improved decision-making and collaboration. Beyond detecting and characterizing heterogeneity in consensus beliefs, i\textsc{dlc-cct} also provides insights at both the question and individual levels, deepening our comprehension of the domain of interest. For instance, by analyzing variations in question difficulty across cultures, a consenus-aware information technology could pinpoint questions that some subsets of respondents may find challenging to agree upon or identify controversial topics unique to specific cultural contexts. Such insights could assist in tailoring interventions and strategies to promote collaboration and consensus-building. By recognizing diverse opinions, a consensus-aware algorithm can generate proposals that effectively represent various perspectives, promoting a more inclusive and equitable approach to consensus formation. It paves the way for information technology applications that either harness collective intelligence under a lack of cross-cultural consensus.

Another promising application of \textsc{cct} is its integration into AI systems to enhance their interoperability. By leveraging \textsc{cct}'s unique capability to interpret shared beliefs within cultural contexts, we can equip AI systems with a more culturally nuanced understanding of consensus beliefs. This integration could significantly enhance these models' generalizability. The integration of \textsc{cct} could also increase the transparency of AI systems by rendering their decision-making processes more intelligible; decisions could be attributed to culturally shared beliefs or norms. Furthermore, applying \textsc{cct} could foster a greater level of trust in AI systems among users. By ensuring that the systems' decision-making processes align more closely with shared beliefs, we can develop systems that are ethically sound and culturally sensitive. Even so, the successful integration of \textsc{cct} into AI systems will require carefully designed methods to translate a consensus derived from \textsc{cct} into workable AI systems.

The i\textsc{dlc-cct} can also be used to support the consensus-building process by surfacing the causes of disagreement between respondents, whether those are individual-level parameters such as shift and scale biases or competency, or culture-level parameters such as the consensus. However, using i\textsc{dlc-cct} for consensus-building may be challenging under adversarial behavior. Consider, for example, voting mechanisms or other consensus-building methods, where knowledge of the aggregation scheme could potentially allow individuals or groups to strategically manipulate the system by skewing their responses, misreporting their preferences, or coordinating their actions with others to drive the consensus toward a desired value \cite{fiedler1996explaining}. This is sometimes known as ``gaming the system'' \cite{fiedler1996explaining}. It is an open question as to the extent to which i\textsc{dlc-cct} is robust to gaming and the most effect means of reducing the risk of such strategic behavior to ensure that each individual's input is a genuine reflection of their own beliefs, views, or preferences, rather than a strategic attempt to influence the final outcome. Doing so can lead to more robust and representative consensus outcomes and more effect consensus-building tools. In summary, while a respondent's naivety about the aggregation scheme may help ensure the genuineness and objectivity of the consensus process, it also implies the need for the design of robust, resistant-to-manipulation aggregation schemes, and vigilance towards potential strategic behaviors.

The i\textsc{idlc-cct} can be extended in various ways. First, building on the insights derived from the i\textsc{idl-cct} approach, researchers and organizations can further explore the potential of multimodal fusion models in capturing and modeling cultural consensus. People perceive the world by simultaneously processing and merging high-dimensional inputs from multiple modalities, including vision and semantic meaning \cite{smith2005development}. With the advancements in deep learning and computing power, multimodal fusion models have gained popularity for their ability to combine different modalities in a way similar to human perception, leading to more accurate predictions of human judgments \cite{ramachandram2017deep}. By integrating these multimodal models within the i\textsc{dlc-cct} framework, future research can combine information from various sources to achieve better alignment between machine representations and cultural consensus, further enhancing our understanding of the complex interplay between culture and perception.

Finally, we note that in the i\textsc{dlc-cct} model, we rely on the Dirichlet Process, which assumes the existence of an infinite number of potential cultures, with each individual belonging to one specific culture. While this simplification facilitates mathematical modeling and computational tractability, it may not accurately represent the multifaceted, overlapping, and fluid nature of cultural identities in reality. Real-world cultures are seldom mutually exclusive, and individuals often navigate multiple cultural spaces simultaneously. They may adhere to one cultural norm in a certain context and another in a different setting. Additionally, cultural affiliations are not static; they can evolve, intersect, and change over time based on personal experiences, societal influences, migration, and many other factors. Further, the distribution over the size of clusters under the DP has exponential tails, which may or may not be a reasonable assumption regarding how respondents are spread among cultures. Extensions of the present work to the Pitman-Yor process could relax this assumption by allowing for non-exponential tail behavior, such as power-law tails. Therefore, while the Dirichlet Process provides a practical method for dealing with an unknown number of cultures, it is essential to understand its limitations. Future studies can explore more intricate models that better capture the dynamic and overlapping nature of cultural identities while relaxing some of the assumptions made about cluster assignments.

\subsection{Conclusion}
In conclusion, we presented the i\textsc{dlc-cct}, which allows culturally held beliefs to be transformed into fine-tuned machine representations. These representations map features of a concept or entity to the consensus response among a subset of respondents using the Dirichlet Process. This method integrates the strengths of the cultural consensus model with machine learning techniques, overcoming their respective limitations. It has demonstrated both predictive and explanatory benefits. The i\textsc{dlc-cct} method offers both scientific and practical applicability, allowing researchers to study group-level variation while benefiting from the latest advances in machine learning. The advantages of incorporating heterogeneity in consensus beliefs are evident, and i\textsc{dlc-cct} proves invaluable for analyzing and devising strategies that promote consensus-building. Through the aggregation of information, it advances collective intelligence. The integration of i\textsc{dlc-cct} into technological systems can enhance our understanding of consensus beliefs, deepening our knowledge of the relevant domain.
\bibliographystyle{unsrt}  
\bibliography{references}

\end{document}